\documentclass{article}

\usepackage{microtype}
\usepackage{graphicx}
\usepackage{subfigure}
\usepackage{booktabs} 

\usepackage{hyperref}

\usepackage[accepted]{icml2025}

\usepackage{amsmath}
\usepackage{amssymb}
\usepackage{mathtools}
\usepackage{amsthm}
\usepackage{bm}

\usepackage[capitalize,noabbrev]{cleveref}

\theoremstyle{plain}
\usepackage{amsmath}
\newtheorem{theorem}{Theorem}[section]

\newtheorem{lemma}[theorem]{Lemma}

\theoremstyle{definition}

\theoremstyle{remark}

\usepackage[textsize=tiny]{todonotes}

\icmltitlerunning{Bellman error centering}

\begin{document}

\twocolumn[
\icmltitle{Bellman Error Centering}




\begin{icmlauthorlist}
\icmlauthor{Xingguo Chen}{yyy}
\icmlauthor{Yu Gong}{yyy}
\icmlauthor{Shangdong Yang}{yyy}
\icmlauthor{Wenhao Wang}{sch}
\end{icmlauthorlist}

\icmlaffiliation{yyy}{Nanjing University of Posts and Telecommunications, Nanjing, China}
\icmlaffiliation{sch}{National University of Defense Technology, Hefei, China}

\icmlcorrespondingauthor{Wenhao Wang}{wangwenhao11@nudt.edu.cn}

\icmlkeywords{Reinforcement Learning}

\vskip 0.3in
]



\printAffiliationsAndNotice{\icmlEqualContribution} 

\begin{abstract}
This paper revisits the recently proposed reward centering algorithms 
including simple reward centering (SRC) and value-based reward centering (VRC), 
and points
out that SRC is indeed the reward centering, while
VRC is essentially Bellman error centering (BEC). Based on
BEC, we provide the centered fixpoint for tabular value
functions, as well as the centered TD fixpoint for linear value function
approximation. We design the on-policy CTD algorithm and the off-policy CTDC
algorithm, and prove the convergence of both algorithms. Finally, we
experimentally validate the stability of our proposed algorithms. 
Bellman error centering facilitates the extension to various
reinforcement learning algorithms.
\end{abstract}

\section{Introduction}
\label{introduction}
Reinforcement learning (RL) has driven transformative advances 
in strategic decision-making (e.g., AlphaGo \citep{silver2016mastering}) and 
human-aligned large language systems 
(e.g., ChatGPT via RLHF \citep{ouyang2022training,carta2023grounding,dai2024safe,guo2025deepseek}). 
However, these breakthroughs incur prohibitive computational costs:
AlphaZero requires billions of environment interactions, while
 RL-based training of state-of-the-art language models demands
  millions of GPU hours \citep{patterson2021carbon}.
   Such challenges necessitate urgent 
   innovations in RL efficiency to enable scalable 
    AI development.
    
 
 To tackle long-term and continuous reinforcement learning problems,
  two primary maximization objectives have been proposed: 
  the average reward criterion and the discounted reward criterion.
 In the context of average reward, 
 \citet{schwartz1993reinforcement} introduced
 R-Learning, employing an adaptive method to estimate average rewards.
 \citet{das1999solving} proposed SMART, which focuses on 
 estimating average rewards  directly. 
 \citet{abounadi2001learning} introduced RVI Q-learning, utilizing the value of a
 reference state to enhance the learning process. 
 \citet{yang2016efficient} proposed
 CSV-learning, which employs a constant shifting values to improve convergence.
 \citet{wan2021learning} removed reference state of RVI Q-learning, proposed
 differential Q-learning and differential TD learning.

Regarding discounted rewards, \citet{perotto2018tuning,grand2024reducing}  highlighted that, under certain
conditions, such as with a large discount factor, Blackwell optimality can be
achieved. \citet{sun2022exploit} demonstrated that reward shifting can effectively
accelerate convergence in deep reinforcement learning. 
\citet{schneckenreither2025average} introduced near-Blackwell-optimal Average Reward Adjusted
Discounted Reinforcement Learning using Laurent Series expansion of the discounted
reward value function. 
\citet{naik2024reward,naik2024reinforcement} proposed the concept of reward centering,
designing simple reward centering and value-based reward centering, and proved
the convergence of tabular Q-learning with reward centering. 
Applying reward centering in tabular Q-learning, 
Q-learning with linear function approximation 
and Deep Q-Networks (DQN) have produced outstanding experimental 
results across all these approaches \citep{naik2024reward}.    
 
However, three issues remain unresolved:
(1) \citet{naik2024reward,naik2024reinforcement}
 pointed out that while reward centering can be combined with 
 other reinforcement learning (RL) algorithms, 
 the specific methods for such integration are 
 not straightforward or trivial. 
 The underlying mechanisms of reward centering warrant further investigation.
(2) Currently, there is only a convergence proof for tabular
 Q-learning with reward centering, 
 leaving the convergence properties of reward centering
  in large state spaces with function approximation still unknown.
(3) If the algorithm converges, what solution it will converge to
 is also an open question.

 In response to these three issues, 
 \textbf{the contributions of this paper} are as follows:
(1) We demonstrated that value-based reward centering is essentially Bellman
error centering. 
(2) Under linear function approximation,  the solution of
Bellman error centering converges to
the centered TD fixpoint.
(3) Building on Bellman error centering, we designed centered temporal difference
learning algorithms, referred to as on-policy CTD and off-policy CTDC, respectively.
(4) We provide convergence proofs under standard assumptions.

\section{Background}
\label{preliminaries}
\subsection{Markov Decision Process}
Reinforcement learning agent keeps interactions with an environment. 
In each interaction $t$, it observes the state $s_t$,
 takes an action $a_t$ to influence the environment, and obtains
 an immediate reward $r_{t+1}$.
 Consider an infinite-horizon Markov Decision Process (MDP), defined by a tuple $\langle S,A,R,P
 \rangle$, where $S=\{1,2,\ldots,N\}$ is a finite set of states of the environment;  $A$
 is a finite set of actions of the agent; 
 $R:S\times A \times S \rightarrow \mathbb{R}$ is a bounded deterministic reward
 function; $P:S\times A\times S \rightarrow [0,1]$ is the transition
 probability distribution \cite{Sutton2018book}.  
 
 A policy is a mapping $\pi:S\times A \rightarrow [0,1]$. The goal of the
 agent is to find an optimal policy $\pi^*$ to maximize the expectation of a
 discounted cumulative rewards over a long period. 
 State value function $V^{\pi}(s)$ for a stationary policy $\pi$ is 
 defined as:  
 \begin{equation}
 V^{\pi}(s)=\mathbb{E}_{\pi}[\sum_{t=0}^{\infty} \gamma^t r_{t+1}|S_0=s],
 \label{valuefunction}
 \end{equation}
 where $\gamma\in (0,1)$
 is a discount factor.
 We have Bellman equation for each state:
 \begin{equation}
 V^{\pi}(s)=\mathbb{E}_{\pi}[ r+\gamma  V^{\pi}(s')],
 \end{equation}
 where $s'$ is the succesor state of state $s$.
 In vector form,
  \begin{equation}
  \begin{split}
 \bm{V}^{\pi}&=\bm{R}^{\pi}+\gamma \bm{\mathbb{P}}^{\pi} \bm{V}^{\pi}\\
 &\dot{=} \bm{\mathcal{T}}^{\pi}\bm{V}^{\pi},
 \end{split}
 \end{equation}
 where $R^{\pi}(s)=\sum_{a}\pi(a|s)\sum_{s'}P(s,a,s')R(s,a,s')$,
 transition probability matrix of states 
 $[\mathbb{P}^{\pi}]_{s,s'}=\sum_{a}\pi(a|s)\sum_{s'}P(s,a,s')$,
 and $\bm{\mathcal{T}}^{\pi}$ is Bellman prediction operator.
 $\bm{V}^{\pi}$ is the fixpoint solution of $\bm{V}=\bm{\mathcal{T}}^{\pi}\bm{V}$.
 
  To deal with large scale state space, 
  linear value function for state $s\in S$ is defined as:
\begin{equation}
 V_{{\bm{\theta}}}(s):= {{\bm{\theta}}}^{\top}{{\bm{\phi}}}(s) = \sum_{i=1}^{m}
\theta_i \phi_i(s),
\label{linearvaluefunction}
\end{equation}
 where ${{{\bm{\theta}}}}:=(\theta_1,\theta_2,\ldots,\theta_m)^{\top}\in
 \mathbb{R}^m$ is a parameter vector, 
 ${{\bm{\phi}}}:=(\phi_1,\phi_2,\ldots,\phi_m)^{\top}\in \mathbb{R}^m$ is a feature
 function defined on state space $S$, and $m\ll |S|$ is the feature size. 
 
However, in the parameter space, 
$\bm{V}_{{\bm{\theta}}}$ cannot be guaranteed to be equal to 
$\bm{\mathcal{T}}^{\pi}\bm{V}_{{\bm{\theta}}}$.
 We typically solve for the TD fixpoint \citep{sutton2008convergent,sutton2009fast}, as follows:
 \begin{equation}
  \bm{V}_{{\bm{\theta}}}=\bm{{\bm{\Pi}}}\bm{\mathcal{T}}^{\pi}\bm{V}_{{\bm{\theta}}},
 \end{equation}
 where projection ${\bm{\Pi}}={\bm{\Phi}}({\bm{\Phi}}^{\top}\textbf{D}{\bm{\Phi}})^{-1}{\bm{\Phi}}^{\top}\textbf{D}$, 
 $\textbf{D}$  is a diagonal matrix of distribution vector $\bm{d}$, each element $\bm{d}_{s}$
  represents the distribution of state $s$.
 In expectation form, it is equal to 
 \begin{equation}
 \mathbb{E}_{\pi}[\delta\bm{\phi}]=0,
 \end{equation}
 where TD error  $\delta=r+\gamma V_{\bm{\theta}}(s')-V_{\bm{\theta}}(s)$. 
 
 In on-policy learning, the target policy $\pi$ and the behavior policy $\mu$
  are the same,
  and the experience is sampled 
  as $\langle s_t,a_t,r_{t+1},s_{t+1}\rangle$.
  The update rule of the on-policy TD learning \citep{sutton2016emphatic} is
  \begin{equation}
  \label{td(0)theta}
  \theta_{t+1} = \theta_{t}+\alpha_t\delta_t\bm{\phi}_t,
  \end{equation}
  where $\alpha_t\in (0,1)$ is a learning stepsize,
  and   $\delta_t=r_{t+1}+\gamma V_{\bm{\theta}}(s_{t+1})-V_{\bm{\theta}}(s_t)$.
  
 In off-policy learning, the target policy $\pi$ and the behavior policy $\mu$ are
 different. The update rule of the off-policy TD learning \citep{sutton2016emphatic} is
 \begin{equation}
  \bm{\theta}_{t+1} = \bm{\theta}_{t}+\alpha_t\rho_t\delta_t\bm{\phi}_t,
  \end{equation}
  where $\rho_t=\frac{\pi(a_t|s_t)}{\mu(a_t|s_t)}$.

\subsection{Reward Centering}
In on-policy learning, \citet{naik2024reward} 
proposed simple reward centering.
The update rule is
\begin{equation}
V_{t+1}(s_t)=V_{t}(s_t)+\alpha_t \bar{\delta}_t,
\label{src1}
\end{equation}
where the new TD error $\bar{\delta}_t$ is
\begin{equation}
\bar{\delta}_t=\delta_t-\bar{r}_{t} = r_{t+1}-\bar{r}_{t}+\gamma V_{t}(s_{t+1})-V_t(s_t),
\label{src2}
\end{equation}
and $\bar{r}_{t}$ is updated as:
\begin{equation}
\bar{r}_{t+1}=\bar{r}_{t}+\beta_t (r_{t+1}-\bar{r}_{t}),
\label{src3}
\end{equation}
where $\beta_t\in (0,1)$ is a learning stepsize.

In off-policy learning, \citet{naik2024reward} 
proposed value-based reward centering.
The update rule is
\begin{equation}
  \label{rewardcentering1}
V_{t+1}(s_t)=V_{t}(s_t)+\alpha_t \rho_t \bar{\delta}_t,
\end{equation}
where $\bar{r}_{t}$ is updated as:
\begin{equation}
  \label{rewardcentering2}
\bar{r}_{t+1}=\bar{r}_{t}+\beta_t \rho_t\bar{\delta}_t.
\end{equation}

\section{Bellman Error Centering}

Centering operator $\mathcal{C}$ for a variable $x(s)$ is defined as follows:
\begin{equation}
\mathcal{C}x(s)\dot{=} x(s)-\mathbb{E}[x(s)]=x(s)-\sum_s{d_{s}x(s)},
\end{equation} 
where $d_s$ is the probability of $s$.
In vector form,
\begin{equation}
\begin{split}
\mathcal{C}\bm{x} &= \bm{x}-\mathbb{E}[x]\bm{1}\\
&=\bm{x}-\bm{x}^{\top}\bm{d}\bm{1},
\end{split}
\end{equation} 
where $\bm{1}$ is an all-ones vector.
For any vector $\bm{x}$ and $\bm{y}$ with a same distribution $\bm{d}$,
we have
\begin{equation}
\begin{split}
\mathcal{C}(\bm{x}+\bm{y})&=(\bm{x}+\bm{y})-(\bm{x}+\bm{y})^{\top}\bm{d}\bm{1}\\
&=\bm{x}-\bm{x}^{\top}\bm{d}\bm{1}+\bm{y}-\bm{y}^{\top}\bm{d}\bm{1}\\
&=\mathcal{C}\bm{x}+\mathcal{C}\bm{y}.
\end{split}
\end{equation}
\subsection{Revisit Reward Centering}

The update (\ref{src3}) is an unbiased estimate of the average reward
with  appropriate learning rate $\beta_t$ conditions.
\begin{equation}
\bar{r}_{t}\approx \lim_{n\rightarrow\infty}\frac{1}{n}\sum_{t=1}^n\mathbb{E}_{\pi}[r_t].
\end{equation}
That is 
\begin{equation}
r_t-\bar{r}_{t}\approx r_t-\lim_{n\rightarrow\infty}\frac{1}{n}\sum_{t=1}^n\mathbb{E}_{\pi}[r_t]= \mathcal{C}r_t.
\end{equation}
Then, the simple reward centering can be rewrited as:
\begin{equation}
V_{t+1}(s_t)=V_{t}(s_t)+\alpha_t [\mathcal{C}r_{t+1}+\gamma V_{t}(s_{t+1})-V_t(s_t)].
\end{equation}
Therefore, the simple reward centering is, in a strict sense, reward centering.

By definition of $\bar{\delta}_t=\delta_t-\bar{r}_{t}$,
let rewrite the update rule of the value-based reward centering as follows:
\begin{equation}
V_{t+1}(s_t)=V_{t}(s_t)+\alpha_t \rho_t (\delta_t-\bar{r}_{t}),
\end{equation}
where $\bar{r}_{t}$ is updated as:
\begin{equation}
\bar{r}_{t+1}=\bar{r}_{t}+\beta_t \rho_t(\delta_t-\bar{r}_{t}).
\label{vrc3}
\end{equation}
The update (\ref{vrc3}) is an unbiased estimate of the TD error
with  appropriate learning rate $\beta_t$ conditions.
\begin{equation}
\bar{r}_{t}\approx \mathbb{E}_{\pi}[\delta_t].
\end{equation}
That is 
\begin{equation}
\delta_t-\bar{r}_{t}\approx \mathcal{C}\delta_t.
\end{equation}
Then, the value-based reward centering can be rewrited as:
\begin{equation}
V_{t+1}(s_t)=V_{t}(s_t)+\alpha_t \rho_t \mathcal{C}\delta_t.
\label{tdcentering}
\end{equation}
Therefore, the value-based reward centering is no more,
 in a strict sense, reward centering.
It is, in a strict sense, \textbf{Bellman error centering}.

It is worth noting that this understanding is crucial, 
as designing new algorithms requires leveraging this concept.

\subsection{On the Fixpoint Solution}

The update rule (\ref{tdcentering}) is a stochastic approximation
of the following update:
\begin{equation}
\begin{split}
V_{t+1}&=V_{t}+\alpha_t [\bm{\mathcal{T}}^{\pi}\bm{V}-\bm{V}-\mathbb{E}[\delta]\bm{1}]\\
&=V_{t}+\alpha_t [\bm{\mathcal{T}}^{\pi}\bm{V}-\bm{V}-(\bm{\mathcal{T}}^{\pi}\bm{V}-\bm{V})^{\top}\bm{d}_{\pi}\bm{1}]\\
&=V_{t}+\alpha_t [\mathcal{C}(\bm{\mathcal{T}}^{\pi}\bm{V}-\bm{V})].
\end{split}
\label{tdcenteringVector}
\end{equation}
If update rule (\ref{tdcenteringVector}) converges, it is expected that
$\mathcal{C}(\mathcal{T}^{\pi}V-V)=\bm{0}$.
That is 
\begin{equation}
    \begin{split}
    \mathcal{C}\bm{V} &= \mathcal{C}\bm{\mathcal{T}}^{\pi}\bm{V} \\
    &= \mathcal{C}(\bm{R}^{\pi} + \gamma \mathbb{P}^{\pi} \bm{V}) \\
    &= \mathcal{C}\bm{R}^{\pi} + \gamma \mathcal{C}\mathbb{P}^{\pi} \bm{V} \\
    &= \mathcal{C}\bm{R}^{\pi} + \gamma (\mathbb{P}^{\pi} \bm{V} - (\mathbb{P}^{\pi} \bm{V})^{\top} \bm{d_{\pi}} \bm{1}) \\
    &= \mathcal{C}\bm{R}^{\pi} + \gamma (\mathbb{P}^{\pi} \bm{V} - \bm{V}^{\top} (\mathbb{P}^{\pi})^{\top} \bm{d_{\pi}} \bm{1}) \\  
    &= \mathcal{C}\bm{R}^{\pi} + \gamma (\mathbb{P}^{\pi} \bm{V} - \bm{V}^{\top} \bm{d_{\pi}} \bm{1}) \\
    &= \mathcal{C}\bm{R}^{\pi} + \gamma (\mathbb{P}^{\pi} \bm{V} - \bm{V}^{\top} \bm{d_{\pi}} \mathbb{P}^{\pi} \bm{1}) \\
    &= \mathcal{C}\bm{R}^{\pi} + \gamma (\mathbb{P}^{\pi} \bm{V} - \mathbb{P}^{\pi} \bm{V}^{\top} \bm{d_{\pi}} \bm{1}) \\
    &= \mathcal{C}\bm{R}^{\pi} + \gamma \mathbb{P}^{\pi} (\bm{V} - \bm{V}^{\top} \bm{d_{\pi}} \bm{1}) \\
    &= \mathcal{C}\bm{R}^{\pi} + \gamma \mathbb{P}^{\pi} \mathcal{C}\bm{V} \\
    &\dot{=} \bm{\mathcal{T}}_c^{\pi} \mathcal{C}\bm{V},
    \end{split}
    \label{centeredfixpoint}
    \end{equation}
where we defined $\bm{\mathcal{T}}_c^{\pi}$ as a centered Bellman operator.
We call equation (\ref{centeredfixpoint}) as centered Bellman equation.
And it is \textbf{centered fixpoint}.

For linear value function approximation, let define
\begin{equation}
\mathcal{C}\bm{V}_{\bm{\theta}}=\bm{\Pi}\bm{\mathcal{T}}_c^{\pi}\mathcal{C}\bm{V}_{\bm{\theta}}.
\label{centeredTDfixpoint}
\end{equation}
We call equation (\ref{centeredTDfixpoint}) as \textbf{centered TD fixpoint}.

\subsection{On-policy and Off-policy Centered TD Algorithms
with Linear Value Function Approximation}
Given the above centered TD fixpoint,
 mean squared centered Bellman error (MSCBE), is proposed as follows:
\begin{align*}
    \label{argminMSBEC}
 &\arg \min_{{\bm{\theta}}}\text{MSCBE}({\bm{\theta}}) \\
 &= \arg \min_{{\bm{\theta}}} \|\bm{\mathcal{T}}_c^{\pi}\mathcal{C}\bm{V}_{\bm{{\bm{\theta}}}}-\mathcal{C}\bm{V}_{\bm{{\bm{\theta}}}}\|_{\bm{D}}^2\notag\\
 &=\arg \min_{{\bm{\theta}}} \|\bm{\mathcal{T}}^{\pi}\bm{V}_{\bm{{\bm{\theta}}}} - \bm{V}_{\bm{{\bm{\theta}}}}-(\bm{\mathcal{T}}^{\pi}\bm{V}_{\bm{{\bm{\theta}}}} - \bm{V}_{\bm{{\bm{\theta}}}})^{\top}\bm{d}\bm{1}\|_{\bm{D}}^2\notag\\
 &=\arg \min_{{\bm{\theta}},\omega} \| \bm{\mathcal{T}}^{\pi}\bm{V}_{\bm{{\bm{\theta}}}} - \bm{V}_{\bm{{\bm{\theta}}}}-\omega\bm{1} \|_{\bm{D}}^2\notag,
\end{align*}
where $\omega$ is is used to estimate the expected value of the Bellman error.

First, the parameter  $\omega$ is derived directly based on
stochastic gradient descent:
\begin{equation}
\omega_{t+1}= \omega_{t}+\beta_t(\delta_t-\omega_t).
\label{omega}
\end{equation}

Then, based on stochastic semi-gradient descent, the update of 
the parameter ${\bm{\theta}}$ is as follows:
\begin{equation}
{\bm{\theta}}_{t+1}=
{\bm{\theta}}_{t}+\alpha_t(\delta_t-\omega_t)\bm{{\bm{\phi}}}_t.
\label{theta}
\end{equation}

We call (\ref{omega}) and (\ref{theta}) the on-policy centered
TD (CTD) algorithm. The convergence analysis with be given in
the following section.

In off-policy learning, we can simply multiply by the importance sampling
 $\rho$.
\begin{equation}
    \omega_{t+1}=\omega_{t}+\beta_t\rho_t(\delta_t-\omega_t),
    \label{omegawithrho}
\end{equation}
\begin{equation}
    {\bm{\theta}}_{t+1}=
    {\bm{\theta}}_{t}+\alpha_t\rho_t(\delta_t-\omega_t)\bm{{\bm{\phi}}}_t.
    \label{thetawithrho}
\end{equation}

We call (\ref{omegawithrho}) and (\ref{thetawithrho}) the off-policy centered
TD (CTD) algorithm.



\subsection{Off-policy Centered TDC Algorithm with Linear Value Function Approximation}
The convergence of the  off-policy centered TD algorithm
may not be guaranteed.

To deal with this problem, we propose another new objective function, 
called mean squared projected centered Bellman error (MSPCBE), 
and derive Centered TDC algorithm (CTDC).


The specific expression of the objective function 
MSPCBE is as follows:
\begin{align}
    \label{MSPBECwithomega}
    &\arg \min_{{\bm{\theta}}}\text{MSPCBE}({\bm{\theta}})\notag\\ 
    &= \arg \min_{{\bm{\theta}}} \|\bm{\Pi}\bm{\mathcal{T}}_c^{\pi}\mathcal{C}\bm{V}_{\bm{{\bm{\theta}}}}-\mathcal{C}\bm{V}_{\bm{{\bm{\theta}}}}\|_{\bm{D}}^2\notag\\
    &= \arg \min_{{\bm{\theta}}} \|\bm{\Pi}(\bm{\mathcal{T}}_c^{\pi}\mathcal{C}\bm{V}_{\bm{{\bm{\theta}}}}-\mathcal{C}\bm{V}_{\bm{{\bm{\theta}}}})\|_{\bm{D}}^2\notag\\
    &= \arg \min_{{\bm{\theta}},\omega}\| {\bm{\Pi}} (\bm{\mathcal{T}}^{\pi}\bm{V}_{\bm{{\bm{\theta}}}} - \bm{V}_{\bm{{\bm{\theta}}}}-\omega\bm{1}) \|_{\bm{D}}^2\notag.
\end{align}
In the process of computing the gradient of the MSPCBE with respect to ${\bm{\theta}}$, 
$\omega$ is treated as a constant.
So, the derivation process of CTDC is the same 
as for the TDC algorithm \cite{sutton2009fast}, the only difference is that the original $\delta$ is replaced by $\delta-\omega$.
Therefore, the updated formulas of the centered TDC  algorithm are as follows:
\begin{equation}
 \bm{{\bm{\theta}}}_{k+1}=\bm{{\bm{\theta}}}_{k}+\alpha_{k}[(\delta_{k}- \omega_k) \bm{\bm{{\bm{\phi}}}}_k\\
 - \gamma\bm{\bm{{\bm{\phi}}}}_{k+1}(\bm{\bm{{\bm{\phi}}}}^{\top}_k \bm{u}_{k})],
\label{thetavmtdc}
\end{equation}
\begin{equation}
 \bm{u}_{k+1}= \bm{u}_{k}+\zeta_{k}[\delta_{k}-\omega_k - \bm{\bm{{\bm{\phi}}}}^{\top}_k \bm{u}_{k}]\bm{\bm{{\bm{\phi}}}}_k,
\label{uvmtdc}
\end{equation}
and
\begin{equation}
 \omega_{k+1}= \omega_{k}+\beta_k (\delta_k- \omega_k).
 \label{omegavmtdc}
\end{equation}
This algorithm is derived to work 
with a given set of sub-samples—in the form of 
triples $(S_k, R_k, S'_k)$ that match transitions 
from both the behavior and target policies.

\section{Proof of Convergence of On-policy CTD}
The purpose of this section
is to establish that the  on-policy CTD algorithm converges with probability one
to the centered TD fixpoint
 under standard assumptions.
\begin{theorem}
    \label{theorem1}(Convergence of on-policy CTD).
    In the case of on-policy learning, consider the iterations (\ref{omega}) and (\ref{theta}).
    Let the step-size sequences $\alpha_k$ and $\beta_k$, $k\geq 0$ satisfy in this case $\alpha_k,\beta_k>0$, for all $k$,
    $
    \sum_{k=0}^{\infty}\alpha_k=\sum_{k=0}^{\infty}\beta_k=\infty,
    $
    $
    \sum_{k=0}^{\infty}\alpha_k^2<\infty,
    $
    $
    \sum_{k=0}^{\infty}\beta_k^2<\infty,
    $
    and  
    $
    \alpha_k = o(\beta_k).
    $
    Assume that $(\bm{\bm{\phi}}_k,r_k,\bm{\bm{\phi}}_k')$ is an i.i.d. sequence with
    uniformly bounded second moments, where $\bm{\bm{\phi}}_k$ and $\bm{\bm{\phi}}'_{k}$ are sampled from the same Markov chain.
    Let $\textbf{A} = \mathrm{Cov}(\bm{\bm{\phi}},\bm{\bm{\phi}}-\gamma\bm{\bm{\phi}}')$,
    $\bm{b}=\mathrm{Cov}(r,\bm{\bm{\phi}})$.
    Assume that matrix $\textbf{A}$ is non-singular. 
    Then the parameter vector $\bm{\bm{\theta}}_k$ converges with probability one 
    to the centered TD fixpoint $\textbf{A}^{-1}\bm{b}$ (\ref{centeredTDfixpoint}).
   \end{theorem}
\begin{proof}
\label{th1proof}   
    The proof is  based on Borkar's Theorem   for
    general stochastic approximation recursions with two time scales
    \cite{borkar1997stochastic}. 
    
    The centered one-step
    linear TD solution is defined
    as: 
    \begin{equation*}
    0=\mathbb{E}[(\delta-\mathbb{E}[\delta]) \bm{\bm{\phi}}]=-\textbf{A}\bm{\bm{\theta}}+\bm{b}.
    \end{equation*}
    Thus, the on-policy CTD algorithm's solution is
    $\bm{\bm{\theta}}_{\text{CTD}}=\textbf{A}^{-1}\bm{b}$.
    
    First, note that recursion (\ref{theta}) can be rewritten as
    \begin{equation*}
    \bm{\bm{\theta}}_{k+1}\leftarrow \bm{\bm{\theta}}_k+\beta_k\xi(k),
    \end{equation*}
    where
    \begin{equation*}
    \xi(k)=\frac{\alpha_k}{\beta_k}(\delta_k-\omega_k)\bm{\bm{\phi}}_k.
    \end{equation*}
    Due to the settings of step-size schedule $\alpha_k = o(\beta_k)$,
    $\xi(k)\rightarrow 0$ almost surely as $k\rightarrow\infty$. 
    That is the increments in iteration (\ref{omega}) are uniformly larger than
    those in (\ref{theta}), thus (\ref{omega}) is the faster recursion.
    Along the faster time scale, iterations of (\ref{omega}) and (\ref{theta})
    are associated to ODEs system as follows:
    \begin{equation}
    \dot{\bm{\bm{\theta}}}(t) = 0,
    \label{thetaFast}
    \end{equation}
    \begin{equation}
    \dot{\omega}(t)=\mathbb{E}[\delta_t|\bm{\bm{\theta}}(t)]-\omega(t).
    \label{omegaFast}
    \end{equation}
    Based on the ODE (\ref{thetaFast}), $\bm{\bm{\theta}}(t)\equiv \bm{\bm{\theta}}$ when
    viewed from the faster timescale. 
    By the Hirsch lemma \cite{hirsch1989convergent}, it follows that
    $||\bm{\bm{\theta}}_k-\bm{\bm{\theta}}||\rightarrow 0$ a.s. as $k\rightarrow \infty$ for some
    $\bm{\bm{\theta}}$ that depends on the initial condition $\bm{\bm{\theta}}_0$ of recursion
    (\ref{theta}).
    Thus, the ODE pair (\ref{thetaFast})-(\ref{omegaFast}) can be written as
    \begin{equation}
    \dot{\omega}(t)=\mathbb{E}[\delta_t|\bm{\bm{\theta}}]-\omega(t).
    \label{omegaFastFinal}
    \end{equation}
    Consider the function $h(\omega)=\mathbb{E}[\delta|\bm{\bm{\theta}}]-\omega$,
    i.e., the driving vector field of the ODE (\ref{omegaFastFinal}).
    It is easy to find that the function $h$ is Lipschitz with coefficient
    $-1$.
    Let $h_{\infty}(\cdot)$ be the function defined by
     $h_{\infty}(\omega)=\lim_{x\rightarrow \infty}\frac{h(x\omega)}{x}$.
     Then  $h_{\infty}(\omega)= -\omega$,  is well-defined. 
     For (\ref{omegaFastFinal}), $\omega^*=\mathbb{E}[\delta|\bm{\bm{\theta}}]$
    is the unique globally asymptotically stable equilibrium.
     For the ODE
      \begin{equation}
     \dot{\omega}(t) = h_{\infty}(\omega(t))= -\omega(t),
     \label{omegaInfty}
     \end{equation}
     apply $\vec{V}(\omega)=(-\omega)^{\top}(-\omega)/2$ as its
    associated strict Liapunov function. Then,
    the origin of (\ref{omegaInfty}) is a globally asymptotically stable
    equilibrium.

    Consider now the recursion (\ref{omega}).
    Let
    $M_{k+1}=(\delta_k-\omega_k)
    -\mathbb{E}[(\delta_k-\omega_k)|\mathcal{F}(k)]$,
    where $\mathcal{F}(k)=\sigma(\omega_l,\bm{\bm{\theta}}_l,l\leq k;\bm{\bm{\phi}}_s,\bm{\bm{\phi}}_s',r_s,s<k)$, 
    $k\geq 1$ are the sigma fields
    generated by $\omega_0,\bm{\bm{\theta}}_0,\omega_{l+1},\bm{\bm{\theta}}_{l+1},\bm{\bm{\phi}}_l,\bm{\bm{\phi}}_l'$,
    $0\leq l<k$.
    It is easy to verify that $M_{k+1},k\geq0$ are integrable random variables that 
    satisfy $\mathbb{E}[M_{k+1}|\mathcal{F}(k)]=0$, $\forall k\geq0$.
    Because $\bm{\bm{\phi}}_k$, $r_k$, and $\bm{\bm{\phi}}_k'$   have
    uniformly bounded second moments, it can be seen that for some constant
    $c_1>0$, $\forall k\geq0$,
    \begin{equation*}
    \mathbb{E}[||M_{k+1}||^2|\mathcal{F}(k)]\leq
    c_1(1+||\omega_k||^2+||\bm{\bm{\theta}}_k||^2).
    \end{equation*}

    Now Assumptions (A1) and (A2) of \cite{borkar2000ode} are verified.
    Furthermore, Assumptions (TS) of \cite{borkar2000ode} is satisfied by our
    conditions on the step-size sequences $\alpha_k$, $\beta_k$. Thus,
    by Theorem 2.2 of \cite{borkar2000ode} we obtain that
    $||\omega_k-\omega^*||\rightarrow 0$ almost surely as $k\rightarrow \infty$.
    
    Consider now the slower time scale recursion (\ref{theta}).
    Based on the above analysis, (\ref{theta}) can be rewritten as 
    \begin{equation*}
    \bm{\bm{\theta}}_{k+1}\leftarrow
    \bm{\bm{\theta}}_{k}+\alpha_k(\delta_k-\mathbb{E}[\delta_k|\bm{\bm{\theta}}_k])\bm{\bm{\phi}}_k.
    \end{equation*}
    
    Let $\mathcal{G}(k)=\sigma(\bm{\bm{\theta}}_l,l\leq k;\bm{\bm{\phi}}_s,\bm{\bm{\phi}}_s',r_s,s<k)$, 
    $k\geq 1$ be the sigma fields
    generated by $\bm{\bm{\theta}}_0,\bm{\bm{\theta}}_{l+1},\bm{\bm{\phi}}_l,\bm{\bm{\phi}}_l'$,
    $0\leq l<k$.
    Let 
    $
    Z_{k+1} = Y_{t}-\mathbb{E}[Y_{t}|\mathcal{G}(k)],
    $
    where
    \begin{equation*}
    Y_{k}=(\delta_k-\mathbb{E}[\delta_k|\bm{\bm{\theta}}_k])\bm{\bm{\phi}}_k.
    \end{equation*}
    Consequently,
    \begin{equation*}
    \begin{array}{ccl}
    \mathbb{E}[Y_t|\mathcal{G}(k)]&=&\mathbb{E}[(\delta_k-\mathbb{E}[\delta_k|\bm{\bm{\theta}}_k])\bm{\bm{\phi}}_k|\mathcal{G}(k)]\\
    &=&\mathbb{E}[\delta_k\bm{\bm{\phi}}_k|\bm{\bm{\theta}}_k]
    -\mathbb{E}[\mathbb{E}[\delta_k|\bm{\bm{\theta}}_k]\bm{\bm{\phi}}_k]\\
    &=&\mathbb{E}[\delta_k\bm{\bm{\phi}}_k|\bm{\bm{\theta}}_k]
    -\mathbb{E}[\delta_k|\bm{\bm{\theta}}_k]\mathbb{E}[\bm{\bm{\phi}}_k]\\
    &=&\mathrm{Cov}(\delta_k|\bm{\bm{\theta}}_k,\bm{\bm{\phi}}_k),
    \end{array}
    \end{equation*}
    where $\mathrm{Cov}(\cdot,\cdot)$ is a covariance operator.
    
     Thus,
     \begin{equation*}
    \begin{array}{ccl}
    Z_{k+1}&=&(\delta_k-\mathbb{E}[\delta_k|\bm{\bm{\theta}}_k])\bm{\bm{\phi}}_k-\mathrm{Cov}(\delta_k|\bm{\bm{\theta}}_k,\bm{\bm{\phi}}_k).
    \end{array}
    \end{equation*}
    It is easy to verify that $Z_{k+1},k\geq0$ are integrable random variables that 
    satisfy $\mathbb{E}[Z_{k+1}|\mathcal{G}(k)]=0$, $\forall k\geq0$.
    Also, because $\bm{\bm{\phi}}_k$, $r_k$, and $\bm{\bm{\phi}}_k'$  have
    uniformly bounded second moments, it can be seen that for some constant
    $c_2>0$, $\forall k\geq0$,
    \begin{equation*}
    \mathbb{E}[||Z_{k+1}||^2|\mathcal{G}(k)]\leq
    c_2(1+||\bm{\bm{\theta}}_k||^2).
    \end{equation*}
    
    Consider now the following ODE associated with (\ref{theta}):
    \begin{equation}
    \begin{array}{ccl}
    \dot{\bm{\bm{\theta}}}(t)&=&\mathrm{Cov}(\delta|\bm{\bm{\theta}}(t),\bm{\bm{\phi}})\\
    &=&\mathrm{Cov}(r+(\gamma\bm{\bm{\phi}}'-\bm{\bm{\phi}})^{\top}\bm{\bm{\theta}}(t),\bm{\bm{\phi}})\\
    &=&\mathrm{Cov}(r,\bm{\bm{\phi}})-\mathrm{Cov}(\bm{\bm{\theta}}(t)^{\top}(\bm{\bm{\phi}}-\gamma\bm{\bm{\phi}}'),\bm{\bm{\phi}})\\
    &=&\mathrm{Cov}(r,\bm{\bm{\phi}})-\bm{\bm{\theta}}(t)^{\top}\mathrm{Cov}(\bm{\bm{\phi}}-\gamma\bm{\bm{\phi}}',\bm{\bm{\phi}})\\
    &=&\mathrm{Cov}(r,\bm{\bm{\phi}})-\mathrm{Cov}(\bm{\bm{\phi}}-\gamma\bm{\bm{\phi}}',\bm{\bm{\phi}})^{\top}\bm{\bm{\theta}}(t)\\
    &=&\mathrm{Cov}(r,\bm{\bm{\phi}})-\mathrm{Cov}(\bm{\bm{\phi}},\bm{\bm{\phi}}-\gamma\bm{\bm{\phi}}')\bm{\bm{\theta}}(t)\\
    &=&-\textbf{A}\bm{\bm{\theta}}(t)+\bm{b}.
    \end{array}
    \label{odetheta}
    \end{equation}
    Let $\vec{h}(\bm{\bm{\theta}}(t))$ be the driving vector field of the ODE
    (\ref{odetheta}).
    \begin{equation*}
    \vec{h}(\bm{\bm{\theta}}(t))=-\textbf{A}\bm{\bm{\theta}}(t)+\bm{b}.
    \end{equation*}
     Consider the cross-covariance matrix,
    \begin{equation}
    \begin{array}{ccl}
        \textbf{A} &=& \mathrm{Cov}(\bm{\bm{\phi}},\bm{\bm{\phi}}-\gamma\bm{\bm{\phi}}')\\
      &=&\frac{\mathrm{Cov}(\bm{\bm{\phi}},\bm{\bm{\phi}})+\mathrm{Cov}(\bm{\bm{\phi}}-\gamma\bm{\bm{\phi}}',\bm{\bm{\phi}}-\gamma\bm{\bm{\phi}}')-\mathrm{Cov}(\gamma\bm{\bm{\phi}}',\gamma\bm{\bm{\phi}}')}{2}\\
      &=&\frac{\mathrm{Cov}(\bm{\bm{\phi}},\bm{\bm{\phi}})+\mathrm{Cov}(\bm{\bm{\phi}}-\gamma\bm{\bm{\phi}}',\bm{\bm{\phi}}-\gamma\bm{\bm{\phi}}')-\gamma^2\mathrm{Cov}(\bm{\bm{\phi}}',\bm{\bm{\phi}}')}{2}\\
      &=&\frac{(1-\gamma^2)\mathrm{Cov}(\bm{\bm{\phi}},\bm{\bm{\phi}})+\mathrm{Cov}(\bm{\bm{\phi}}-\gamma\bm{\bm{\phi}}',\bm{\bm{\phi}}-\gamma\bm{\bm{\phi}}')}{2},\\
    \end{array}
    \label{covariance}
    \end{equation}
    where we eventually used $\mathrm{Cov}(\bm{\bm{\phi}}',\bm{\bm{\phi}}')=\mathrm{Cov}(\bm{\bm{\phi}},\bm{\bm{\phi}})$
    \footnote{The covariance matrix $\mathrm{Cov}(\bm{\bm{\phi}}',\bm{\bm{\phi}}')$ is equal to
    the covariance matrix $\mathrm{Cov}(\bm{\bm{\phi}},\bm{\bm{\phi}})$ if the initial state is re-reachable or
    initialized randomly in a Markov chain for on-policy update.}.
    Note that the covariance matrix $\mathrm{Cov}(\bm{\bm{\phi}},\bm{\bm{\phi}})$ and
    $\mathrm{Cov}(\bm{\bm{\phi}}-\gamma\bm{\bm{\phi}}',\bm{\bm{\phi}}-\gamma\bm{\bm{\phi}}')$ are semi-positive
    definite. Then, the matrix $\textbf{A}$ is semi-positive definite because  $\textbf{A}$ is
    linearly combined  by  two positive-weighted semi-positive definite matrice
    (\ref{covariance}).
    Furthermore, $\textbf{A}$ is nonsingular due to the assumption.
    Hence, the cross-covariance matrix $\textbf{A}$ is positive definite.
    
    Therefore,
    $\bm{\bm{\theta}}^*=\textbf{A}^{-1}\bm{b}$ can be seen to be the unique globally asymptotically
    stable equilibrium for ODE (\ref{odetheta}).
    Let $\vec{h}_{\infty}(\bm{\bm{\theta}})=\lim_{r\rightarrow
    \infty}\frac{\vec{h}(r\bm{\bm{\theta}})}{r}$. Then
    $\vec{h}_{\infty}(\bm{\bm{\theta}})=-\textbf{A}\bm{\bm{\theta}}$ is well-defined. 
    Consider now the ODE
    \begin{equation}
    \dot{\bm{\bm{\theta}}}(t)=-\textbf{A}\bm{\bm{\theta}}(t).
    \label{odethetafinal}
    \end{equation}
    The ODE (\ref{odethetafinal}) has the origin as its unique globally asymptotically stable equilibrium.
    Thus, the assumption (A1) and (A2) are verified.
\end{proof}

\section{Proof of Convergence of Off-policy CTDC}
The purpose of this section
is to establish that the  off-policy CTDC algorithm converges with probability one
to the centered TD fixpoint
 under standard assumptions.

\begin{theorem}
 \label{theorem2}(Convergence of off-policy CTDC).
 In the case of off-policy learning, consider the iterations (\ref{omegavmtdc}), (\ref{uvmtdc}) and (\ref{thetavmtdc}).
 Let the step-size sequences $\alpha_k$, $\zeta_k$ and $\beta_k$, $k\geq 0$ satisfy in this case $\alpha_k,\zeta_k,\beta_k>0$, for all $k$,
 $
 \sum_{k=0}^{\infty}\alpha_k=\sum_{k=0}^{\infty}\beta_k=\sum_{k=0}^{\infty}\zeta_k=\infty,
 $
 $
 \sum_{k=0}^{\infty}\alpha_k^2<\infty,
 $
 $
 \sum_{k=0}^{\infty}\zeta_k^2<\infty,
 $
 $
 \sum_{k=0}^{\infty}\beta_k^2<\infty,
 $
 and  
 $
 \alpha_k = o(\zeta_k),
 $
 $
 \zeta_k = o(\beta_k).
 $
 Assume that $(\bm{\bm{\phi}}_k,r_k,\bm{\bm{\phi}}_k')$ is an i.i.d. sequence with
 uniformly bounded second moments.
 Let $\textbf{A} = \mathrm{Cov}(\bm{\bm{\phi}},\bm{\bm{\phi}}-\gamma\bm{\bm{\phi}}')$,
 $\bm{b}=\mathrm{Cov}(r,\bm{\bm{\phi}})$, and $\textbf{C}=\mathbb{E}[\bm{\bm{\phi}}\bm{\bm{\phi}}^{\top}]$.
 Assume that  $\textbf{A}$ and $\textbf{C}$ are non-singular matrices. 
 Then the parameter vector $\bm{\bm{\theta}}_k$ converges with probability one 
 to the centered TD fixpoint $\textbf{A}^{-1}\bm{b}$ (\ref{centeredTDfixpoint}).
\end{theorem}
\begin{proof}
    The proof is  based on multi-time-scale stochastic approximation that is    
    similar to that given by \cite{sutton2009fast} for TDC. 
    
    For the off-policy CTDC algorithm, the centered  one-step linear TD solution is defined as:
    \begin{equation*}
        \begin{array}{ccl}
            0 &=& \mathbb{E}[({\bm{\phi}} - \gamma {\bm{\phi}}' - \mathbb{E}[{\bm{\phi}} - \gamma {\bm{\phi}}']){\bm{\phi}}^\top]\mathbb{E}[{\bm{\phi}} {\bm{\phi}}^{\top}]^{-1}\\
            &&\mathbb{E}[(\delta -\mathbb{E}[\delta]){\bm{\phi}}]\\
          &=&\textbf{A}^{\top}\textbf{C}^{-1}(-\textbf{A}{\bm{\theta}}+{\bm{b}}).
         \end{array}
    \end{equation*}
    The matrix $\textbf{A}^{\top}\textbf{C}^{-1}\textbf{A}$ is positive definite. Thus,   the off-policy CTDC  algorithm's solution is
    ${\bm{\theta}}_{\text{CTDC}}=\textbf{A}^{-1}{b}$.
    
    First, note that recursion (\ref{thetavmtdc}) and (\ref{uvmtdc}) can be rewritten as, respectively, 
    \begin{equation*}
     {\bm{\theta}}_{k+1}\leftarrow {\bm{\theta}}_k+\zeta_k {x}(k),
    \end{equation*}
    \begin{equation*}
     {u}_{k+1}\leftarrow {u}_k+\beta_k {y}(k),
    \end{equation*}
    where 
    \begin{equation*}
     {x}(k)=\frac{\alpha_k}{\zeta_k}[(\delta_{k}- \omega_k) {\bm{\phi}}_k - \gamma{\bm{\phi}}'_{k}({\bm{\phi}}^{\top}_k {u}_k)],
    \end{equation*}
    \begin{equation*}
     {y}(k)=\frac{\zeta_k}{\beta_k}[\delta_{k}-\omega_k - {\bm{\phi}}^{\top}_k {u}_k]{\bm{\phi}}_k.
    \end{equation*}
    
    Recursion (\ref{thetavmtdc}) can also be rewritten as
    \begin{equation*}
     {\bm{\theta}}_{k+1}\leftarrow {\bm{\theta}}_k+\beta_k z(k),
    \end{equation*}
    where
    \begin{equation*}
     z(k)=\frac{\alpha_k}{\beta_k}[(\delta_{k}- \omega_k) {\bm{\phi}}_k - \gamma{\bm{\phi}}'_{k}({\bm{\phi}}^{\top}_k {u}_k)].
    \end{equation*}
    
    Due to the settings of the step-size schedule 
    $\alpha_k = o(\zeta_k)$, $\zeta_k = o(\beta_k)$, ${x}(k)\rightarrow 0$, ${y}(k)\rightarrow 0$, $z(k)\rightarrow 0$ almost surely as $k\rightarrow 0$.
    That is the increments in iteration (\ref{omegavmtdc}) are uniformly larger than
    those in (\ref{uvmtdc}) and  the increments in iteration (\ref{uvmtdc}) are uniformly larger than
    those in (\ref{thetavmtdc}), thus (\ref{omegavmtdc}) is the fastest recursion, (\ref{uvmtdc}) is the second fast recursion, and (\ref{thetavmtdc}) is the slower recursion.
    Along the fastest time scale, iterations of (\ref{thetavmtdc}), (\ref{uvmtdc}) and (\ref{omegavmtdc})
    are associated with the ODEs system as follows:
    \begin{equation}
     \dot{{\bm{\theta}}}(t) = 0,
        \label{thetavmtdcFastest}
    \end{equation}
    \begin{equation}
     \dot{{u}}(t) = 0,
        \label{uvmtdcFastest}
    \end{equation}
    \begin{equation}
     \dot{\omega}(t)=\mathbb{E}[\delta_t|{u}(t),{\bm{\theta}}(t)]-\omega(t).
        \label{omegavmtdcFastest}
    \end{equation}
    
    Based on the ODE (\ref{thetavmtdcFastest}) and (\ref{uvmtdcFastest}), both ${\bm{\theta}}(t)\equiv {\bm{\theta}}$
    and ${u}(t)\equiv {u}$ when viewed from the fastest timescale.
    By the Hirsch lemma \cite{hirsch1989convergent}, it follows that
    $||{\bm{\theta}}_k-{\bm{\theta}}||\rightarrow 0$ a.s. as $k\rightarrow \infty$ for some
    ${\bm{\theta}}$ that depends on the initial condition ${\bm{\theta}}_0$ of recursion
    (\ref{thetavmtdc}) and $||{u}_k-{u}||\rightarrow 0$ a.s. as $k\rightarrow \infty$ for some
    $u$ that depends on the initial condition $u_0$ of recursion
    (\ref{uvmtdc}). Thus, the ODE pair (\ref{thetavmtdcFastest})-(\ref{omegavmtdcFastest})
    can be written as 
    \begin{equation}
     \dot{\omega}(t)=\mathbb{E}[\delta_t|{u},{\bm{\theta}}]-\omega(t).
        \label{omegavmtdcFastestFinal}
    \end{equation}
    
    Consider the function $h(\omega)=\mathbb{E}[\delta|{\bm{\theta}},{u}]-\omega$,
    i.e., the driving vector field of the ODE (\ref{omegavmtdcFastestFinal}).
    It is easy to find that the function $h$ is Lipschitz with coefficient
    $-1$.
    Let $h_{\infty}(\cdot)$ be the function defined by
     $h_{\infty}(\omega)=\lim_{r\rightarrow \infty}\frac{h(r\omega)}{r}$.
     Then  $h_{\infty}(\omega)= -\omega$,  is well-defined. 
     For (\ref{omegavmtdcFastestFinal}), $\omega^*=\mathbb{E}[\delta|{\bm{\theta}},{u}]$
    is the unique globally asymptotically stable equilibrium.
    For the ODE
    \begin{equation}
     \dot{\omega}(t) = h_{\infty}(\omega(t))= -\omega(t),
     \label{omegavmtdcInfty}
    \end{equation}
    apply $\vec{V}(\omega)=(-\omega)^{\top}(-\omega)/2$ as its
    associated strict Liapunov function. Then,
    the origin of (\ref{omegavmtdcInfty}) is a globally asymptotically stable
    equilibrium.
    
    Consider now the recursion (\ref{omegavmtdc}).
    Let
    $M_{k+1}=(\delta_k-\omega_k)
    -\mathbb{E}[(\delta_k-\omega_k)|\mathcal{F}(k)]$,
    where $\mathcal{F}(k)=\sigma(\omega_l,{u}_l,{\bm{\theta}}_l,l\leq k;{\bm{\phi}}_s,{\bm{\phi}}_s',r_s,s<k)$, 
    $k\geq 1$ are the sigma fields
    generated by $\omega_0,u_0,{\bm{\theta}}_0,\omega_{l+1},{u}_{l+1},{\bm{\theta}}_{l+1},{\bm{\phi}}_l,{\bm{\phi}}_l'$,
    $0\leq l<k$.
    It is easy to verify that $M_{k+1},k\geq0$ are integrable random variables that 
    satisfy $\mathbb{E}[M_{k+1}|\mathcal{F}(k)]=0$, $\forall k\geq0$.
    Because ${\bm{\phi}}_k$, $r_k$, and ${\bm{\phi}}_k'$ have
    uniformly bounded second moments, it can be seen that for some constant
    $c_1>0$, $\forall k\geq0$,
    \begin{equation*}
    \mathbb{E}[||M_{k+1}||^2|\mathcal{F}(k)]\leq
    c_1(1+||\omega_k||^2+||{u}_k||^2+||{\bm{\theta}}_k||^2).
    \end{equation*}

    Now Assumptions (A1) and (A2) of \cite{borkar2000ode} are verified.
    Furthermore, Assumptions (TS) of \cite{borkar2000ode} is satisfied by our
    conditions on the step-size sequences $\alpha_k$,$\zeta_k$, $\beta_k$. Thus,
    by Theorem 2.2 of \cite{borkar2000ode} we obtain that
    $||\omega_k-\omega^*||\rightarrow 0$ almost surely as $k\rightarrow \infty$.
    
    Consider now the second time scale recursion (\ref{uvmtdc}).
    Based on the above analysis, (\ref{uvmtdc}) can be rewritten as
    \begin{equation}
     \dot{{\bm{\theta}}}(t) = 0,
        \label{thetavmtdcFaster}
    \end{equation}
    \begin{equation}
     \dot{u}(t) = \mathbb{E}[(\delta_t-\mathbb{E}[\delta_t|{u}(t),{\bm{\theta}}(t)]){\bm{\phi}}_t|{\bm{\theta}}(t)] - \textbf{C}{u}(t).
        \label{uvmtdcFaster}
    \end{equation}
    The ODE (\ref{thetavmtdcFaster}) suggests that ${\bm{\theta}}(t)\equiv {\bm{\theta}}$ (i.e., a time-invariant parameter)
    when viewed from the second fast timescale.
    By the Hirsch lemma \cite{hirsch1989convergent}, it follows that
    $||{\bm{\theta}}_k-{\bm{\theta}}||\rightarrow 0$ a.s. as $k\rightarrow \infty$ for some
    ${\bm{\theta}}$ that depends on the initial condition ${\bm{\theta}}_0$ of recursion
    (\ref{thetavmtdc}). 
    
    Consider now the recursion (\ref{uvmtdc}).
    Let
    $N_{k+1}=((\delta_k-\mathbb{E}[\delta_k]) - {\bm{\phi}}_k {\bm{\phi}}^{\top}_k {u}_k) -\mathbb{E}[((\delta_k-\mathbb{E}[\delta_k]) - {\bm{\phi}}_k {\bm{\phi}}^{\top}_k {u}_k)|\mathcal{I} (k)]$,
    where $\mathcal{I}(k)=\sigma({u}_l,{\bm{\theta}}_l,l\leq k;{\bm{\phi}}_s,{\bm{\phi}}_s',r_s,s<k)$, 
    $k\geq 1$ are the sigma fields
    generated by ${u}_0,{\bm{\theta}}_0,{u}_{l+1},{\bm{\theta}}_{l+1},{\bm{\phi}}_l,{\bm{\phi}}_l'$,
    $0\leq l<k$.
    It is easy to verify that $N_{k+1},k\geq0$ are integrable random variables that 
    satisfy $\mathbb{E}[N_{k+1}|\mathcal{I}(k)]=0$, $\forall k\geq0$.
    Because ${\bm{\phi}}_k$, $r_k$, and ${\bm{\phi}}_k'$ have
    uniformly bounded second moments, it can be seen that for some constant
    $c_2>0$, $\forall k\geq0$,
    \begin{equation*}
    \mathbb{E}[||N_{k+1}||^2|\mathcal{I}(k)]\leq
    c_2(1+||{u}_k||^2+||{\bm{\theta}}_k||^2).
    \end{equation*}
    
    Because ${\bm{\theta}}(t)\equiv {\bm{\theta}}$ from (\ref{thetavmtdcFaster}), the ODE pair (\ref{thetavmtdcFaster})-(\ref{uvmtdcFaster})
    can be written as 
    \begin{equation}
     \dot{{u}}(t) = \mathbb{E}[(\delta_t-\mathbb{E}[\delta_t|{\bm{\theta}}]){\bm{\phi}}_t|{\bm{\theta}}] - \textbf{C}{u}(t).
        \label{uvmtdcFasterFinal}
    \end{equation}
    Now consider the function $h({u})=\mathbb{E}[\delta_t-\mathbb{E}[\delta_t|{\bm{\theta}}]|{\bm{\theta}}] -\textbf{C}{u}$, i.e., the
    driving vector field of the ODE (\ref{uvmtdcFasterFinal}). For (\ref{uvmtdcFasterFinal}),
    ${u}^* = \textbf{C}^{-1}\mathbb{E}[(\delta-\mathbb{E}[\delta|{\bm{\theta}}]){\bm{\phi}}|{\bm{\theta}}]$ is the unique globally asymptotically
    stable equilibrium. Let $h_{\infty}({u})=-\textbf{C}{u}$.
    For the ODE
    \begin{equation}
     \dot{{u}}(t) = h_{\infty}({u}(t))= -\textbf{C}{u}(t),
        \label{uvmtdcInfty}
    \end{equation}
    the origin of (\ref{uvmtdcInfty}) is a globally asymptotically stable
    equilibrium because $\textbf{C}$ is a positive definite matrix (because it is nonnegative definite and nonsingular).
    Now Assumptions (A1) and (A2) of \cite{borkar2000ode} are verified.
    Furthermore, Assumptions (TS) of \cite{borkar2000ode} is satisfied by our
    conditions on the step-size sequences $\alpha_k$,$\zeta_k$, $\beta_k$. Thus,
    by Theorem 2.2 of \cite{borkar2000ode} we obtain that
    $||{u}_k-{u}^*||\rightarrow 0$ almost surely as $k\rightarrow \infty$.
    
    Consider now the slower timescale recursion (\ref{thetavmtdc}). In the light of the above,
    (\ref{thetavmtdc}) can be rewritten as 
    \begin{equation*}
        \begin{array}{ccl}
            {\bm{\theta}}_{k+1} &\leftarrow& {\bm{\theta}}_{k} + \alpha_k (\delta_k -\mathbb{E}[\delta_k|{\bm{\theta}}_k]) {\bm{\phi}}_k\\
            &&- \alpha_k \gamma{\bm{\phi}}'_{k}({\bm{\phi}}^{\top}_k \textbf{C}^{-1}\mathbb{E}[(\delta_k -\mathbb{E}[\delta_k|{\bm{\theta}}_k]){\bm{\phi}}|{\bm{\theta}}_k]).
         \end{array}
    \end{equation*}
    Let $\mathcal{G}(k)=\sigma({\bm{\theta}}_l,l\leq k;{\bm{\phi}}_s,{\bm{\phi}}_s',r_s,s<k)$, 
    $k\geq 1$ be the sigma fields
    generated by ${\bm{\theta}}_0,{\bm{\theta}}_{l+1},{\bm{\phi}}_l,{\bm{\phi}}_l'$,
    $0\leq l<k$. Let
    \begin{equation*}
        \begin{array}{ccl}
     Z_{k+1}&=&\big((\delta_k -\mathbb{E}[\delta_k|{\bm{\theta}}_k]) {\bm{\phi}}_k \\
            &&- \gamma {\bm{\phi}}'_{k}{\bm{\phi}}^{\top}_k \textbf{C}^{-1}\mathbb{E}[(\delta_k -\mathbb{E}[\delta_k|{\bm{\theta}}_k]){\bm{\phi}}|{\bm{\theta}}_k]\big)\\ 
         & &-\big(\mathbb{E}[((\delta_k -\mathbb{E}[\delta_k|{\bm{\theta}}_k]) {\bm{\phi}}_k \\
        &&- \gamma {\bm{\phi}}'_{k}{\bm{\phi}}^{\top}_k \textbf{C}^{-1}\mathbb{E}[(\delta_k -\mathbb{E}[\delta_k|{\bm{\theta}}_k]){\bm{\phi}}|{\bm{\theta}}_k])|\mathcal{G}(k)]\big)\\
        &=&\big((\delta_k -\mathbb{E}[\delta_k|{\bm{\theta}}_k]) {\bm{\phi}}_k \\
    &&-\gamma {\bm{\phi}}'_{k}{\bm{\phi}}^{\top}_k \textbf{C}^{-1}\mathbb{E}[(\delta_k -\mathbb{E}[\delta_k|{\bm{\theta}}_k]){\bm{\phi}}|{\bm{\theta}}_k]\big)\\
        & &-\big(\mathbb{E}[(\delta_k -\mathbb{E}[\delta_k|{\bm{\theta}}_k]) {\bm{\phi}}_k|{\bm{\theta}}_k]\\
    &&- \gamma\mathbb{E}[{\bm{\phi}}' {\bm{\phi}}^{\top}]\textbf{C}^{-1}\mathbb{E}[(\delta_k -\mathbb{E}[\delta_k|{\bm{\theta}}_k]) {\bm{\phi}}_k|{\bm{\theta}}_k]\big).
        \end{array}
    \end{equation*}
    It is easy to see that $Z_k$, $k\geq 0$ are integrable random variables and $\mathbb{E}[Z_{k+1}|\mathcal{G}(k)]=0$, $\forall k\geq0$. Further,
    \begin{equation*}
    \mathbb{E}[||Z_{k+1}||^2|\mathcal{G}(k)]\leq
    c_3(1+||{\bm{\theta}}_k||^2), k\geq 0,
    \end{equation*}
    for some constant $c_3 \geq 0$, again because ${\bm{\phi}}_k$, $r_k$, and ${\bm{\phi}}_k'$ have
    uniformly bounded second moments, it can be seen that for some constant.
    
    Consider now the following ODE associated with (\ref{thetavmtdc}):
    \begin{equation}
        \label{thetavmtdcSlowerFinal}
        \begin{array}{ccl}
            \dot{{\bm{\theta}}}(t) &=& \mathbb{E}[({\bm{\phi}} - \gamma {\bm{\phi}}' - \mathbb{E}[{\bm{\phi}} - \gamma {\bm{\phi}}']){\bm{\phi}}^\top]\mathbb{E}[{\bm{\phi}} {\bm{\phi}}^{\top}]^{-1}\\
            &&\quad \quad \quad\mathbb{E}[(\delta -\mathbb{E}[\delta|{\bm{\theta}}(t)]) {\bm{\phi}}|{\bm{\theta}}(t)].
         \end{array}
    \end{equation}
    Let 
    \begin{equation*}
    \begin{array}{ccl}
     \vec{h}({\bm{\theta}}(t))&=&\mathbb{E}[({\bm{\phi}} - \gamma {\bm{\phi}}' - \mathbb{E}[{\bm{\phi}} - \gamma {\bm{\phi}}']){\bm{\phi}}^\top]\mathbb{E}[{\bm{\phi}} {\bm{\phi}}^{\top}]^{-1}\\
     &&\quad \quad \quad\mathbb{E}[(\delta -\mathbb{E}[\delta|{\bm{\theta}}(t)]) {\bm{\phi}}|{\bm{\theta}}(t)]\\
        &=& \textbf{A}^{\top}\textbf{C}^{-1}(-\textbf{A}{\bm{\theta}}(t)+{\bm{b}}),
    \end{array}
    \end{equation*}
    because $\mathbb{E}[(\delta -\mathbb{E}[\delta|{\bm{\theta}}(t)]) {\bm{\phi}}|{\bm{\theta}}(t)]=-\textbf{A}{\bm{\theta}}(t)+{b}$, where 
    $\textbf{A} = \mathrm{Cov}({\bm{\phi}},{\bm{\phi}}-\gamma{\bm{\phi}}')$, ${\bm{b}}=\mathrm{Cov}(r,{\bm{\phi}})$, and $\textbf{C}=\mathbb{E}[{\bm{\phi}}{\bm{\phi}}^{\top}]$.
    
    Therefore,
    ${\bm{\theta}}^*=\textbf{A}^{-1}{\bm{b}}$ can be seen to be the unique globally asymptotically
    stable equilibrium for ODE (\ref{thetavmtdcSlowerFinal}).
    Let $\vec{h}_{\infty}({\bm{\theta}})=\lim_{r\rightarrow
    \infty}\frac{\vec{h}(r{\bm{\theta}})}{r}$. Then
    $\vec{h}_{\infty}({\bm{\theta}})=-\textbf{A}^{\top}\textbf{C}^{-1}\textbf{A}{\bm{\theta}}$ is well-defined. 
    Consider now the ODE
    \begin{equation}
    \dot{{\bm{\theta}}}(t)=-\textbf{A}^{\top}\textbf{C}^{-1}\textbf{A}{\bm{\theta}}(t).
    \label{odethetavmtdcfinal}
    \end{equation}
    
    Because $\textbf{C}^{-1}$ is positive definite and $\textbf{A}$ has full rank (as it
    is nonsingular by assumption), the matrix $\textbf{A}^{\top} \textbf{C}^{-1}\textbf{A}$ is also
    positive definite. 
    The ODE (\ref{odethetavmtdcfinal}) has the origin of its unique globally asymptotically stable equilibrium.
    Thus, the assumption (A1) and (A2) are verified.
    
    The proof is given above.
    In the fastest time scale, the parameter $w$ converges to
    $\mathbb{E}[\delta|{u}_k,{\bm{\theta}}_k]$.
    In the second fast time scale,
    the parameter $u$ converges to $\textbf{C}^{-1}\mathbb{E}[(\delta-\mathbb{E}[\delta|{\bm{\theta}}_k]){\bm{\phi}}|{\bm{\theta}}_k]$.
    In the slower time scale,
    the parameter ${\bm{\theta}}$ converges to $\textbf{A}^{-1}{\bm{b}}$.
    \end{proof}

\section{Experimental Studies}

This section aims to validates the convergence of the algorithms
in the policy evaluation experiments.
We employ  three policy evaluation experiments, 
including {Boyanchian}, {2-state off-policy counterexample} 
and {7-state off-policy counterexample}.
Details are shown in the appendix.
We compare the performance of 
the proposed CTD and CTDC algorithms with  the TD and TDC algorithms.

The vertical axis is unified as root mean squared centered Bellman error (RMSCBE).
Each experiment was trained for 50 iterations, with each 
episode of BoyanChain consisting of 1,000 steps and each 
episode of the other two experiments consisting of 2,000 
steps. We show the shaded regions  as the standard 
deviation (std).

\begin{figure}[H]
    \vskip 0.2in
    \begin{center}
    \subfigure[Boyanchian]{
        \includegraphics[width=\columnwidth, height=0.68\columnwidth]{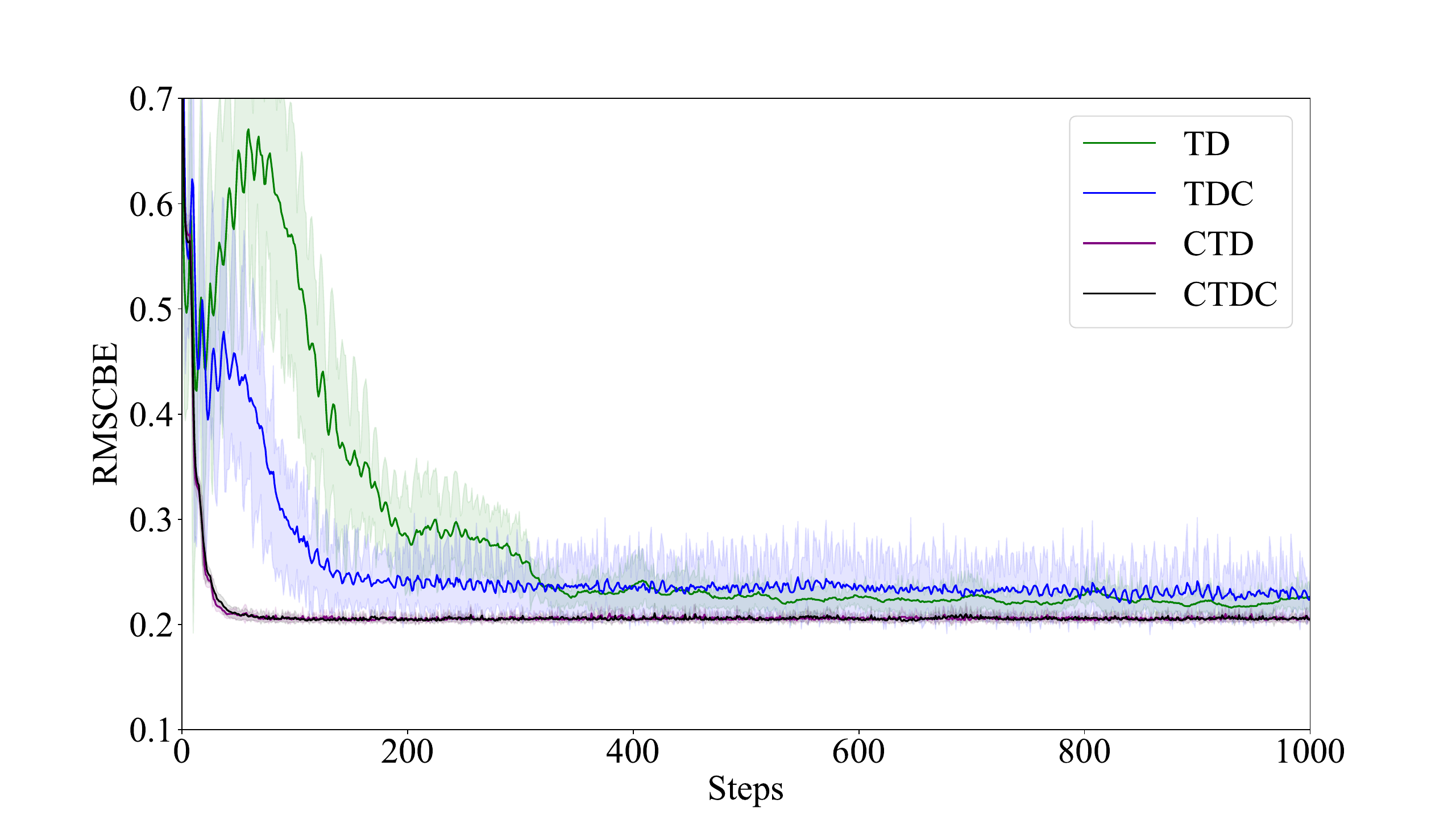}
        \label{Boyanchian}
    }
    \subfigure[2-state counterexample]{
        \includegraphics[width=\columnwidth, height=0.68\columnwidth]{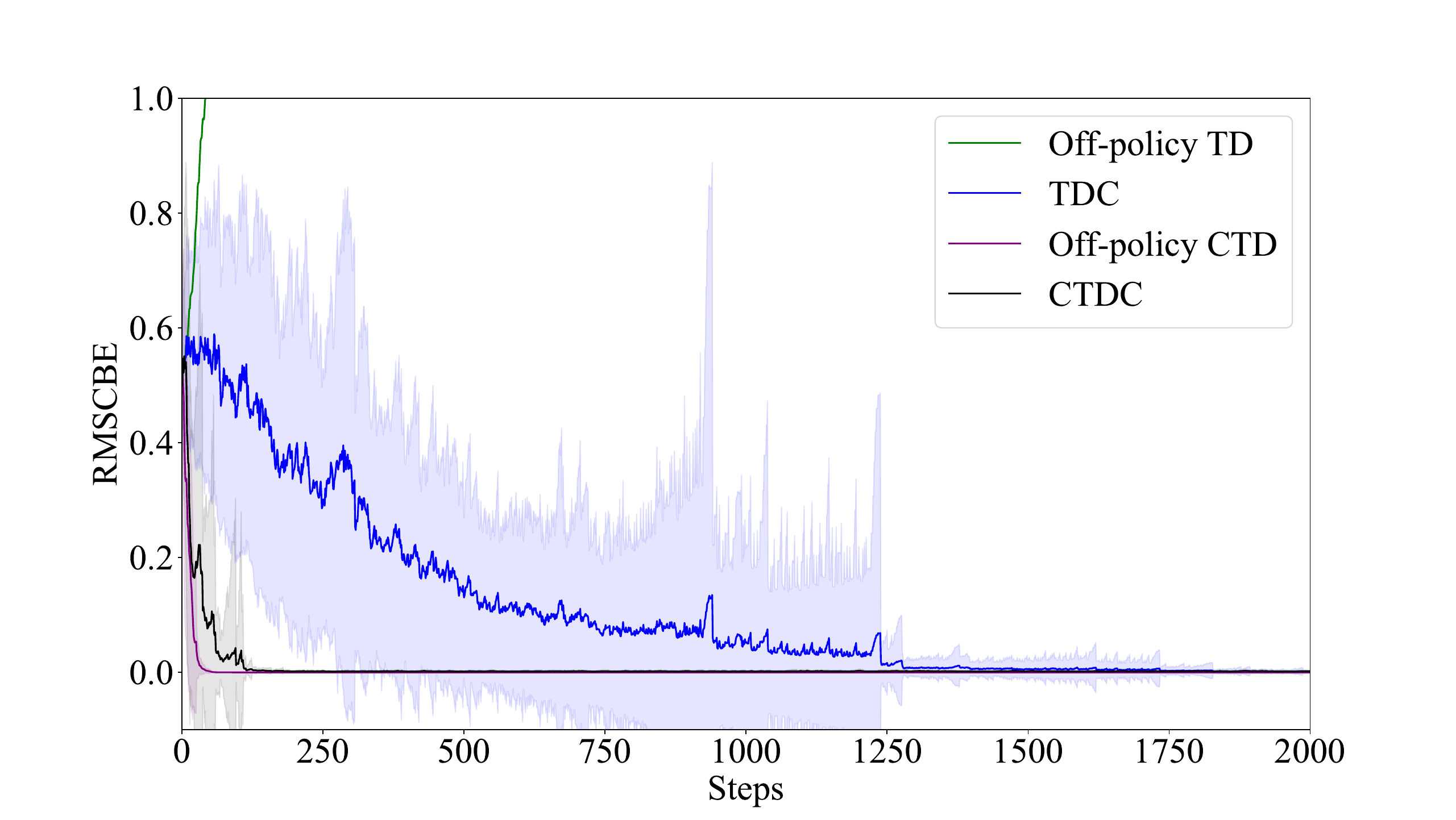}
        \label{2statestep}
    }
    \subfigure[7-state counterexample]{
        \includegraphics[width=\columnwidth, height=0.68\columnwidth]{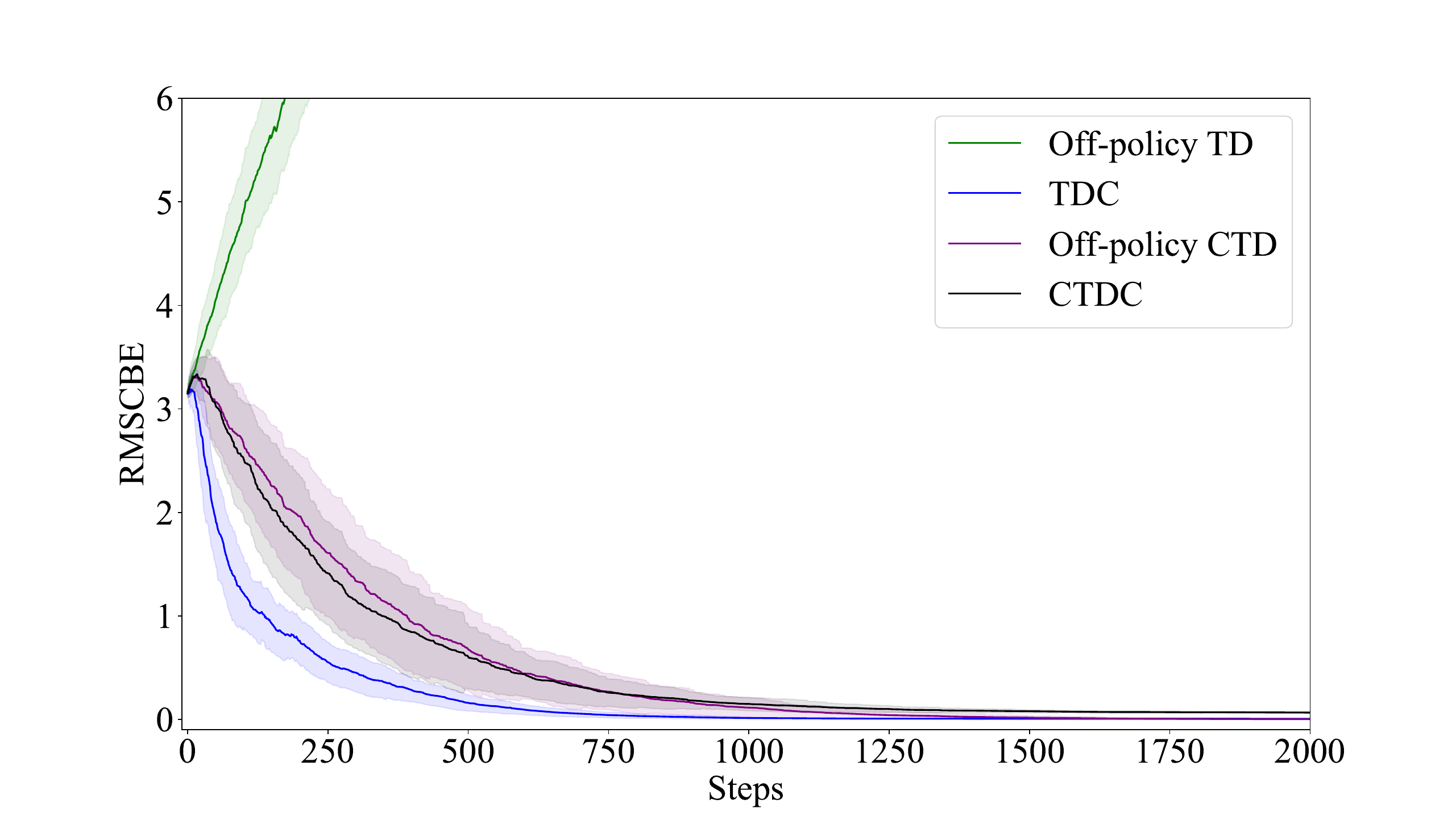}
        \label{7statestep}
    }
        \caption{Learning curses of three evaluation environments.}
        \label{learningcurvesofbairdexample}
    \end{center}
    \vskip -0.2in
\end{figure}

Figure \ref{Boyanchian} shows the learning curses in Boyanchain.
This environment is an on-policy setting. 
All algorithms converge, with the CTD and CTDC 
algorithms exhibiting the fastest convergence and 
the best performance. The convergence of the CTD algorithm 
validates the theoretical proof presented earlier regarding 
its stable convergence under on-policy conditions.

Figure \ref{2statestep} and \ref{7statestep} 
show the learning curses in 2-state counterexample and 
7-state counterexample, respectively.
Both environments are off-policy settings. 
Off-policy TD diverges, while the other 
three algorithms converge.  
The convergence of the CTDC algorithm validates 
the theoretical proof presented earlier regarding 
its stable convergence under off-policy conditions.
In the 2-state counterexample environment, both the
off-policy CTD algorithm and the off-policy CTDC algorithm
 perform exceptionally well.

Figure \ref{2statestep} and \ref{7statestep} show that
 the off-policy TD algorithm diverges 
 whereas off-policy CTD converge stably. This result is 
particularly surprising because we have only proven that CTD converges 
stably in the on-policy setting.

Comparing Equation (\ref{theta}) and Equation (\ref{td(0)theta}), we find that $\omega$ in 
(\ref{theta}) is essentially an adjustment or correction of the TD update. We think 
the presence  of $\omega$ leads to stabilizing gradient estimation and reducing the variance
of gradient estimation.

One may wonder why the off-policy CTD algorithm converges in the 2-state counterexample.
 We provide the following analysis.
\begin{lemma}
    Consider a general two-state off-policy counterexample with state features  
    $
    \bm{\Phi} = (m, n)^{\top}$, where $m\neq 0$, $n\neq 0$, and $m \neq n$.
    The behavior policy is given by  
    $
    \textbf{P}_{\mu}=
    \begin{bmatrix}
    a & 1-a \\
    b & 1-b
    \end{bmatrix},
    $
    where \( 0 \leqslant a \leqslant 1 \) and \( 0 \leqslant b \leqslant 1 \).  
    The target policy is defined as  
    $
    \textbf{P}_{\pi}=
    \begin{bmatrix}
    x & 1-x \\
    y & 1-y
    \end{bmatrix},
    $
    where \( 0 \leqslant x \leqslant 1 \) and \( 0 \leqslant y \leqslant 1 \).  
    In this setting, $\textbf{A}_{\textbf{off-policy CTD}}$ is  positive definite.
\end{lemma}
\begin{proof}
    \label{th3proof}   

The stationary distribution $\bm{d}_{\mu}$ and 
$\textbf{P}_{\mu}$ satisfy the  invariance $\bm{d}_\mu^{\top}\mathbf{P}_\mu = \bm{d}_\mu^{\top}$. 
Then, 
$\bm{d}_\mu =(\frac{b}{1-a+b},\frac{1-a}{1-a+b})^{\top}$.

Given that 
$\textbf{A}_{\textbf{off-policy CTD}}=\bm{\Phi^{\top}}(\textbf{D}-\bm{d}_{\mu}\bm{d}_{\mu}^{\top})(\bm{I}-\gamma\textbf{P}_{\pi})\bm{\Phi}$.
\begin{equation*}
    \textbf{D}-\bm{d}_{\mu}\bm{d}_{\mu}^{\top}=\frac{(1-a)b}{(1-a+b)^{2}}\begin{bmatrix} 1 & -1 \\ -1 & 1 \end{bmatrix},
\end{equation*}
\begin{equation*}
    (\textbf{D}-\bm{d}_{\mu}\bm{d}_{\mu}^{\top})(\bm{I}-\gamma\textbf{P}_{\pi})=\frac{(1-a)b}{(1-a+b)^{2}}(1-x\gamma +y\gamma)\begin{bmatrix} 1 & -1 \\ -1 & 1 \end{bmatrix},
\end{equation*}
where, 
\begin{equation*}
    \textbf{A}_{\textbf{off-policy CTD}}=\frac{(1-a)b}{(1-a+b)^{2}}(1-x\gamma +y\gamma)(m-n)^2.
\end{equation*}
Since $m\neq n$, $\textbf{A}_{\textbf{off-policy CTD}}$ is a positive definite matrix.
\end{proof}
This is the reason why the off-policy CTD algorithm converges
 in the 2-state counterexample.

\section{Conclusion and Future Work}
This paper explores the principles of reward centering, interpreting value-based
reward centering as Bellman error centering. We derive the fixed-point solutions
under tabular value and linear function approximation. Additionally, we present
both on-policy and off-policy algorithms, along with proofs of their
convergence.

Future work includes, but is not limited to,

(1) extensions of the CTD and CTDC algorithms to learning for control, 
especially the episodic problems;

(2) extending Bellman error centering to the $\lambda$ returns;

(3) extending Bellman error centering to the emphatic TD \cite{sutton2016emphatic}
and its convergence analysis;

(4) extending Bellman error centering to policy gradient and actor-critic RL algorithms.

\nocite{langley00}

\section{Impact Statement}
This paper presents work whose goal is to advance the field 
of Reinforcement Learning. There are many potential societal
 consequences of our work, none which we feel must be
specifically highlighted here.
\bibliography{example_paper}

\begin{thebibliography}{27}
\providecommand{\natexlab}[1]{#1}
\providecommand{\url}[1]{\texttt{#1}}
\expandafter\ifx\csname urlstyle\endcsname\relax
  \providecommand{\doi}[1]{doi: #1}\else
  \providecommand{\doi}{doi: \begingroup \urlstyle{rm}\Url}\fi

\bibitem[Abounadi et~al.(2001)Abounadi, Bertsekas, and
  Borkar]{abounadi2001learning}
Abounadi, J., Bertsekas, D., and Borkar, V.~S.
\newblock Learning algorithms for markov decision processes with average cost.
\newblock \emph{SIAM Journal on Control and Optimization}, 40\penalty0
  (3):\penalty0 681--698, 2001.

\bibitem[Baird et~al.(1995)]{baird1995residual}
Baird, L. et~al.
\newblock Residual algorithms: Reinforcement learning with function
  approximation.
\newblock In \emph{Proc. 12th Int. Conf. Mach. Learn.}, pp.\  30--37, 1995.

\bibitem[Borkar(1997)]{borkar1997stochastic}
Borkar, V.~S.
\newblock Stochastic approximation with two time scales.
\newblock \emph{Syst. \& Control Letters}, 29\penalty0 (5):\penalty0 291--294,
  1997.

\bibitem[Borkar \& Meyn(2000)Borkar and Meyn]{borkar2000ode}
Borkar, V.~S. and Meyn, S.~P.
\newblock The ode method for convergence of stochastic approximation and
  reinforcement learning.
\newblock \emph{SIAM J. Control Optim.}, 38\penalty0 (2):\penalty0 447--469,
  2000.

\bibitem[Carta et~al.(2023)Carta, Romac, Wolf, Lamprier, Sigaud, and
  Oudeyer]{carta2023grounding}
Carta, T., Romac, C., Wolf, T., Lamprier, S., Sigaud, O., and Oudeyer, P.-Y.
\newblock Grounding large language models in interactive environments with
  online reinforcement learning.
\newblock In \emph{International Conference on Machine Learning}, pp.\
  3676--3713. PMLR, 2023.

\bibitem[Dai et~al.(2024)Dai, Pan, Sun, Ji, Xu, Liu, Wang, and
  Yang]{dai2024safe}
Dai, J., Pan, X., Sun, R., Ji, J., Xu, X., Liu, M., Wang, Y., and Yang, Y.
\newblock Safe rlhf: Safe reinforcement learning from human feedback.
\newblock In \emph{The Twelfth International Conference on Learning
  Representations}, 2024.

\bibitem[Das et~al.(1999)Das, Gosavi, Mahadevan, and
  Marchalleck]{das1999solving}
Das, T.~K., Gosavi, A., Mahadevan, S., and Marchalleck, N.
\newblock Solving semi-markov decision problems using average reward
  reinforcement learning.
\newblock \emph{Management Science}, 45\penalty0 (4):\penalty0 560--574, 1999.

\bibitem[Grand-Cl{\'e}ment \& Petrik(2024)Grand-Cl{\'e}ment and
  Petrik]{grand2024reducing}
Grand-Cl{\'e}ment, J. and Petrik, M.
\newblock Reducing blackwell and average optimality to discounted mdps via the
  blackwell discount factor.
\newblock \emph{Advances in Neural Information Processing Systems}, pp.\
  52628--52647, 2024.

\bibitem[Guo et~al.(2025)Guo, Yang, Zhang, Song, Zhang, Xu, Zhu, Ma, Wang, Bi,
  et~al.]{guo2025deepseek}
Guo, D., Yang, D., Zhang, H., Song, J., Zhang, R., Xu, R., Zhu, Q., Ma, S.,
  Wang, P., Bi, X., et~al.
\newblock Deepseek-r1: Incentivizing reasoning capability in llms via
  reinforcement learning.
\newblock \emph{arXiv preprint arXiv:2501.12948}, 2025.

\bibitem[Hirsch(1989)]{hirsch1989convergent}
Hirsch, M.~W.
\newblock Convergent activation dynamics in continuous time networks.
\newblock \emph{Neural Netw.}, 2\penalty0 (5):\penalty0 331--349, 1989.

\bibitem[Langley(2000)]{langley00}
Langley, P.
\newblock Crafting papers on machine learning.
\newblock In Langley, P. (ed.), \emph{Proceedings of the 17th International
  Conference on Machine Learning (ICML 2000)}, pp.\  1207--1216, Stanford, CA,
  2000. Morgan Kaufmann.

\bibitem[Maei(2011)]{maei2011gradient}
Maei, H.~R.
\newblock \emph{Gradient temporal-difference learning algorithms}.
\newblock PhD thesis, University of Alberta, 2011.

\bibitem[Naik(2024)]{naik2024reinforcement}
Naik, A.
\newblock \emph{Reinforcement Learning for Continuing Problems Using Average
  Reward}.
\newblock PhD thesis, Department of Computing Science, University of Alberta,
  2024.

\bibitem[Naik et~al.(2024)Naik, Wan, Tomar, and Sutton]{naik2024reward}
Naik, A., Wan, Y., Tomar, M., and Sutton, R.
\newblock Reward centering.
\newblock In \emph{Proceedings of Reinforcement Learning Conference}, 2024.

\bibitem[Ouyang et~al.(2022)Ouyang, Wu, Jiang, Almeida, Wainwright, Mishkin,
  Zhang, Agarwal, Slama, Ray, et~al.]{ouyang2022training}
Ouyang, L., Wu, J., Jiang, X., Almeida, D., Wainwright, C., Mishkin, P., Zhang,
  C., Agarwal, S., Slama, K., Ray, A., et~al.
\newblock Training language models to follow instructions with human feedback.
\newblock \emph{Advances in neural information processing systems},
  35:\penalty0 27730--27744, 2022.

\bibitem[Patterson et~al.(2021)Patterson, Gonzalez, Le, Liang, Munguia,
  Rothchild, So, Texier, and Dean]{patterson2021carbon}
Patterson, D., Gonzalez, J., Le, Q., Liang, C., Munguia, L.-M., Rothchild, D.,
  So, D., Texier, M., and Dean, J.
\newblock Carbon emissions and large neural network training.
\newblock \emph{arXiv preprint arXiv:2104.10350}, 2021.

\bibitem[Perotto \& Vercouter(2018)Perotto and Vercouter]{perotto2018tuning}
Perotto, F.~S. and Vercouter, L.
\newblock Tuning the discount factor in order to reach average optimality on
  deterministic mdps.
\newblock In \emph{Artificial Intelligence XXXV: 38th SGAI International
  Conference on Artificial Intelligence, AI 2018, Cambridge, UK, December
  11--13, 2018, Proceedings 38}, pp.\  92--105. Springer, 2018.

\bibitem[Schneckenreither \& Moser(2025)Schneckenreither and
  Moser]{schneckenreither2025average}
Schneckenreither, M. and Moser, G.
\newblock Average reward adjusted discounted reinforcement learning.
\newblock \emph{Neural Computing and Applications}, pp.\  1--32, 2025.

\bibitem[Schwartz(1993)]{schwartz1993reinforcement}
Schwartz, A.
\newblock A reinforcement learning method for maximizing undiscounted rewards.
\newblock In \emph{Proceedings of the tenth international conference on machine
  learning}, volume 298, pp.\  298--305, 1993.

\bibitem[Silver et~al.(2016)Silver, Huang, Maddison, Guez, Sifre, van~den
  Driessche, Schrittwieser, Antonoglou, Panneershelvam, Lanctot,
  et~al.]{silver2016mastering}
Silver, D., Huang, A., Maddison, C.~J., Guez, A., Sifre, L., van~den Driessche,
  G., Schrittwieser, J., Antonoglou, I., Panneershelvam, V., Lanctot, M.,
  et~al.
\newblock Mastering the game of go with deep neural networks and tree search.
\newblock \emph{Nature}, 529\penalty0 (7587):\penalty0 484--489, 2016.

\bibitem[Sun et~al.(2022)Sun, Han, Yang, Ma, Guo, and Zhou]{sun2022exploit}
Sun, H., Han, L., Yang, R., Ma, X., Guo, J., and Zhou, B.
\newblock Exploit reward shifting in value-based deep-rl: Optimistic
  curiosity-based exploration and conservative exploitation via linear reward
  shaping.
\newblock \emph{Advances in neural information processing systems},
  35:\penalty0 37719--37734, 2022.

\bibitem[Sutton et~al.(2009)Sutton, Maei, Precup, Bhatnagar, Silver,
  Szepesv{\'a}ri, and Wiewiora]{sutton2009fast}
Sutton, R., Maei, H., Precup, D., Bhatnagar, S., Silver, D., Szepesv{\'a}ri,
  C., and Wiewiora, E.
\newblock Fast gradient-descent methods for temporal-difference learning with
  linear function approximation.
\newblock In \emph{Proc. 26th Int. Conf. Mach. Learn.}, pp.\  993--1000, 2009.

\bibitem[Sutton \& Barto(2018)Sutton and Barto]{Sutton2018book}
Sutton, R.~S. and Barto, A.~G.
\newblock \emph{Reinforcement Learning: An Introduction}.
\newblock The MIT Press, second edition, 2018.

\bibitem[Sutton et~al.(2008)Sutton, Maei, and
  Szepesv{\'a}ri]{sutton2008convergent}
Sutton, R.~S., Maei, H.~R., and Szepesv{\'a}ri, C.
\newblock A convergent $ o (n) $ temporal-difference algorithm for off-policy
  learning with linear function approximation.
\newblock In \emph{Advances in Neural Information Processing Systems}, pp.\
  1609--1616. Cambridge, MA: MIT Press, 2008.

\bibitem[Sutton et~al.(2016)Sutton, Mahmood, and White]{sutton2016emphatic}
Sutton, R.~S., Mahmood, A.~R., and White, M.
\newblock An emphatic approach to the problem of off-policy temporal-difference
  learning.
\newblock \emph{The Journal of Machine Learning Research}, 17\penalty0
  (1):\penalty0 2603--2631, 2016.

\bibitem[Wan et~al.(2021)Wan, Naik, and Sutton]{wan2021learning}
Wan, Y., Naik, A., and Sutton, R.~S.
\newblock Learning and planning in average-reward markov decision processes.
\newblock In \emph{International Conference on Machine Learning}, pp.\
  10653--10662. PMLR, 2021.

\bibitem[Yang et~al.(2016)Yang, Gao, An, Wang, and Chen]{yang2016efficient}
Yang, S., Gao, Y., An, B., Wang, H., and Chen, X.
\newblock Efficient average reward reinforcement learning using constant
  shifting values.
\newblock In \emph{Proceedings of the Thirtieth AAAI Conference on Artificial
  Intelligence}, pp.\  2258--2264, 2016.

\end{thebibliography}
\bibliographystyle{icml2025}

\newpage
\appendix
\onecolumn
\section{Experimental Details}
\subsection{Boychain}
\label{appendixboyanchian}
\textbf{Boyanchian:} BoyanChain is a classic on-policy environment 
with 13 states, each represented by a 4-dimensional feature vector. The agent 
can choose between two actions at each state: move to the next state (action 1) or skip to 
the one after (action 2). State 11 always transitions to state 12, while state 12 resets to 
state 0. Rewards are structured as follows: -2 for moving from state 11 to 12, 0 
for resetting from state 12 to 0, and -3 for all other actions. The policy is 
deterministic at state 11 (always choosing action 1) and stochastic elsewhere, 
randomly selecting between action 1 and 2. 
The discount factor $\gamma =0.9$. 
The feature matrix of Boyanchain is
defined as follows:
\begin{equation*}
    \bm{\Phi}_{Boyanchain}=\left[ 
    \begin{array}{cccc}
     1& 0& 0& 0\\
     0.75& 0.25& 0& 0\\
     0.5& 0.5& 0& 0\\
     0.25&0.75&0&0\\
     0& 1& 0& 0\\
     0& 0.75& 0.25& 0\\
    0 & 0.5& 0.5& 0\\
     0& 0.25& 0.75& 0\\
     0& 0& 1& 0\\
     0& 0& 0.75& 0.25\\
    0 & 0&0.5& 0.5\\
     0& 0& 0.25& 0.75\\
     0& 0& 0& 1
    \end{array}\right]
\end{equation*}
\begin{figure}[H]
    \begin{center}
    \label{boyanchianonpolicy}
    \includegraphics[width=0.5\columnwidth, height=0.2\columnwidth]{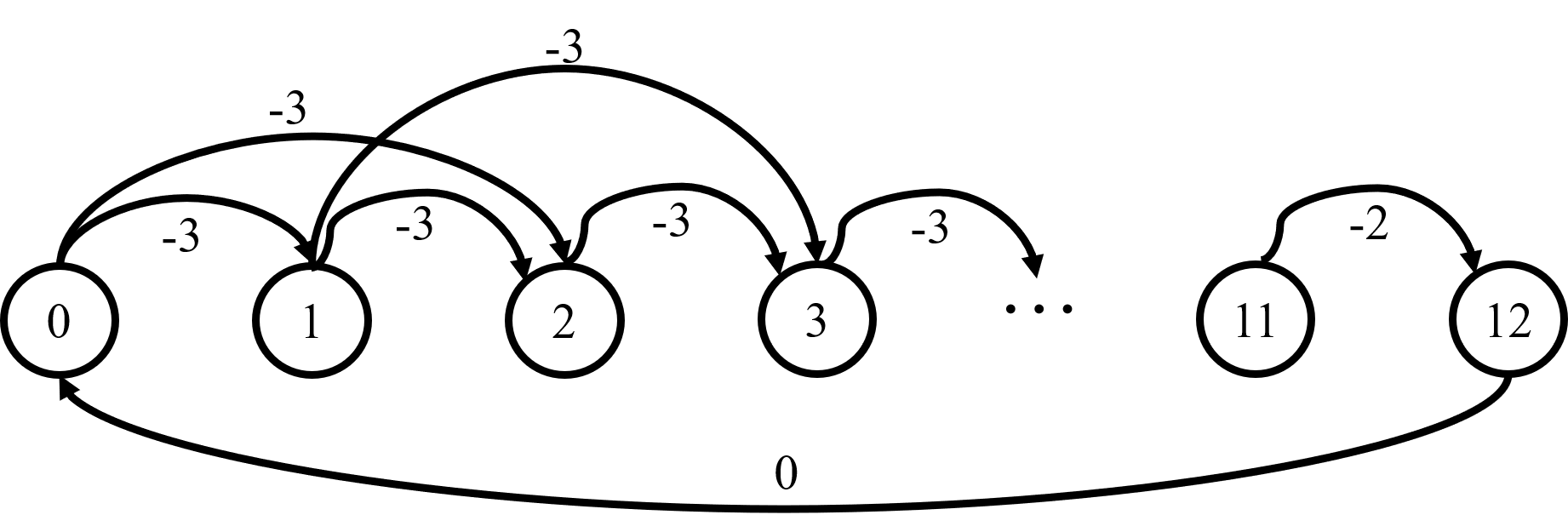}
        \caption{Boychain.}
    \end{center}
\end{figure}
The $\alpha$ values for all algorithms are in the range of $\{0.0001, 0.0005, 0.001, 0.005, 0.01, 0.05, 0.1, 0.2, 0.3\}$. 
For the TDC and CTDC algorithm, the $\zeta$ values are in the range of $\{0.0005, 0.001, 0.005, 0.01, 0.05, 0.1, 0.2, 0.5\}$. 
For the CTD and CTDC algorithm, the $\beta$ values are in the range of $\{0.0005, 0.001, 0.005, 0.01, 0.05, 0.1, 0.2, 0.5\}$. 
\begin{figure}
    \vskip 0.2in
    \begin{center}
    \subfigure[TD]{
        \includegraphics[width=0.3\columnwidth, height=0.25\columnwidth]{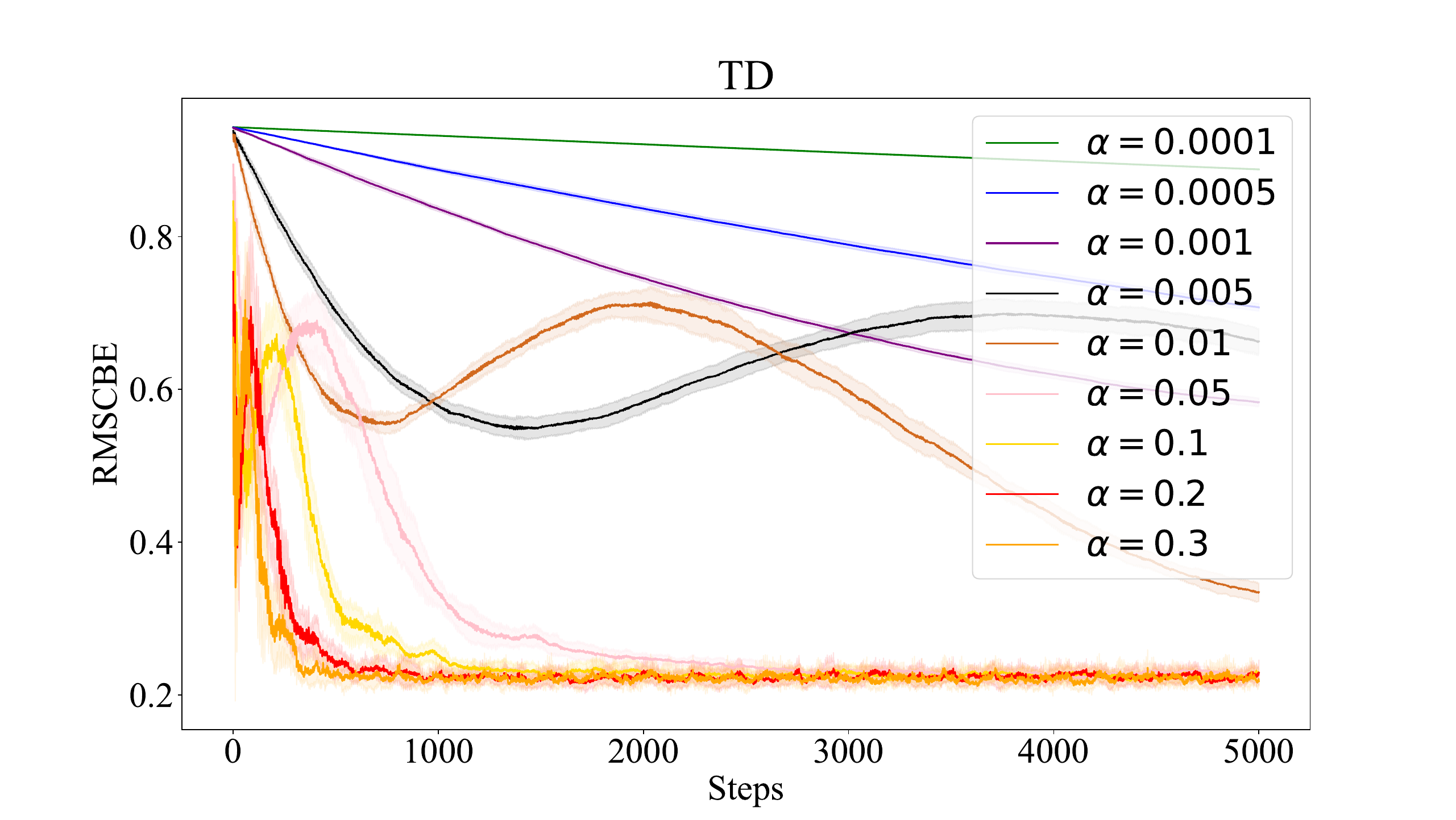}
        \label{td_boyanchain}
    }
    \subfigure[TDC]{
        \includegraphics[width=0.3\columnwidth, height=0.25\columnwidth]{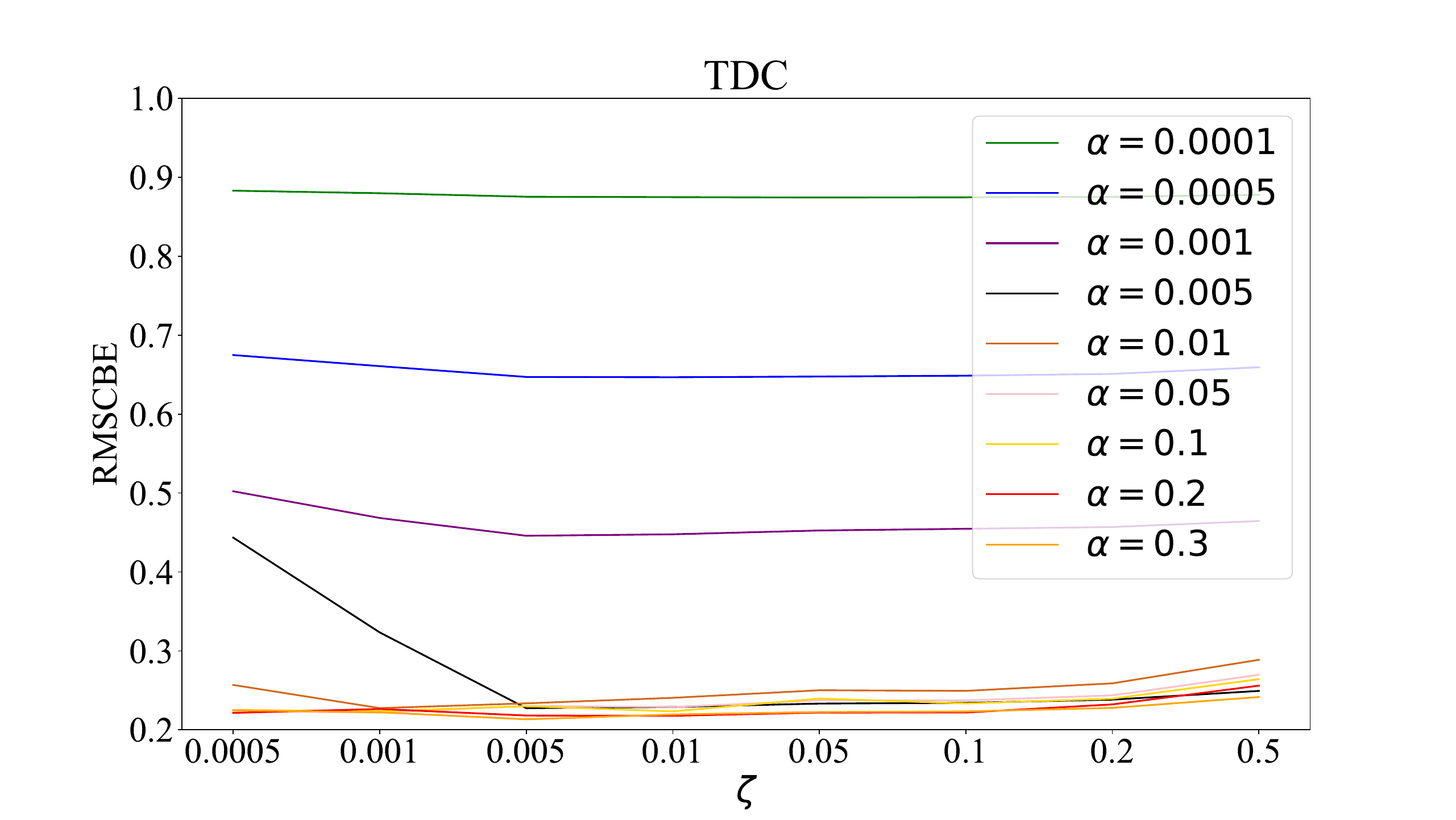}
        \label{tdc_boyanchain}
    }
    \subfigure[CTD]{
        \includegraphics[width=0.3\columnwidth, height=0.25\columnwidth]{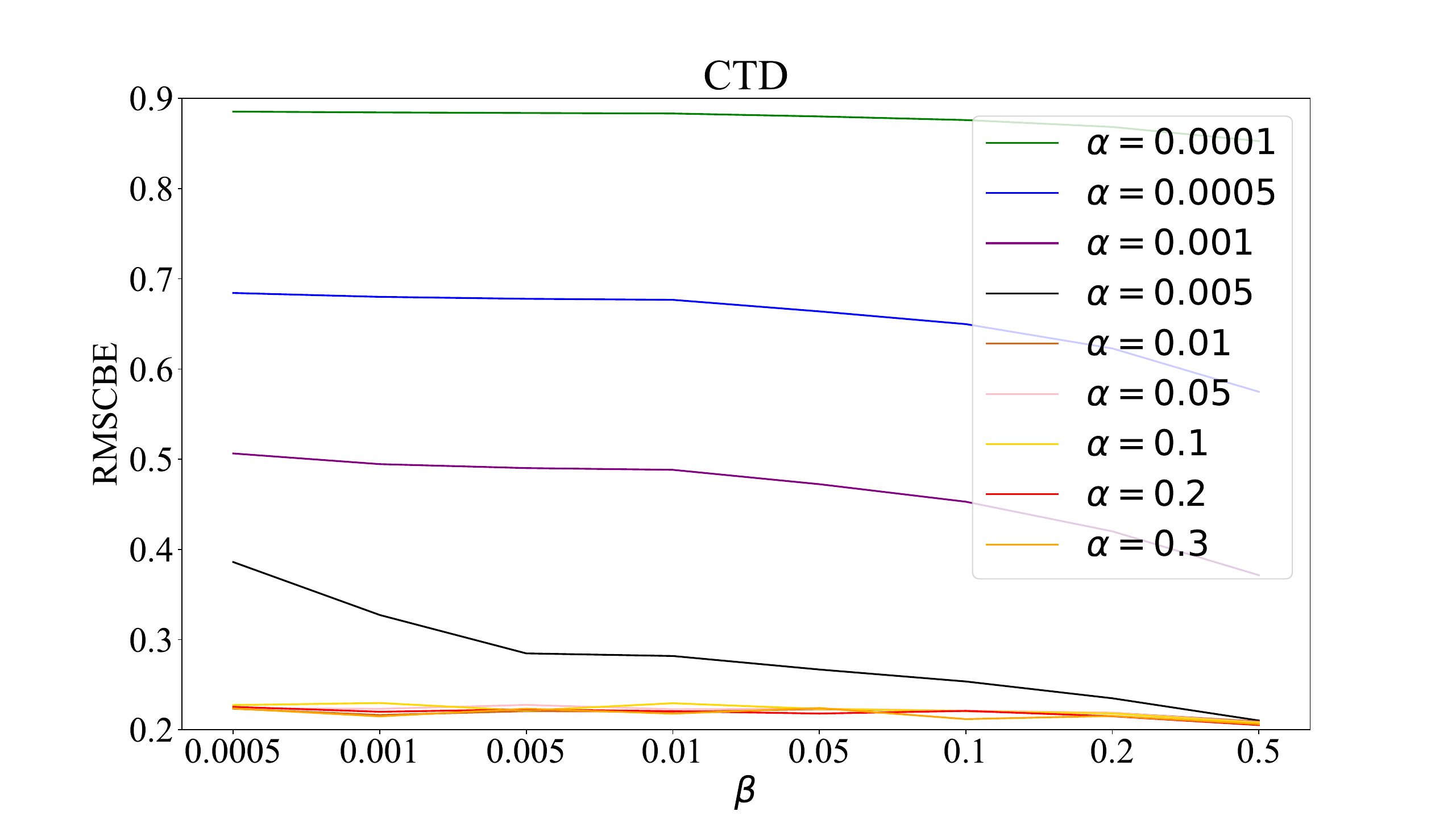}
        \label{ctd_boyanchain}
    }
    \\
    \subfigure[CTDC($\alpha=0.0001$)]{
        \includegraphics[width=0.3\columnwidth, height=0.25\columnwidth]{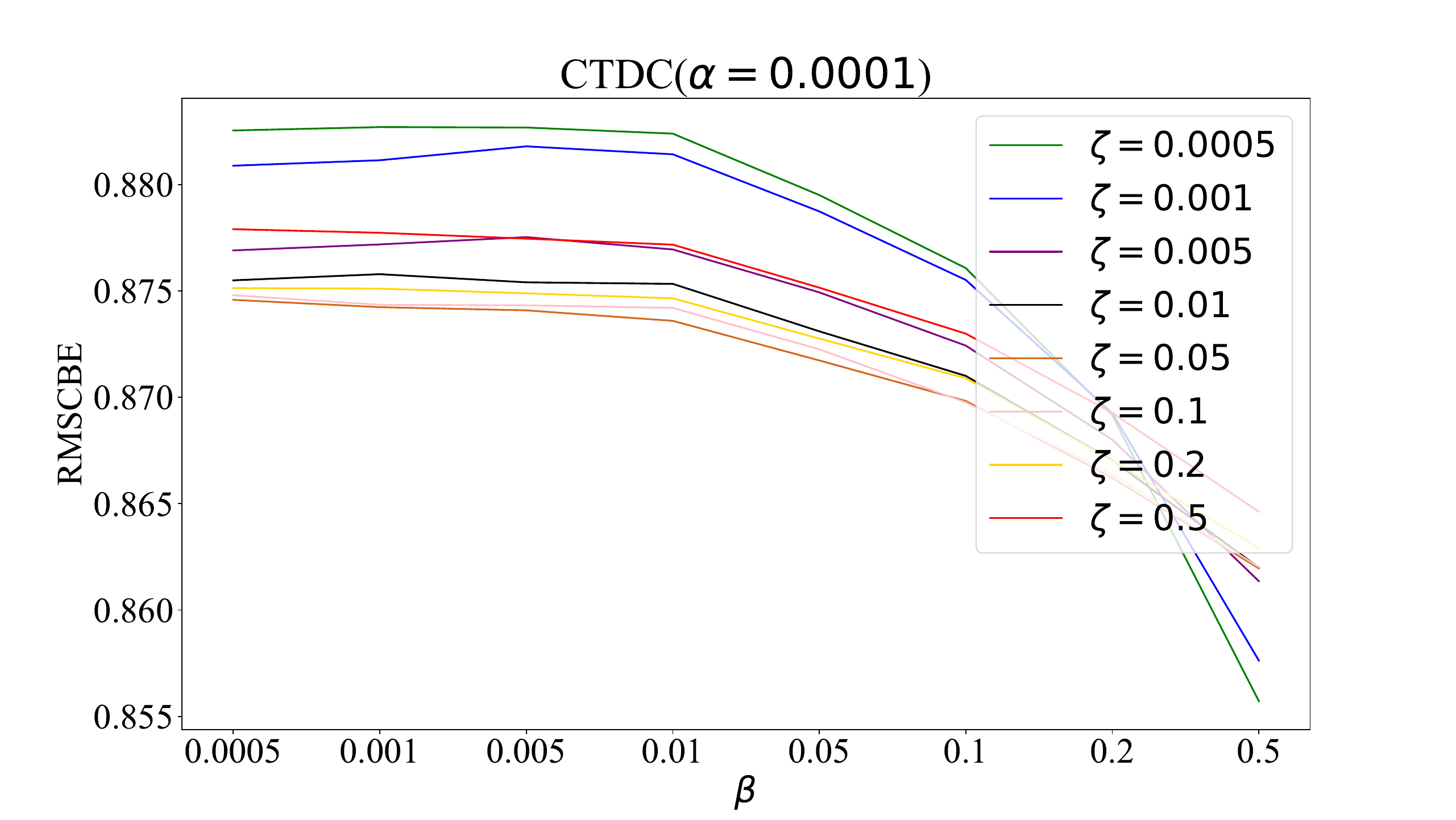}
        \label{ctdc_alpha_0001_boyanchain}
    }
    \subfigure[CTDC($\alpha=0.0005$)]{
        \includegraphics[width=0.3\columnwidth, height=0.25\columnwidth]{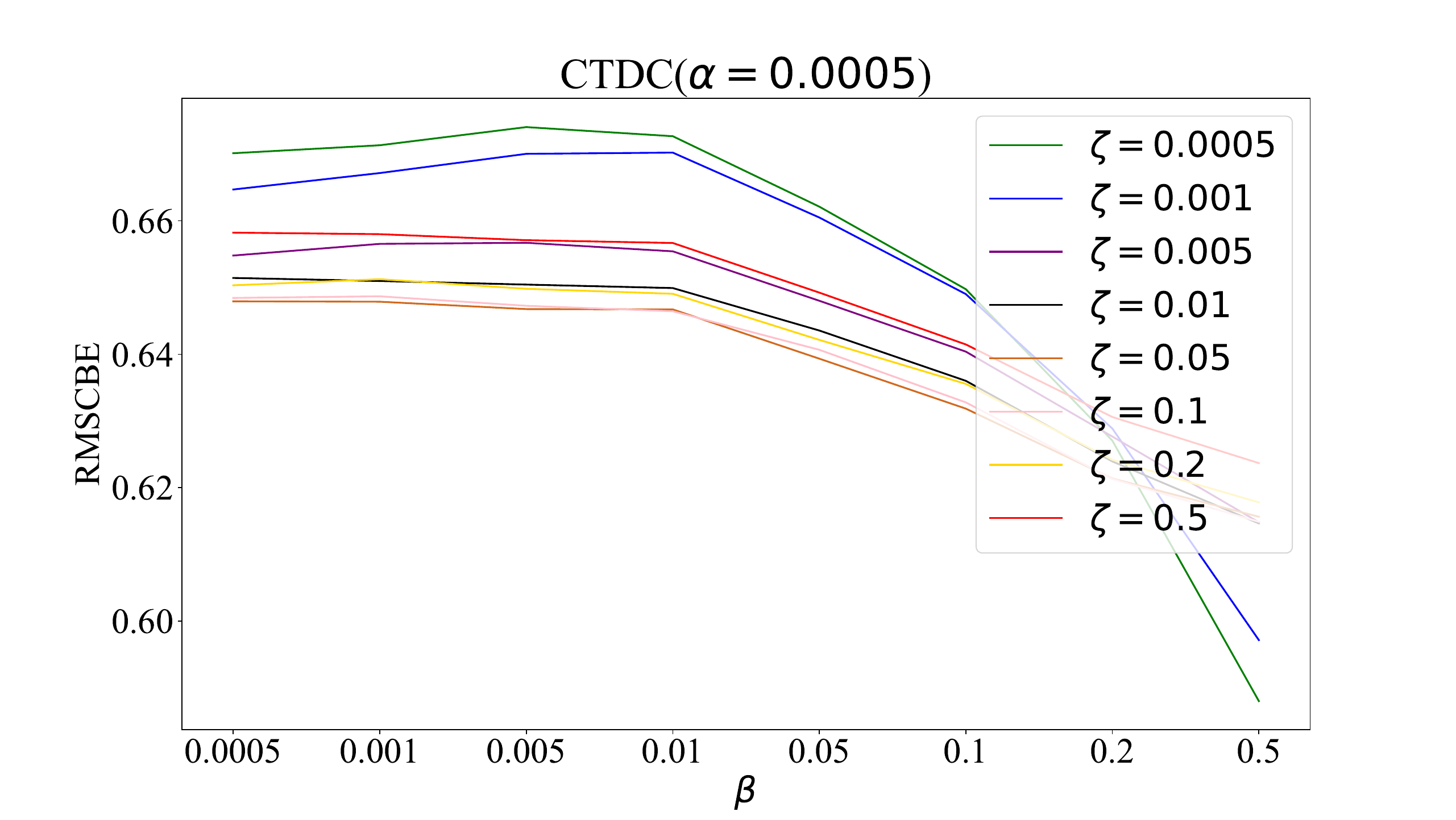}
        \label{ctdc_alpha_0005_boyanchain}
    }
    \subfigure[CTDC($\alpha=0.001$)]{
        \includegraphics[width=0.3\columnwidth, height=0.25\columnwidth]{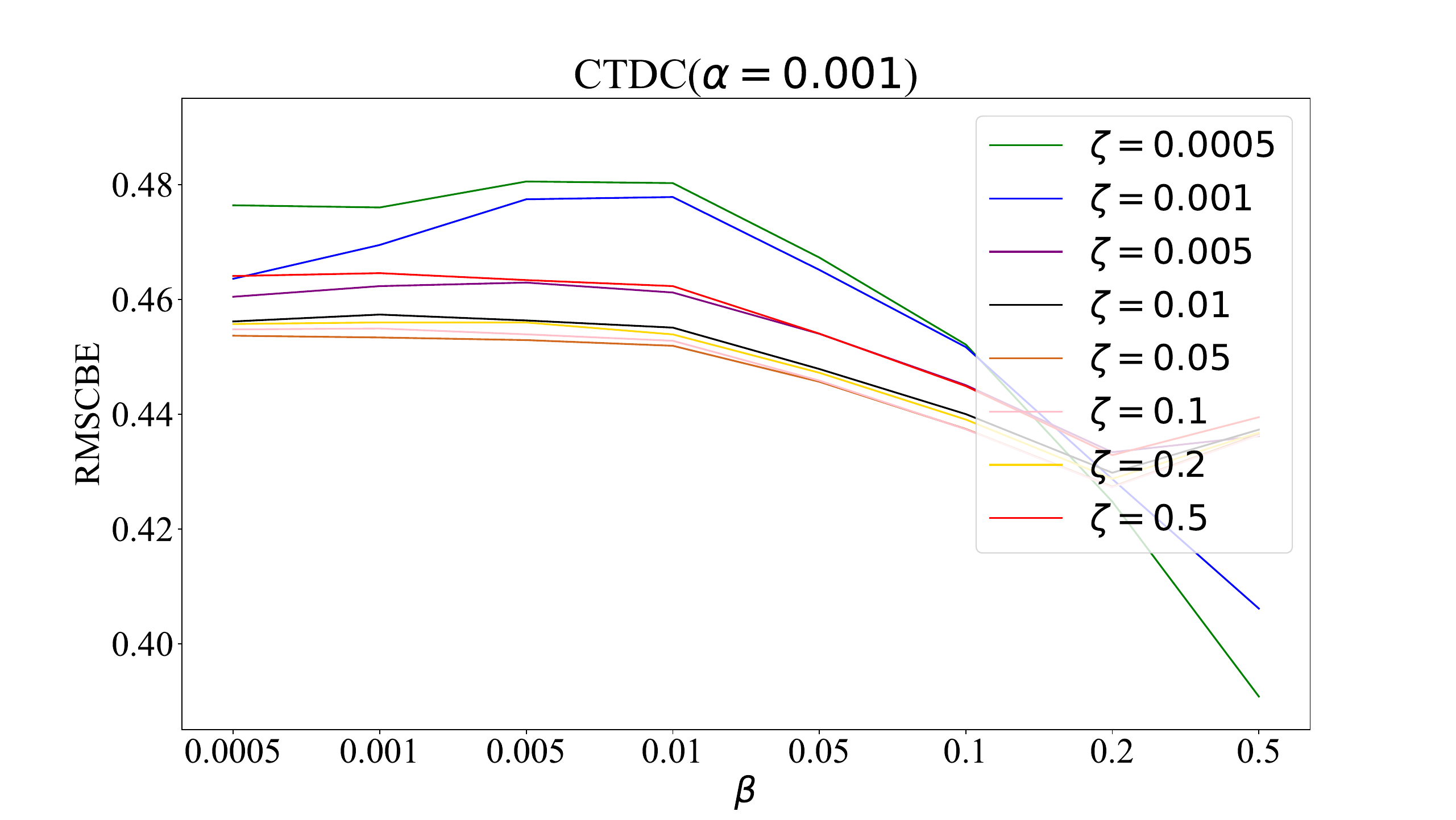}
        \label{ctdc_alpha_001_boyanchain}
    }
    \\
    \subfigure[CTDC($\alpha=0.005$)]{
        \includegraphics[width=0.3\columnwidth, height=0.25\columnwidth]{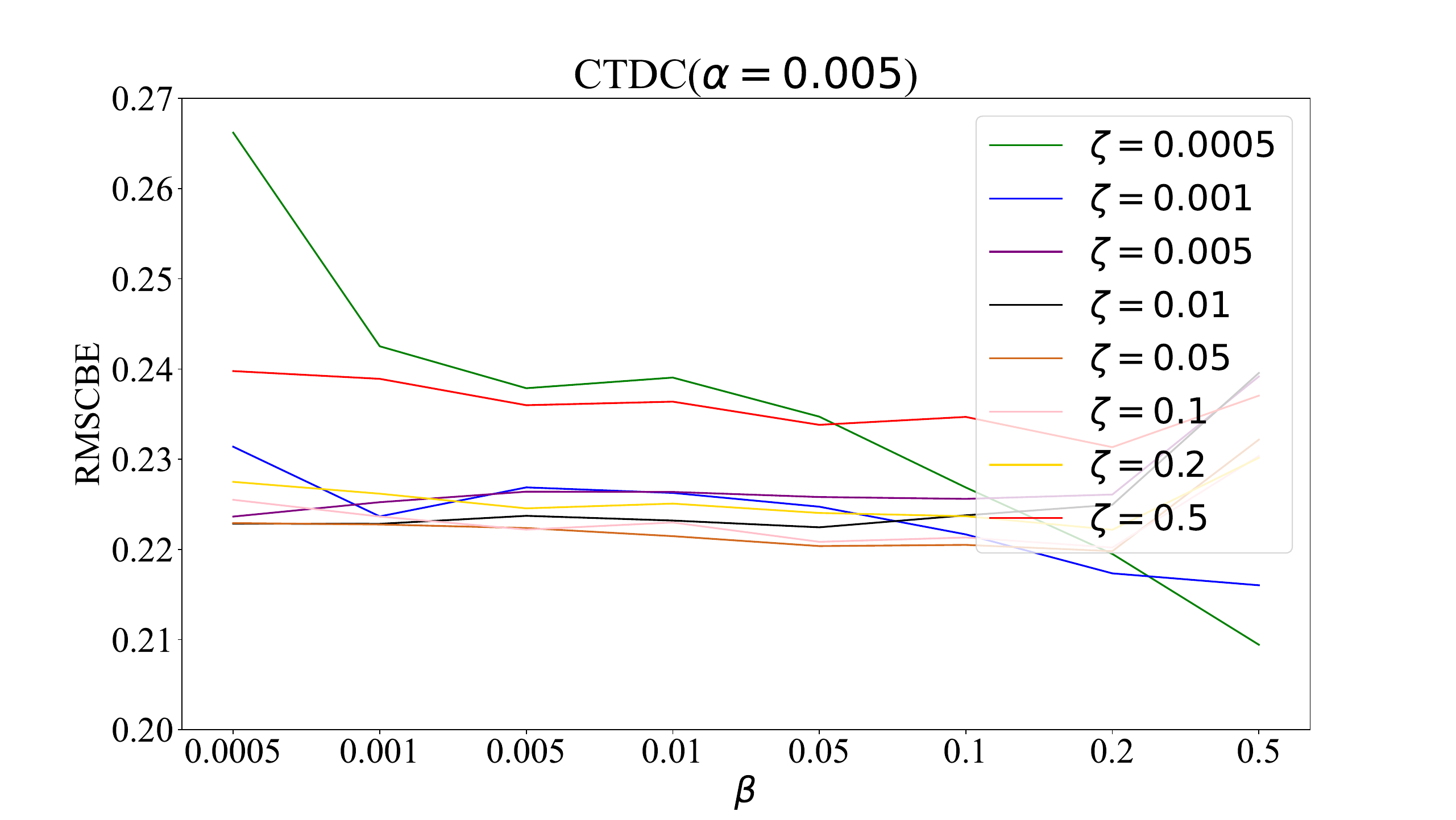}
        \label{ctdc_alpha_005_boyanchain}
    }
    \subfigure[CTDC($\alpha=0.01$)]{
        \includegraphics[width=0.3\columnwidth, height=0.25\columnwidth]{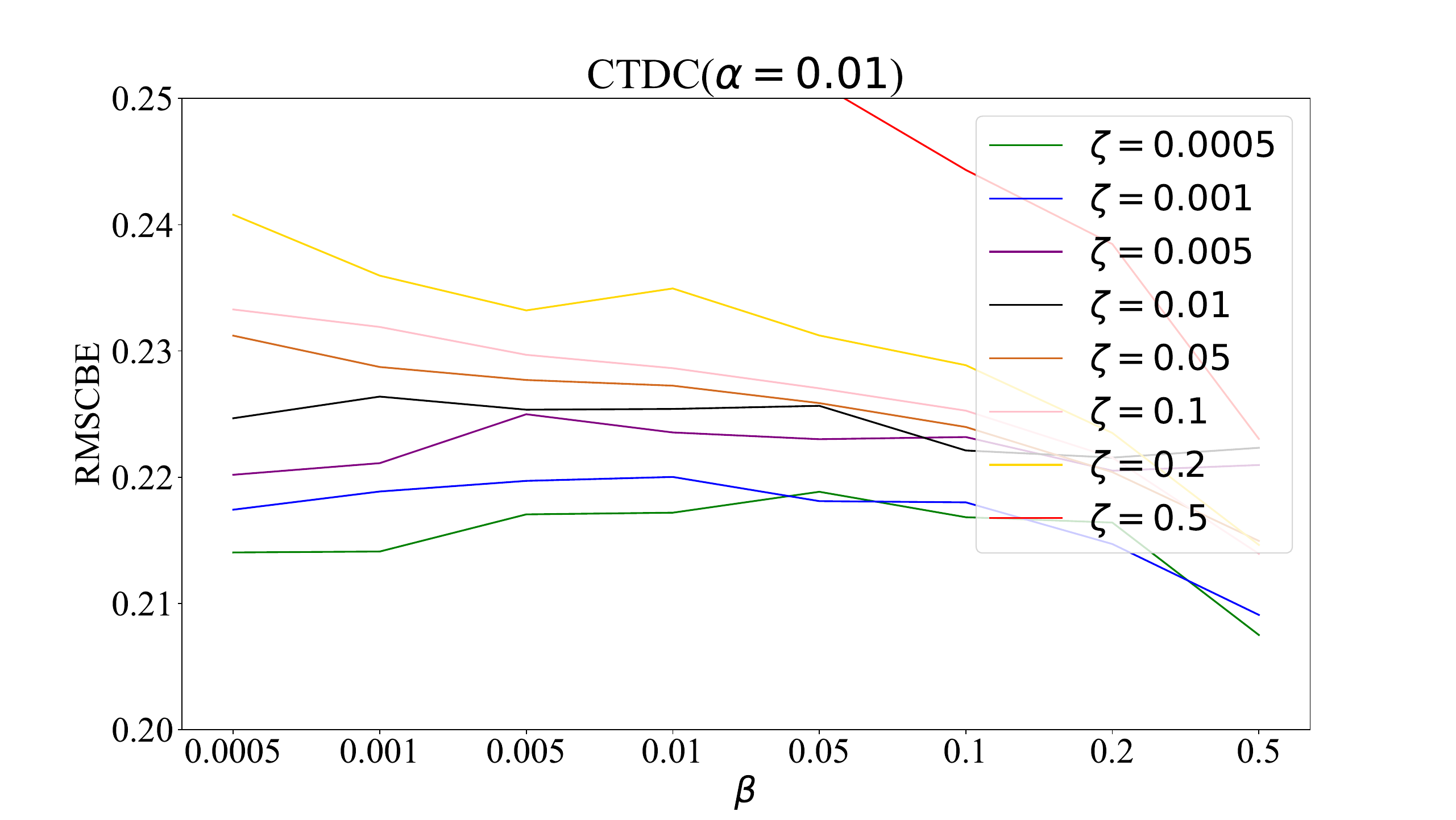}
        \label{ctdc_alpha_01_boyanchain}
    }
    \subfigure[CTDC($\alpha=0.05$)]{
        \includegraphics[width=0.3\columnwidth, height=0.25\columnwidth]{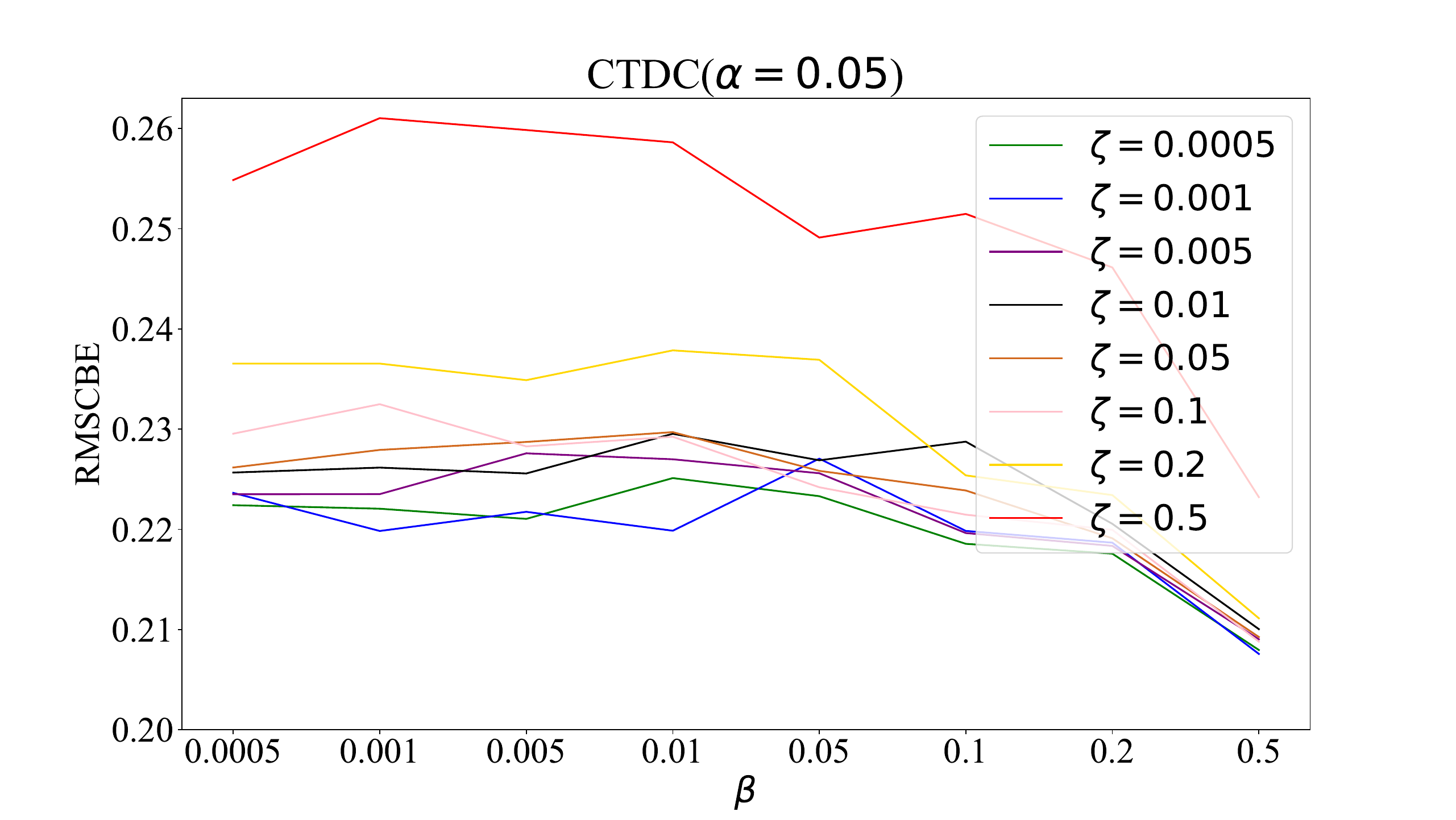}
        \label{ctdc_alpha_05_boyanchain}
    }
    \\
    \subfigure[CTDC($\alpha=0.1$)]{
        \includegraphics[width=0.3\columnwidth, height=0.25\columnwidth]{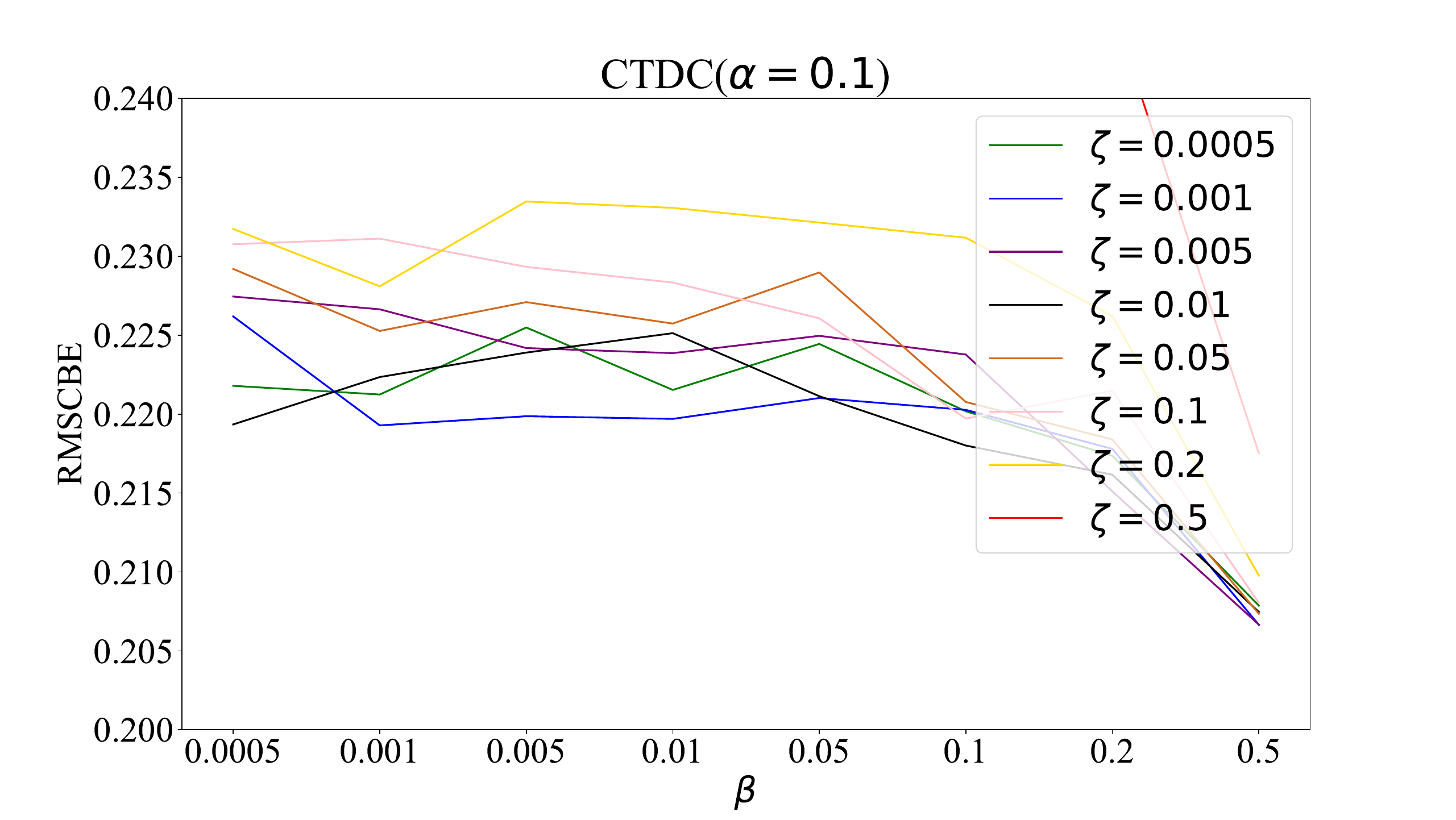}
        \label{ctdc_alpha_1_boyanchain}
    }
    \subfigure[CTDC($\alpha=0.2$)]{
        \includegraphics[width=0.3\columnwidth, height=0.25\columnwidth]{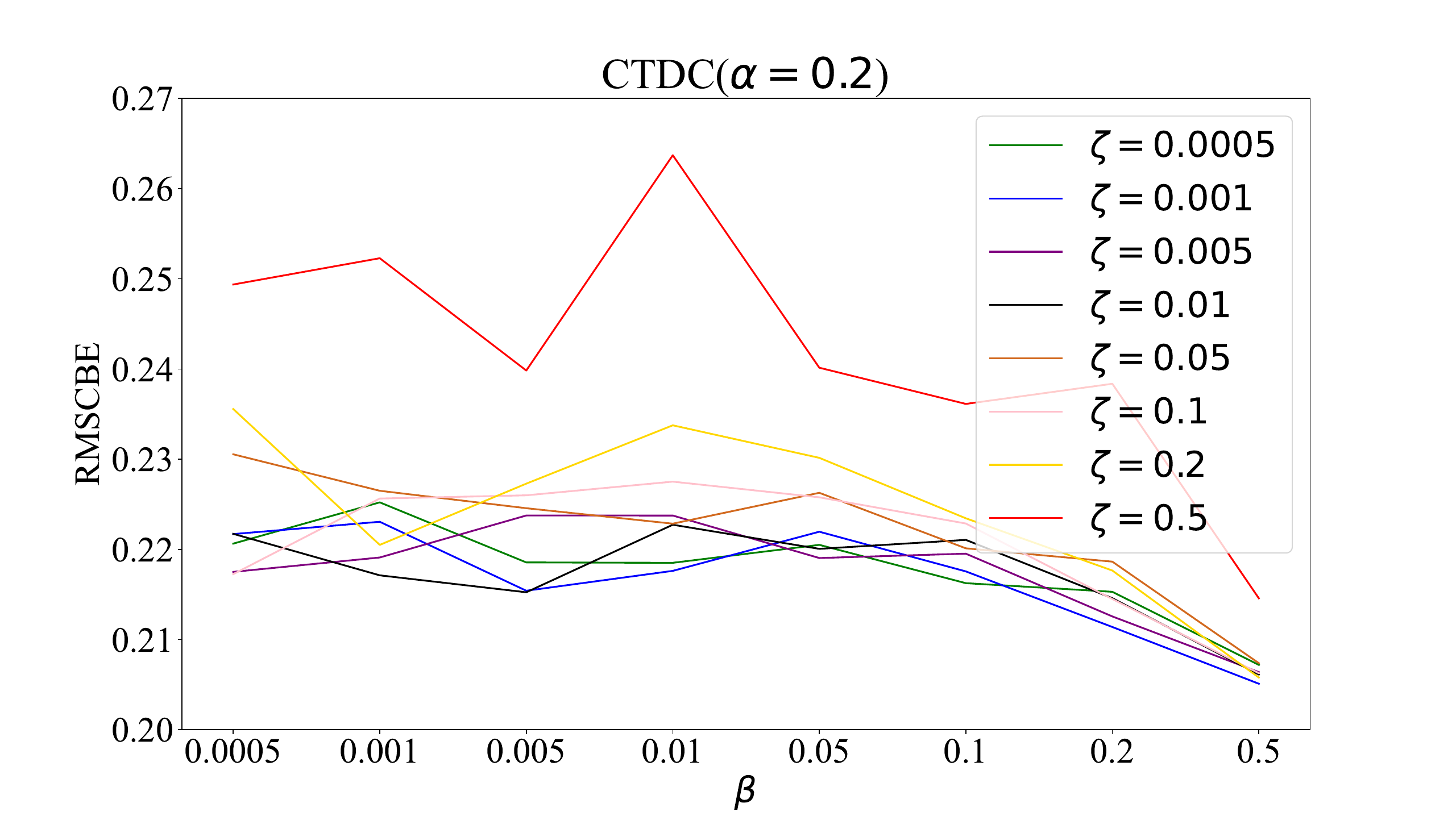}
        \label{ctdc_alpha_2_boyanchain}
    }
    \subfigure[CTDC($\alpha=0.3$)]{
        \includegraphics[width=0.3\columnwidth, height=0.25\columnwidth]{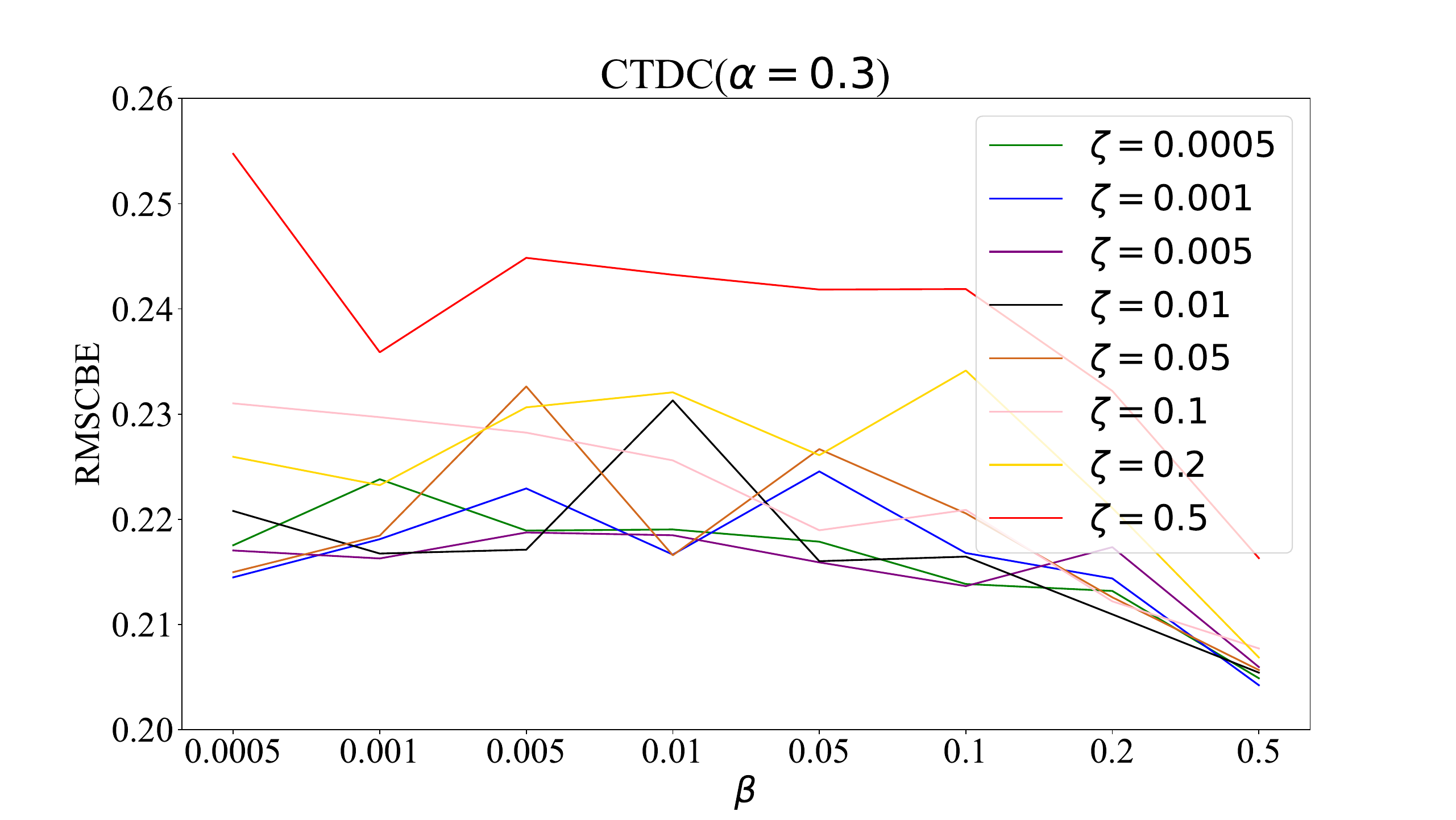}
        \label{ctdc_alpha_3_boyanchain}
    }
        \caption{Sensitivity of various algorithms to learning rates for Boyanchain.}
        \label{SensitivityBoyanchain}
    \end{center}
    \vskip -0.2in
\end{figure}
\subsection{2-state counterexample}
\label{appendix2state}
\textbf{2-state off-policy counterexample:} The ``1''$\rightarrow$``2'' problem has only two states \cite{sutton2016emphatic}. From each
state, there are two actions, left and right, which take 
the agent to the left or right state. All rewards are zero.
The feature $\bm{\Phi}=(1,2)^{\top}$ 
are assigned to the left and the right 
state. The behavior policy takes equal
probability to left or right
in both states, i.e., 
$
\textbf{P}_{1}=
\begin{bmatrix}
0.5 & 0.5 \\
0.5 & 0.5
\end{bmatrix}
$.
The target policy only selects action rights in both states, i.e., 
$
\textbf{P}_{2}=
\begin{bmatrix}
0 & 1 \\
0 & 1
\end{bmatrix}
$.
The state distribution of
the first policy is $\bm{d}_1 =(0.5,0.5)^{\top}$.
The state distribution of
the second policy is $\bm{d}_1 =(0,1)^{\top}$.
The discount factor is $\gamma=0.9$.
\begin{figure}[H]
    \begin{center}
    \label{bairdexampleoffpolicy}
    \includegraphics[width=0.3\columnwidth, height=0.1\columnwidth]{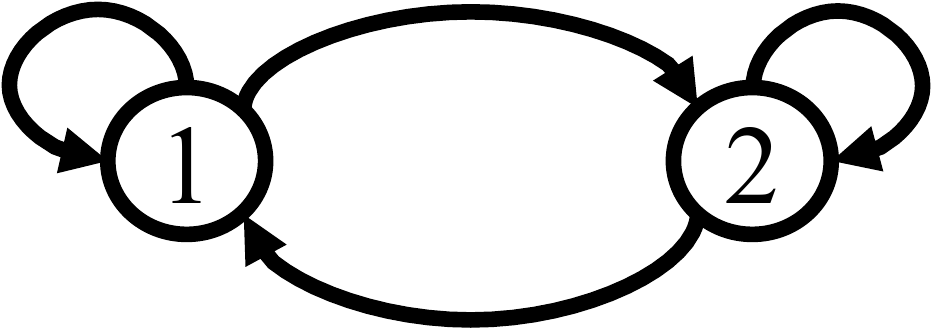}
        \caption{2-state off-policy counterexample.}
    \end{center}
\end{figure}
The $\alpha$ values for all algorithms are in the range of $\{0.0001, 0.0005, 0.001, 0.005, 0.01\}$. 
For the TDC and CTDC algorithm, the $\zeta$ values are in the range of $\{0.0005, 0.001, 0.005, 0.01, 0.05\}$. 
For the CTD and CTDC algorithm, the $\beta$ values are in the range of $\{0.0005, 0.001, 0.005, 0.01, 0.05, 0.1, 0.2\}$. 
\begin{figure}
    \vskip 0.2in
    \begin{center}
    \subfigure[TD]{
        \includegraphics[width=0.3\columnwidth, height=0.25\columnwidth]{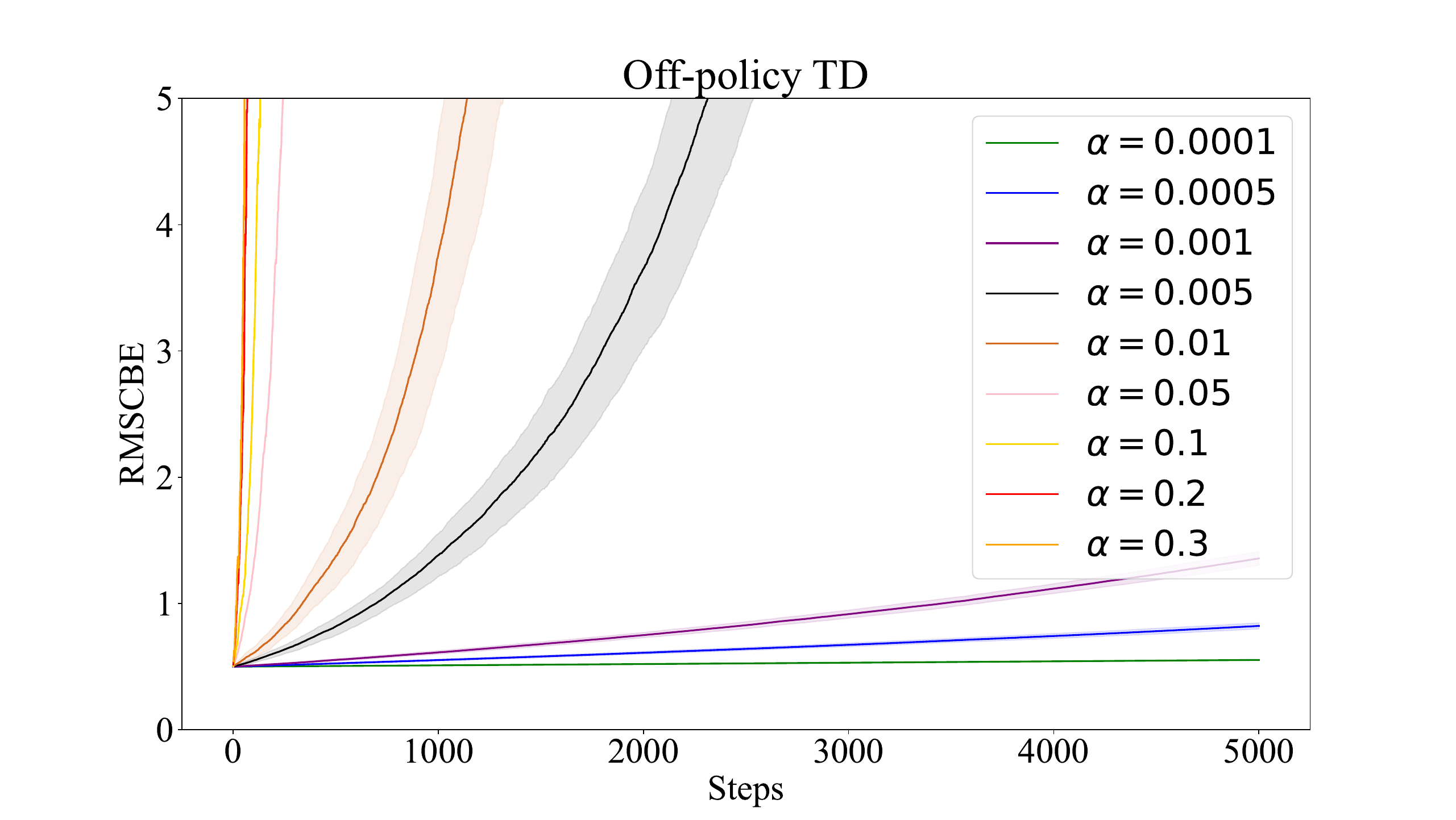}
        \label{td_2state}
    }
    \subfigure[TDC]{
        \includegraphics[width=0.3\columnwidth, height=0.25\columnwidth]{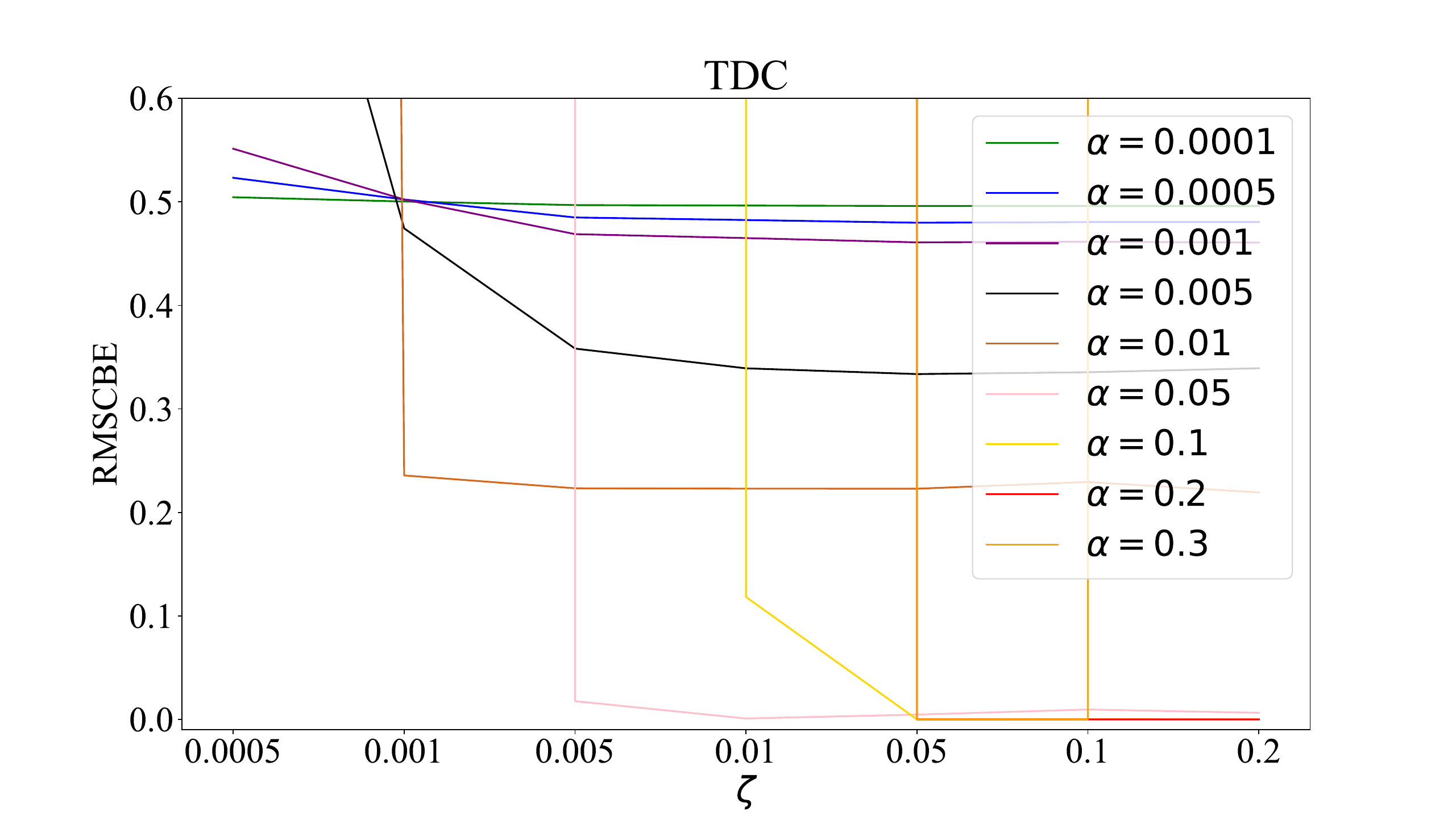}
        \label{tdc_2state}
    }
    \subfigure[CTD]{
        \includegraphics[width=0.3\columnwidth, height=0.25\columnwidth]{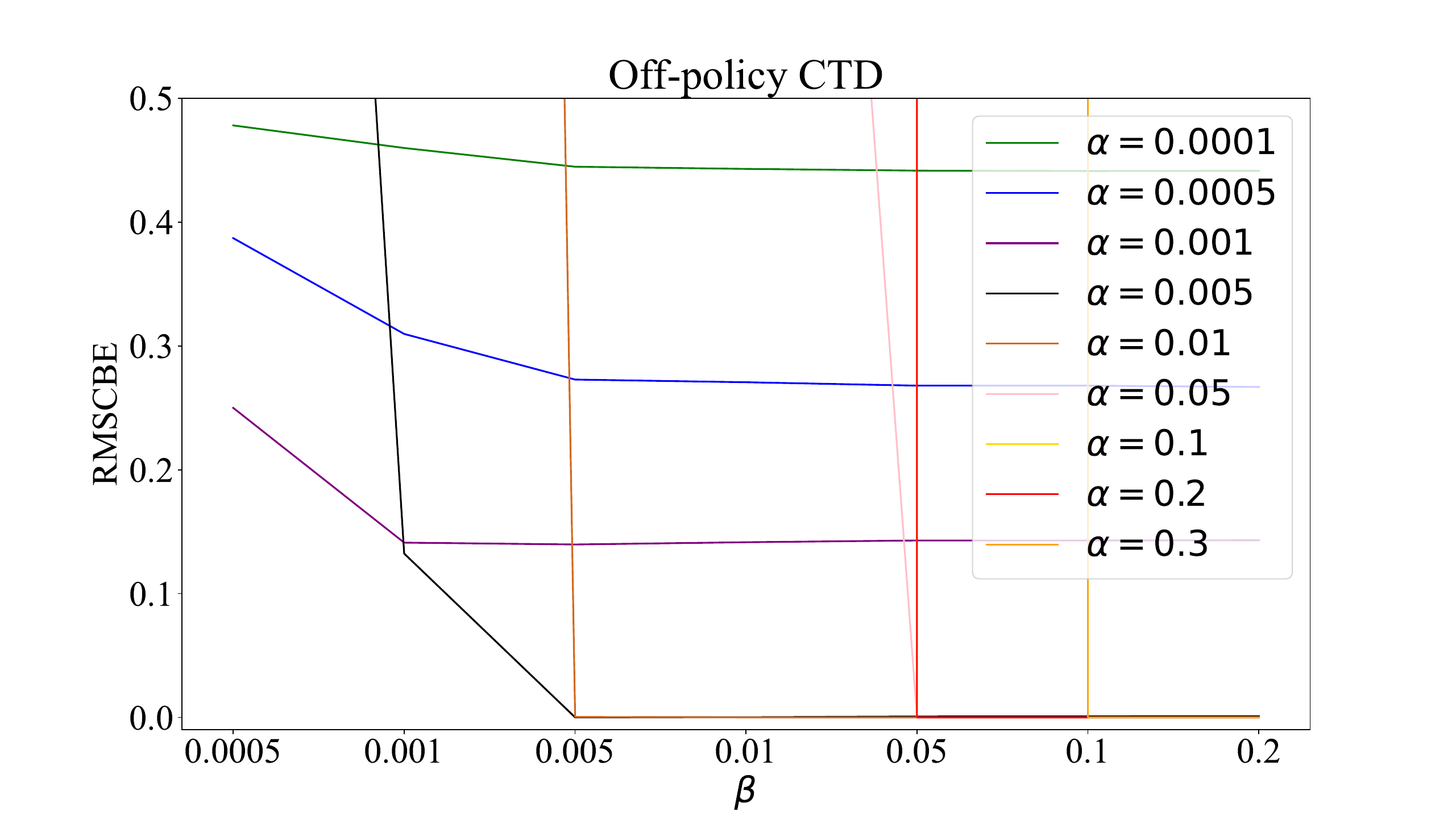}
        \label{ctd_2state}
    }
    \\
    \subfigure[CTDC($\alpha=0.0001$)]{
        \includegraphics[width=0.3\columnwidth, height=0.25\columnwidth]{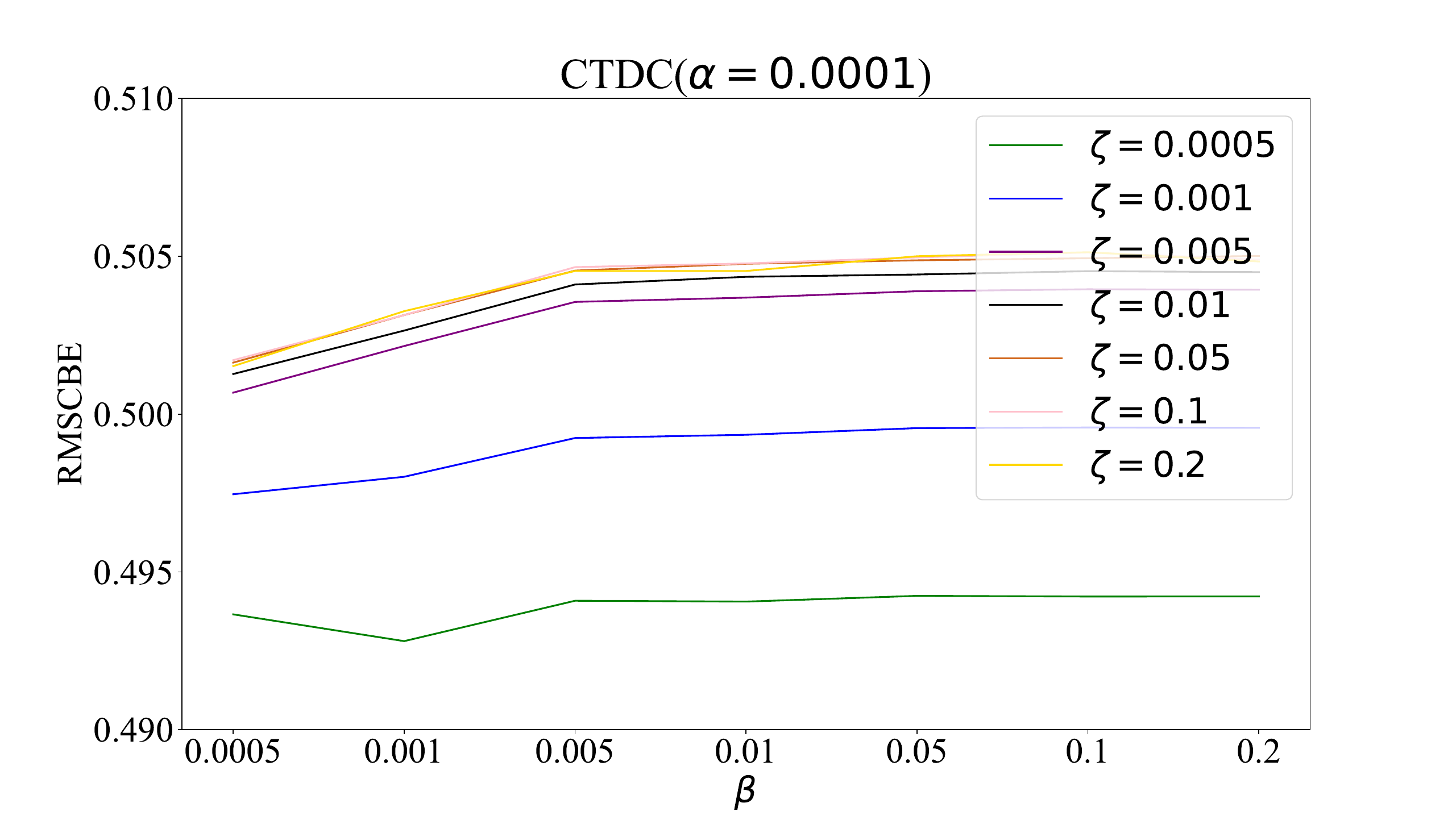}
        \label{ctdc_alpha_0001_2state}
    }
    \subfigure[CTDC($\alpha=0.0005$)]{
        \includegraphics[width=0.3\columnwidth, height=0.25\columnwidth]{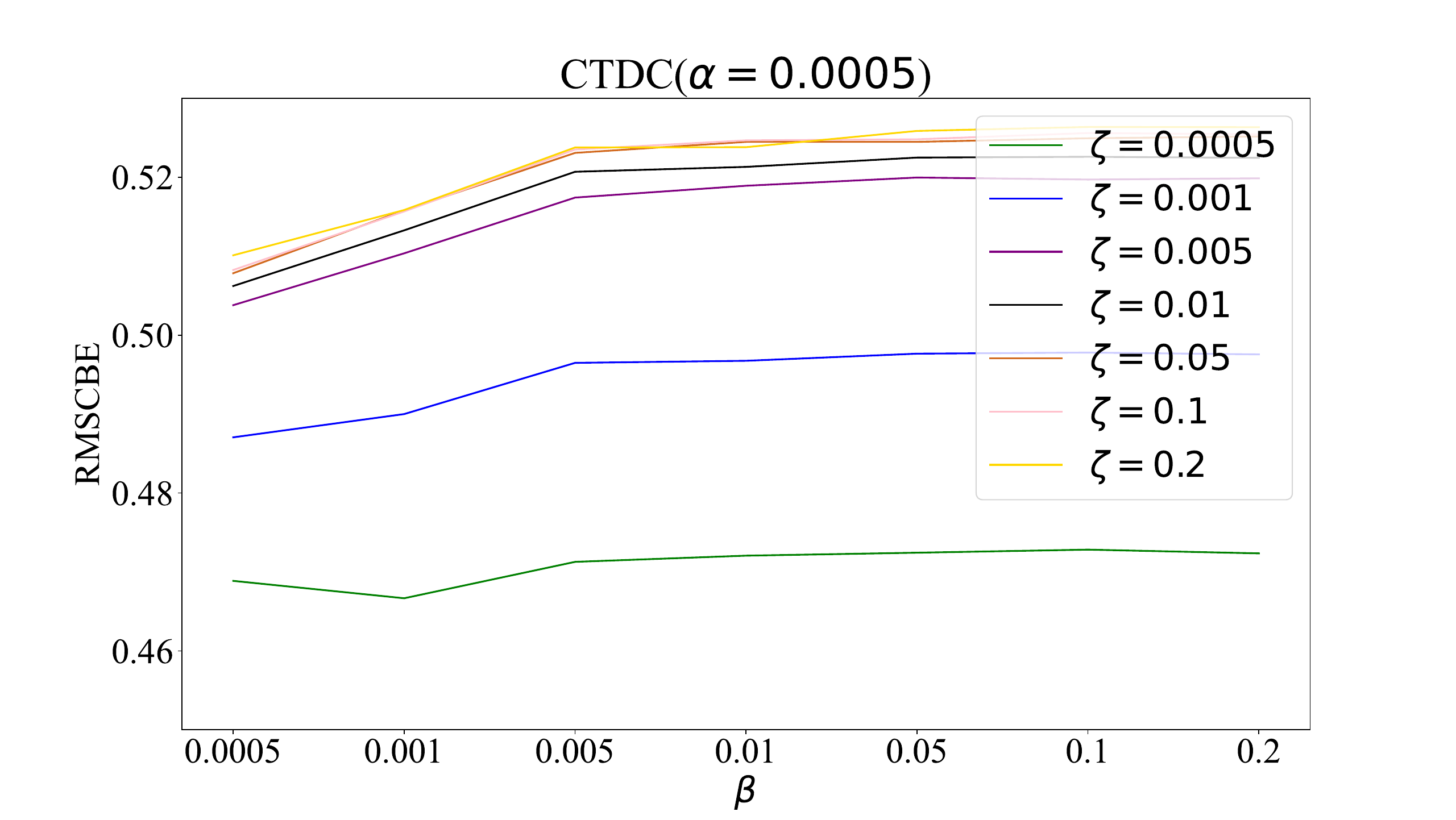}
        \label{ctdc_alpha_0005_2state}
    }
    \subfigure[CTDC($\alpha=0.001$)]{
        \includegraphics[width=0.3\columnwidth, height=0.25\columnwidth]{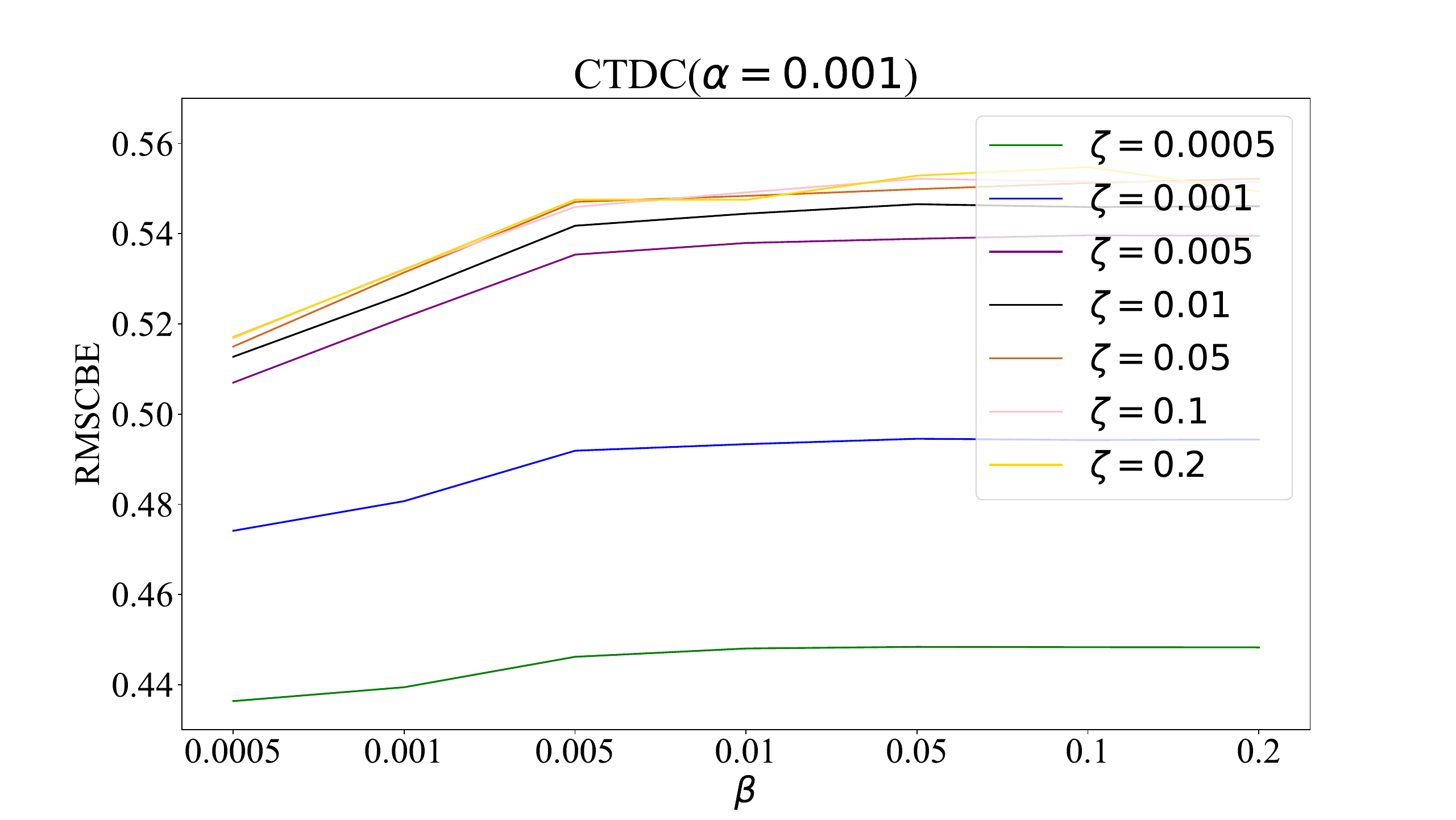}
        \label{ctdc_alpha_001_2state}
    }
    \\
    \subfigure[CTDC($\alpha=0.005$)]{
        \includegraphics[width=0.3\columnwidth, height=0.25\columnwidth]{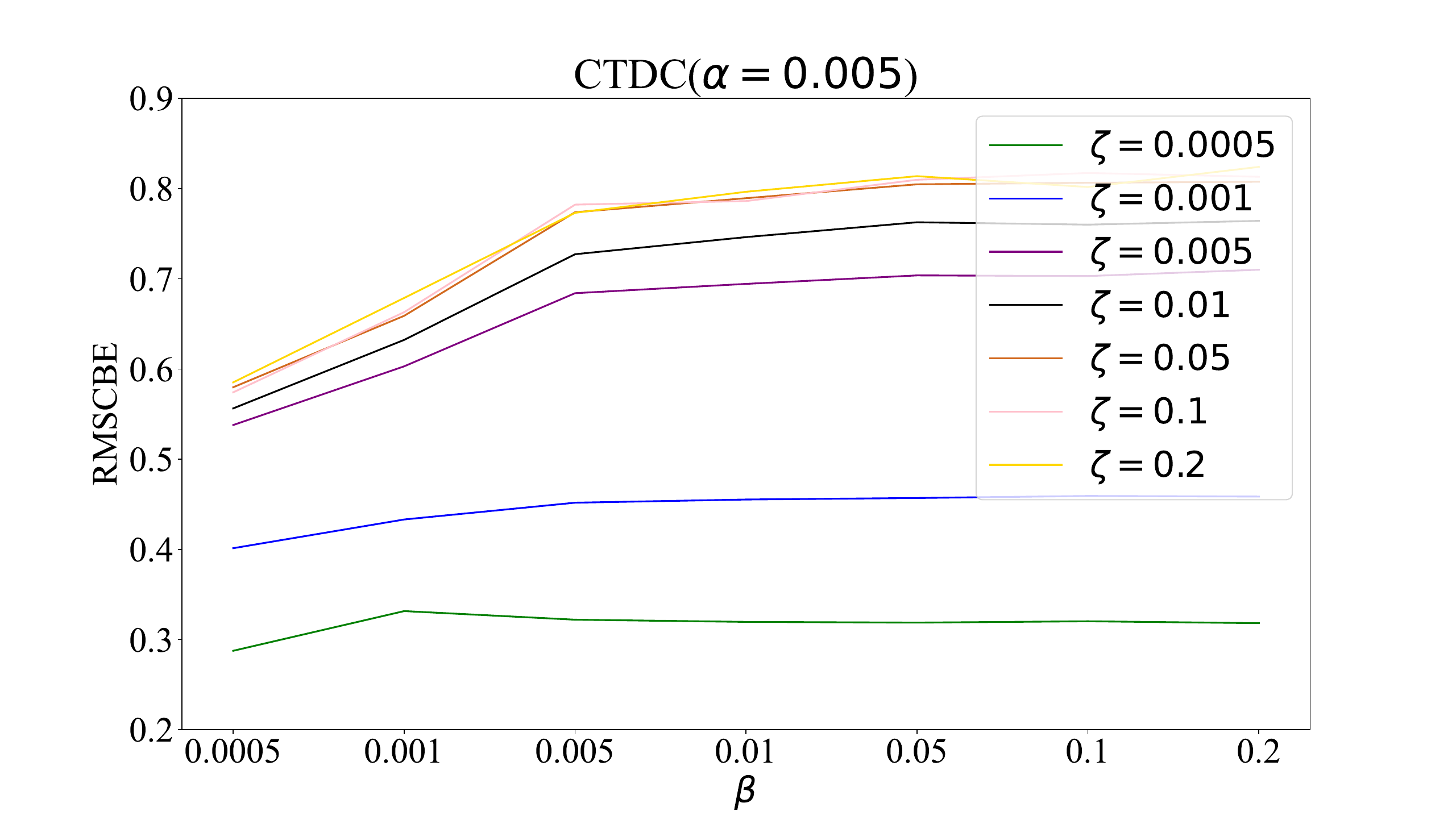}
        \label{ctdc_alpha_005_2state}
    }
    \subfigure[CTDC($\alpha=0.01$)]{
        \includegraphics[width=0.3\columnwidth, height=0.25\columnwidth]{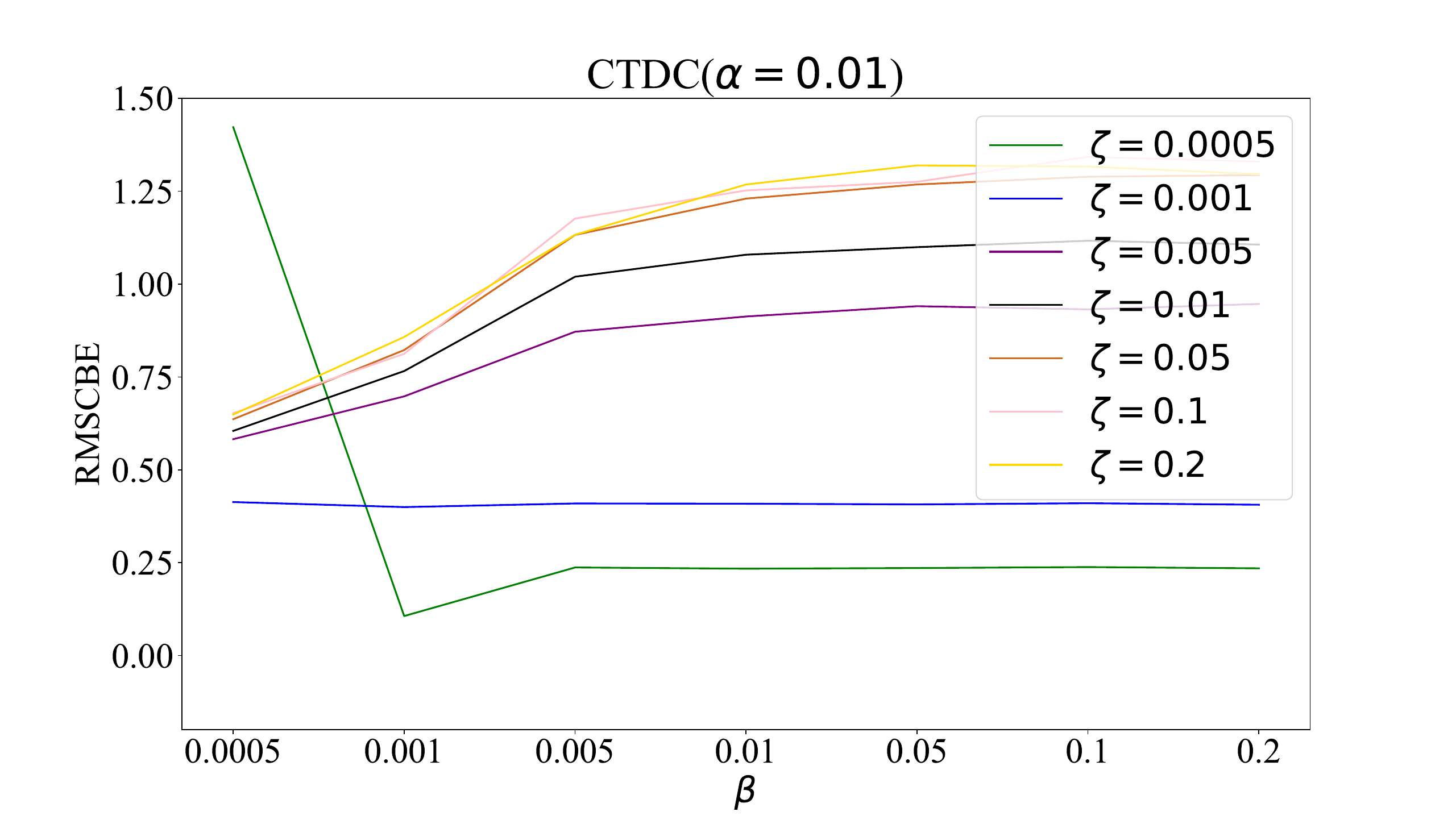}
        \label{ctdc_alpha_01_2state}
    }
    \subfigure[CTDC($\alpha=0.05$)]{
        \includegraphics[width=0.3\columnwidth, height=0.25\columnwidth]{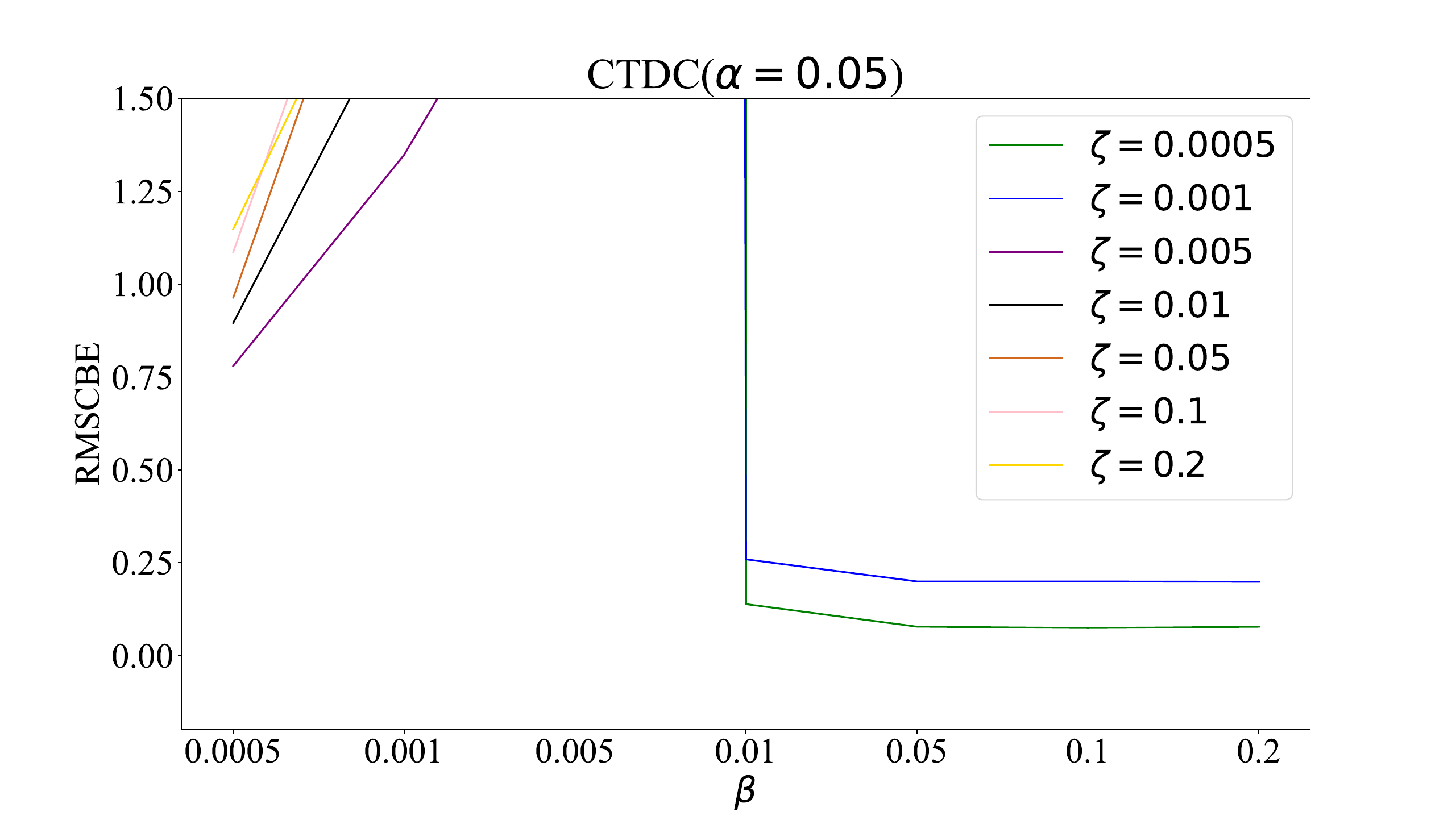}
        \label{ctdc_alpha_05_2staten}
    }
    \\
    \subfigure[CTDC($\alpha=0.1$)]{
        \includegraphics[width=0.3\columnwidth, height=0.25\columnwidth]{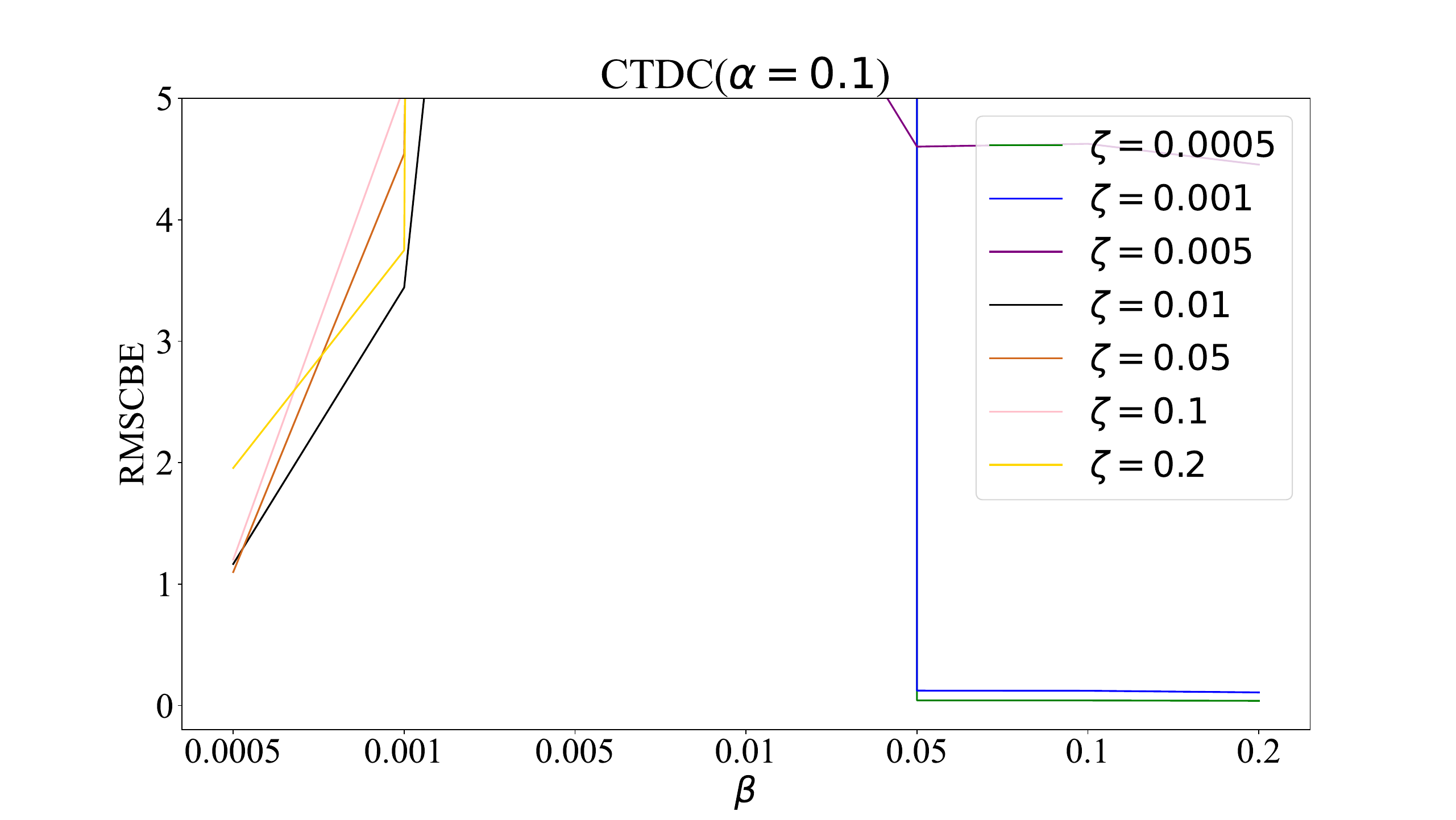}
        \label{ctdc_alpha_1_2state}
    }
    \subfigure[CTDC($\alpha=0.2$)]{
        \includegraphics[width=0.3\columnwidth, height=0.25\columnwidth]{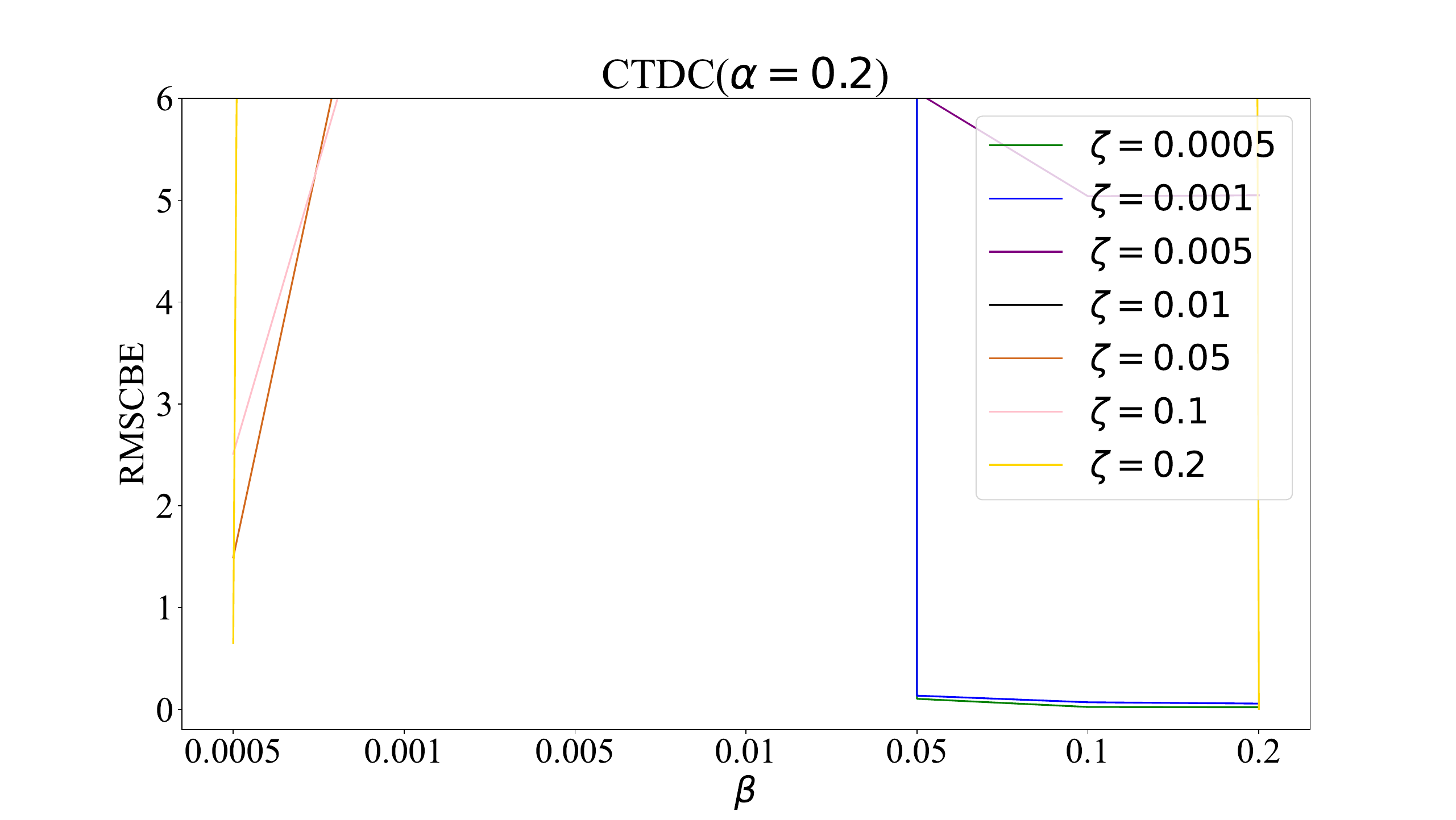}
        \label{ctdc_alpha_2_2state}
    }
    \subfigure[CTDC($\alpha=0.3$)]{
        \includegraphics[width=0.3\columnwidth, height=0.25\columnwidth]{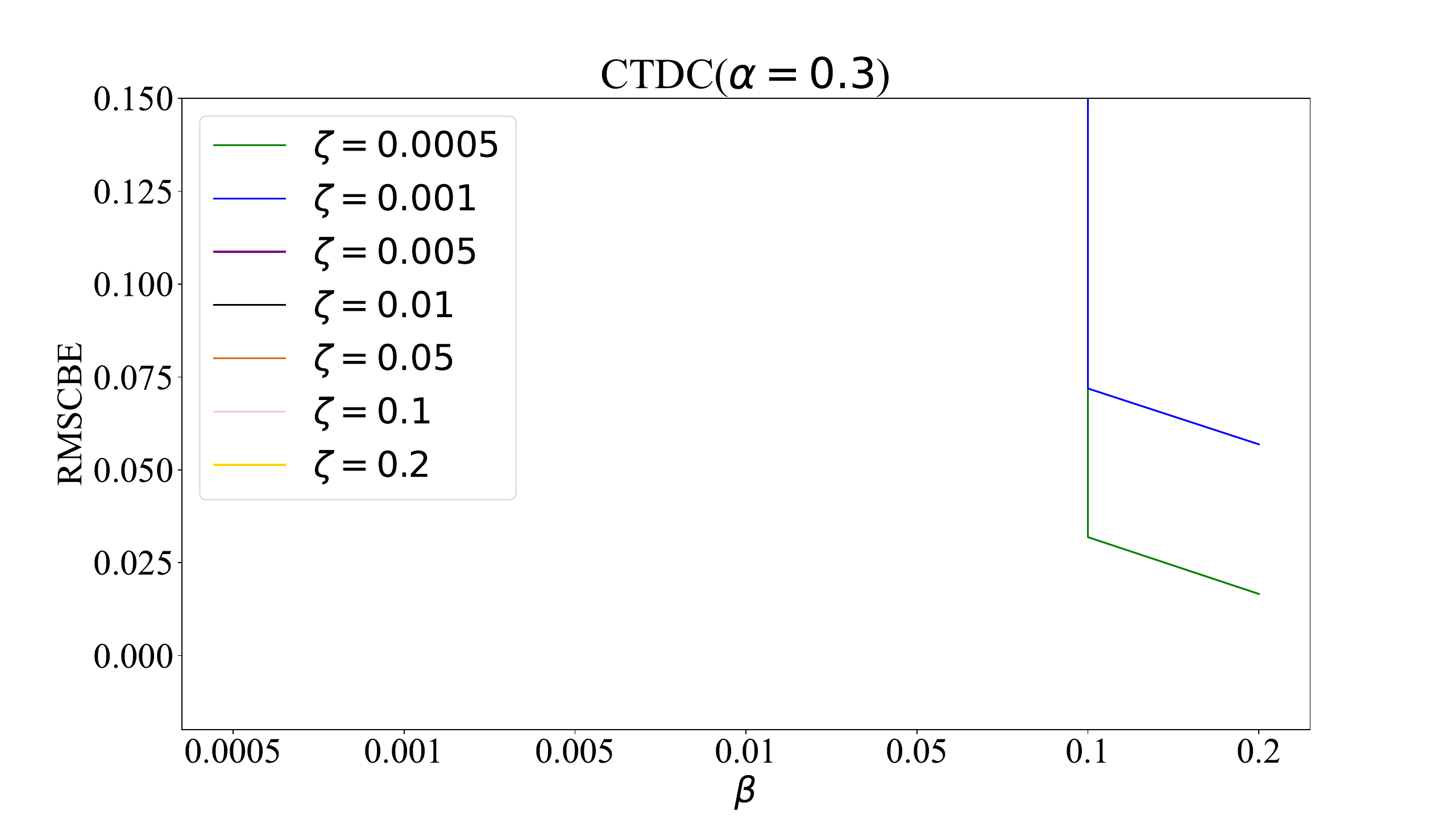}
        \label{ctdc_alpha_3_2staten}
    }
        \caption{Sensitivity of various algorithms to learning rates for 2-state counterexample.}
        \label{Sensitivity2state}
    \end{center}
    \vskip -0.2in
\end{figure}
\subsection{7-state counterexample}
\label{appendix7state}
\begin{figure}[H]
    \begin{center}
        \resizebox{5cm}{3cm}{
\begin{tikzpicture}[smooth]
\node[coordinate] (origin) at (0.3,0) {};
\node[coordinate] (num7) at (3,0) {};
\node[coordinate] (num1) at (1,2.5) {};
\path (num7) ++ (-10:0.5cm) node (num7_bright1) [coordinate] {};
\path (num7) ++ (-30:0.7cm) node (num7_bright2) [coordinate] {};
\path (num7) ++ (-60:0.35cm) node (num7_bright3) [coordinate] {};
\path (num7) ++ (-60:0.6cm) node (num7_bright4) [coordinate] {};
\path (origin) ++ (90:3cm) node (origin_above) [coordinate] {};
\path (origin_above) ++ (0:5.7cm) node (origin_aright) [coordinate] {};
\path (num1) ++ (90:0.5cm) node (num1_a) [coordinate] {};
\path (num1) ++ (-90:0.3cm) node (num1_b) [coordinate] {};

\path (num1) ++ (0:1cm) node (num2) [coordinate] {};
\path (num1_a) ++ (0:1cm) node (num2_a) [coordinate] {};
\path (num1_b) ++ (0:1cm) node (num2_b) [coordinate] {};
\path (num2) ++ (0:1cm) node (num3) [coordinate] {};
\path (num2_a) ++ (0:1cm) node (num3_a) [coordinate] {};
\path (num2_b) ++ (0:1cm) node (num3_b) [coordinate] {};
\path (num3) ++ (0:1cm) node (num4) [coordinate] {};
\path (num3_a) ++ (0:1cm) node (num4_a) [coordinate] {};
\path (num3_b) ++ (0:1cm) node (num4_b) [coordinate] {};
\path (num4) ++ (0:1cm) node (num5) [coordinate] {};
\path (num4_a) ++ (0:1cm) node (num5_a) [coordinate] {};
\path (num4_b) ++ (0:1cm) node (num5_b) [coordinate] {};
\path (num5) ++ (0:1cm) node (num6) [coordinate] {};
\path (num5_a) ++ (0:1cm) node (num6_a) [coordinate] {};
\path (num5_b) ++ (0:1cm) node (num6_b) [coordinate] {};


\draw[dashed,line width = 0.03cm,xshift=3cm] plot[tension=0.06]
coordinates{(num7) (origin) (origin_above) (origin_aright)}; 

\draw[->,>=stealth,line width = 0.02cm,xshift=3cm] plot[tension=0.5]
coordinates{(num7) (num7_bright1) (num7_bright2)(num7_bright4) (num7_bright3)};
 
\node[line width = 0.02cm,shape=circle,fill=white,draw=black] (g) at (num7) {7};

\draw[<->,>=stealth,dashed,line width = 0.03cm,] (num1) -- (num1_a) ;
\node[line width = 0.02cm,shape=circle,fill=white,draw=black] (a) at (num1_b) {1};
\draw[<->,>=stealth,dashed,line width = 0.03cm,] (num2) -- (num2_a) ;
\node[line width = 0.02cm,shape=circle,fill=white,draw=black] (b) at (num2_b) {2};
\draw[<->,>=stealth,dashed,line width = 0.03cm,] (num3) -- (num3_a) ;
\node[line width = 0.02cm,shape=circle,fill=white,draw=black] (c) at (num3_b) {3};
\draw[<->,>=stealth,dashed,line width = 0.03cm,] (num4) -- (num4_a) ;
\node[line width = 0.02cm,shape=circle,fill=white,draw=black] (d) at (num4_b) {4};
\draw[<->,>=stealth,dashed,line width = 0.03cm,] (num5) -- (num5_a) ;
\node[line width = 0.02cm,shape=circle,fill=white,draw=black] (e) at (num5_b) {5};
\draw[<->,>=stealth,dashed,line width = 0.03cm,] (num6) -- (num6_a) ;
\node[line width = 0.02cm,shape=circle,fill=white,draw=black] (f) at (num6_b) {6};

\draw[->,>=stealth,line width = 0.02cm] (a)--(g);
\draw[->,>=stealth,line width = 0.02cm] (b)--(g);
\draw[->,>=stealth,line width = 0.02cm] (c)--(g);
\draw[->,>=stealth,line width = 0.02cm] (d)--(g);
\draw[->,>=stealth,line width = 0.02cm] (e)--(g);
\draw[->,>=stealth,line width = 0.02cm] (f)--(g);
\end{tikzpicture}
}
        \label{7statebairdexample}
        \caption{7-state off-policy counterexample.}
    \end{center}
\end{figure}
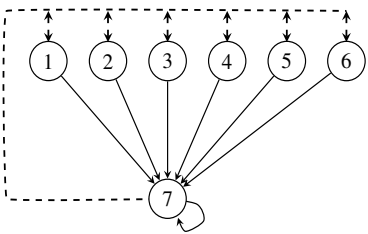

\textbf{7-state off-policy counterexample:} This task is well known as a
counterexample, in which TD diverges \cite{baird1995residual,sutton2009fast}. As
shown in Figure \ref{7statebairdexample}, reward for each transition is zero. Thus the true values are zeros for all states and for any given policy. The behaviour policy
chooses actions represented by solid lines with a probability of $\frac{1}{7}$
and actions represented by dotted lines with a probability of $\frac{6}{7}$. The
target policy is expected to choose the solid line with more probability than $\frac{1}{7}$,
and it chooses the solid line with probability of $1$ in this paper.
 The discount factor $\gamma =0.99$. 
The feature matrix of 7-state version of Baird's off-policy counterexample is
defined as follows \ref{appendix7state} \cite{baird1995residual,sutton2009fast,maei2011gradient}:
\begin{equation*}
    \bm{\Phi}_{Counter}=\left[ 
\begin{array}{cccccccc}
1 & 2& 0& 0& 0& 0& 0& 0\\
1 & 0& 2& 0& 0& 0& 0& 0\\
1 & 0& 0& 2& 0& 0& 0& 0\\
1 & 0& 0& 0& 2& 0& 0& 0\\
1 & 0& 0& 0& 0& 2& 0& 0\\
1 & 0& 0& 0& 0& 0& 2& 0\\
2 & 0& 0& 0& 0& 0& 0& 1
\end{array}\right]
\end{equation*}
The $\alpha$ values for all algorithms are in the range of $\{0.0001, 0.0005, 0.001, 0.005, 0.01, 0.05, 0.1, 0.2, 0.3\}$. 
For the TDC and CTDC algorithm, the $\zeta$ values are in the range of $\{0.0005, 0.001, 0.005, 0.01, 0.05, 0.1, 0.2\}$. 
For the CTD and CTDC algorithm, the $\beta$ values are in the range of $\{0.0005, 0.001, 0.005, 0.01, 0.05, 0.1, 0.2\}$. 
\begin{figure}
    \vskip 0.2in
    \begin{center}
    \subfigure[TD]{
        \includegraphics[width=0.3\columnwidth, height=0.25\columnwidth]{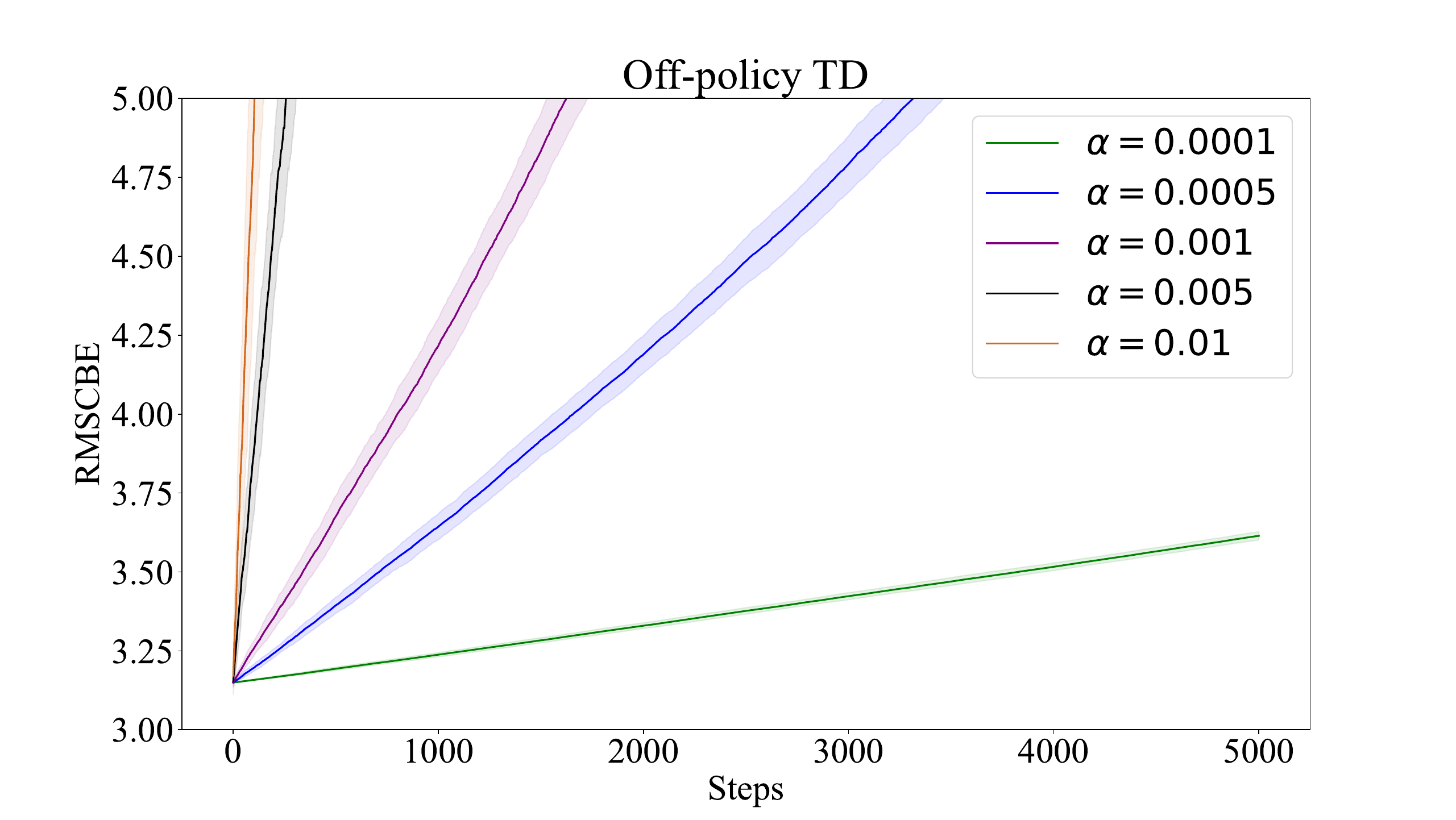}
        \label{td_7state}
    }
    \subfigure[TDC]{
        \includegraphics[width=0.3\columnwidth, height=0.25\columnwidth]{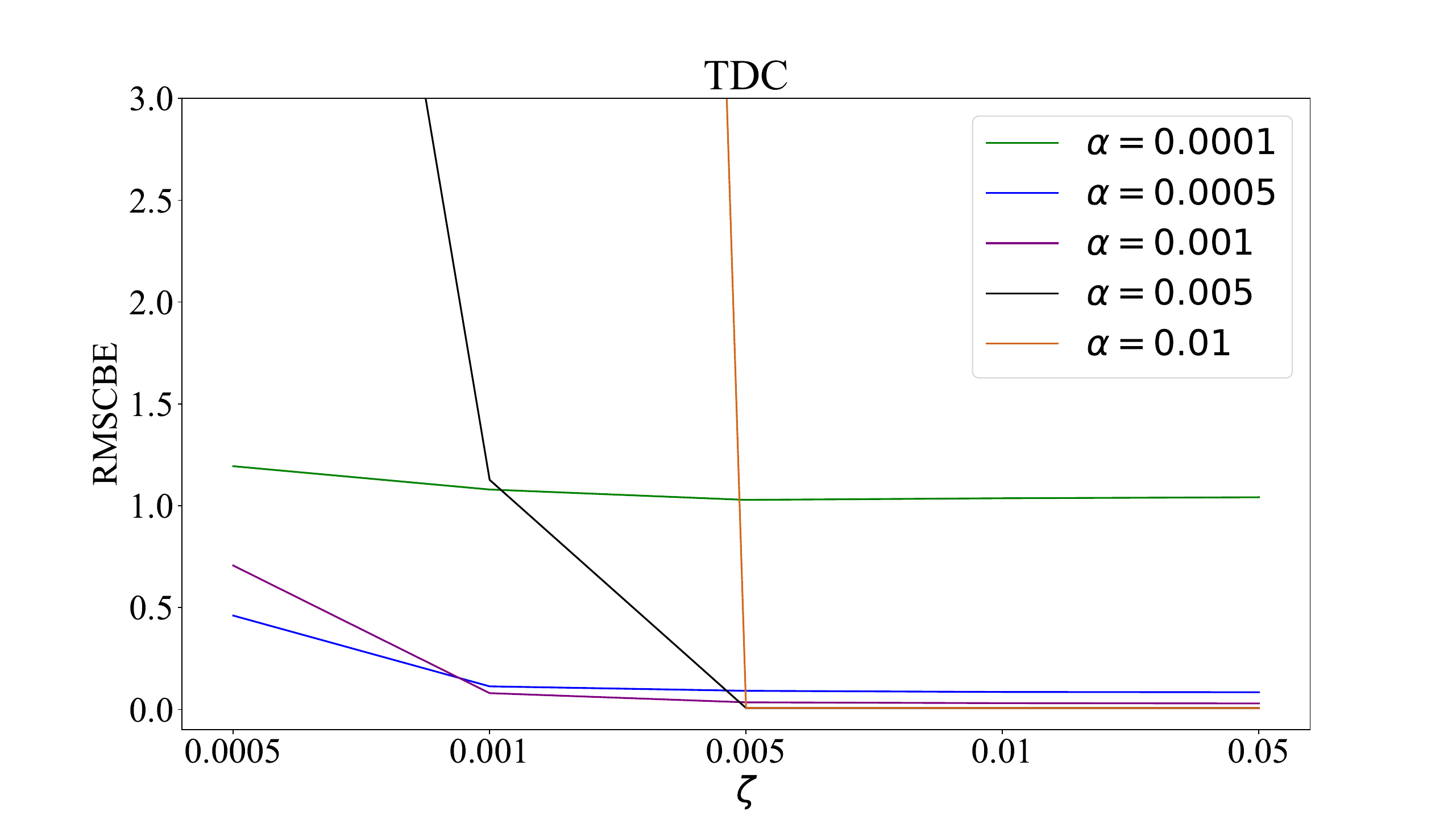}
        \label{tdc_7state}
    }
    \subfigure[CTD]{
        \includegraphics[width=0.3\columnwidth, height=0.25\columnwidth]{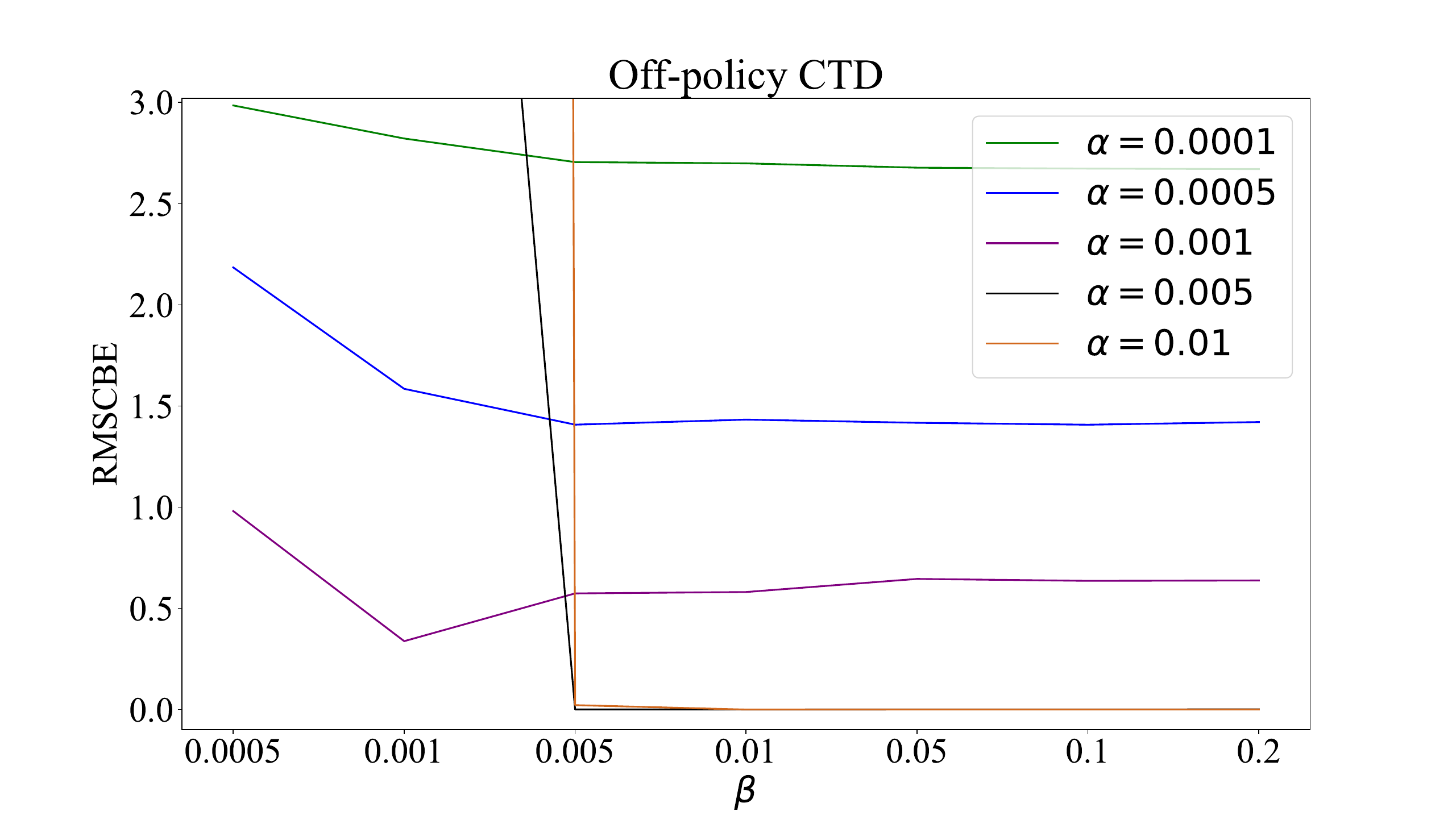}
        \label{ctd_7state}
    }
    \\
    \subfigure[CTDC($\alpha=0.0001$)]{
        \includegraphics[width=0.3\columnwidth, height=0.25\columnwidth]{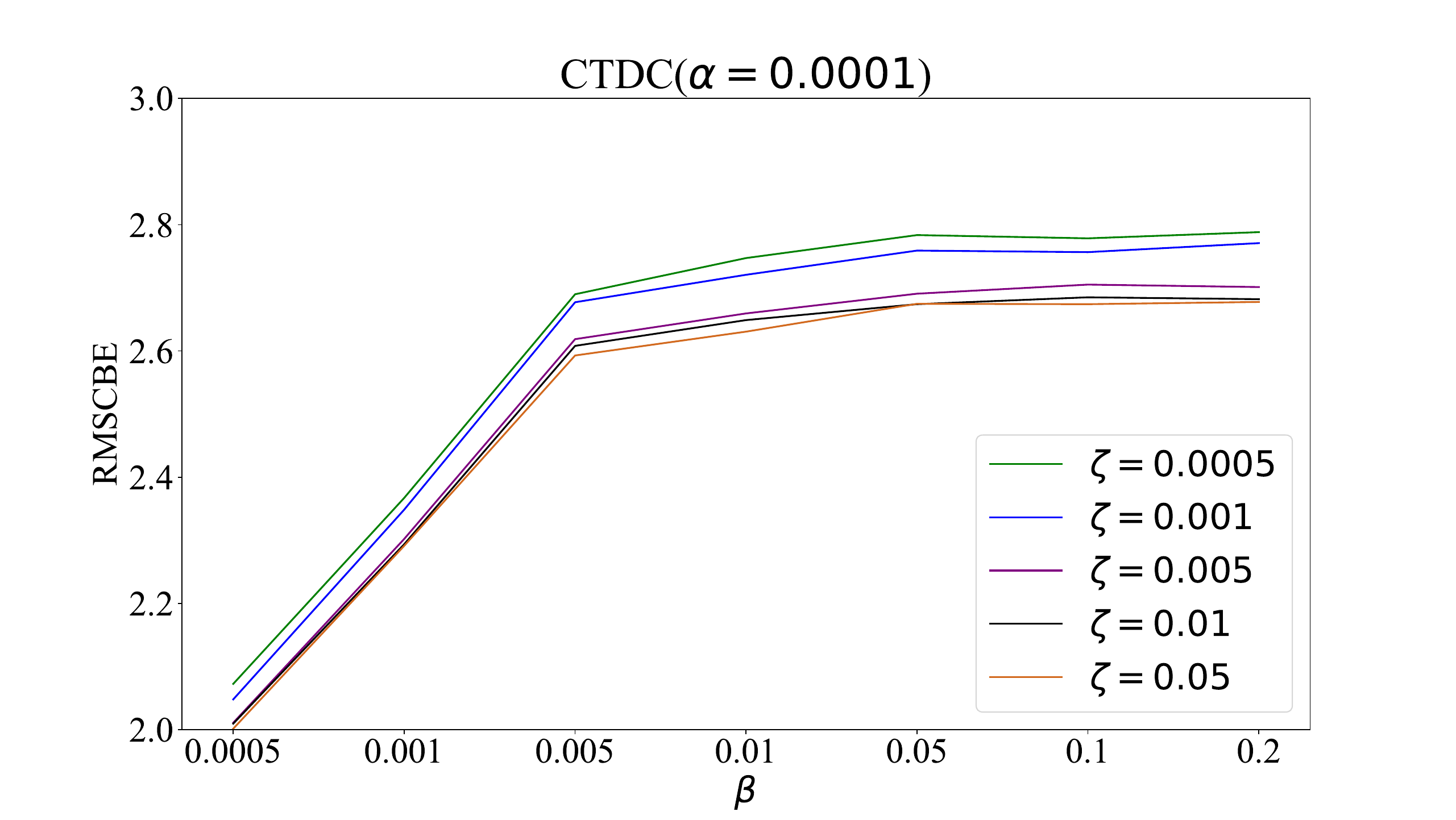}
        \label{ctdc_alpha_0001_7state}
    }
    \subfigure[CTDC($\alpha=0.0005$)]{
        \includegraphics[width=0.3\columnwidth, height=0.25\columnwidth]{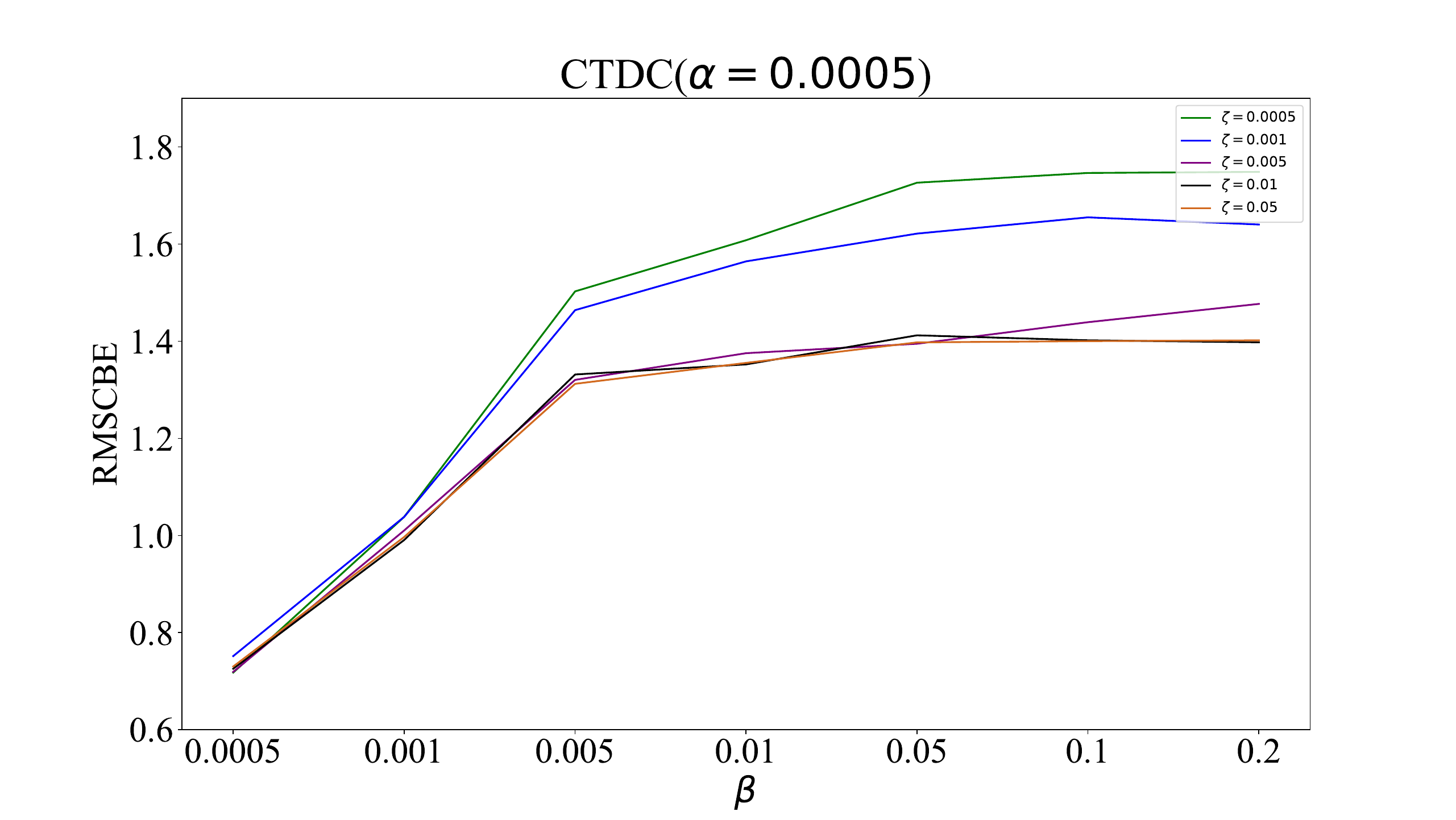}
        \label{ctdc_alpha_0005_7state}
    }
    \subfigure[CTDC($\alpha=0.001$)]{
        \includegraphics[width=0.3\columnwidth, height=0.25\columnwidth]{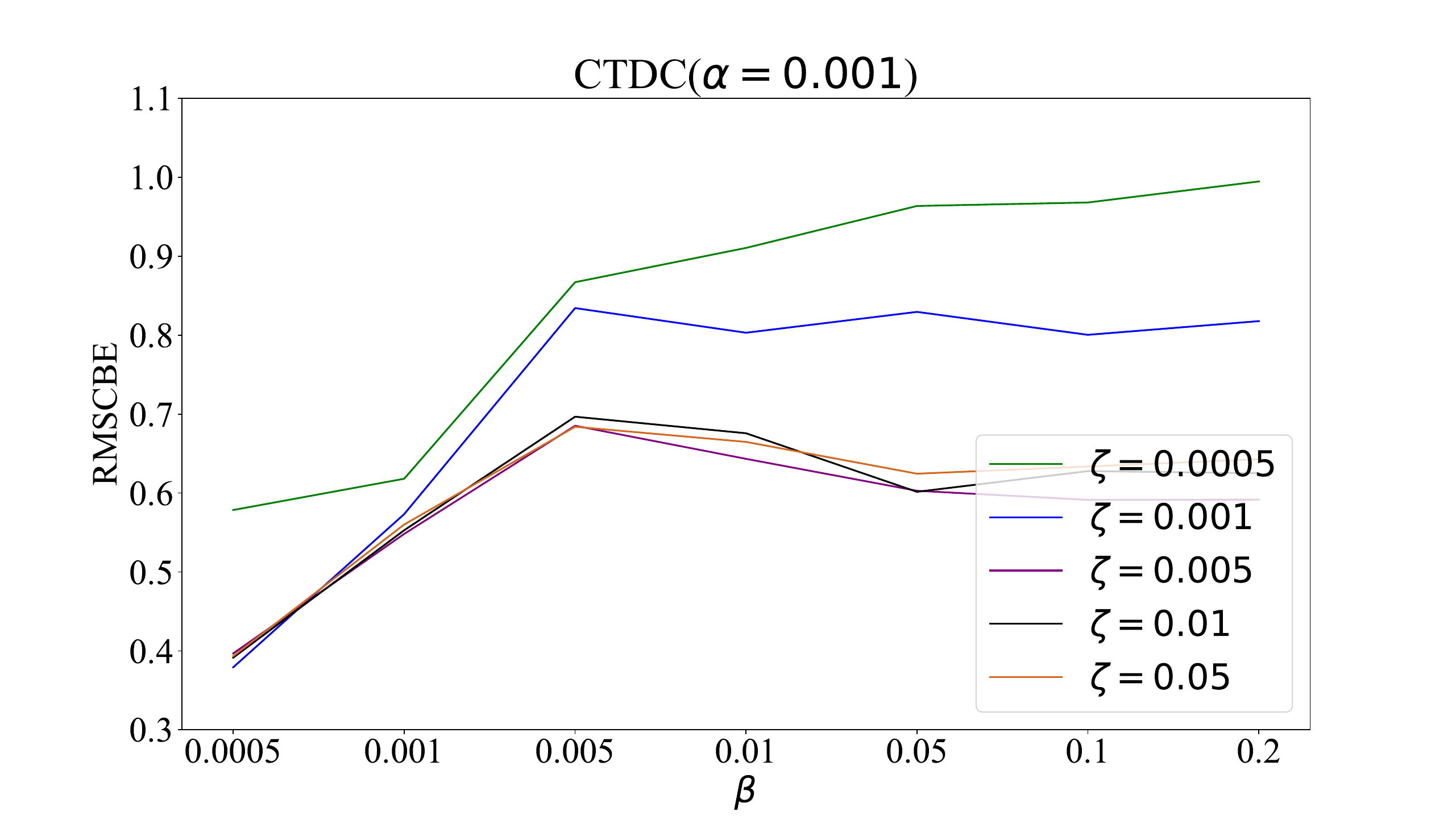}
        \label{ctdc_alpha_001_7state}
    }
    \\
    \subfigure[CTDC($\alpha=0.005$)]{
        \includegraphics[width=0.3\columnwidth, height=0.25\columnwidth]{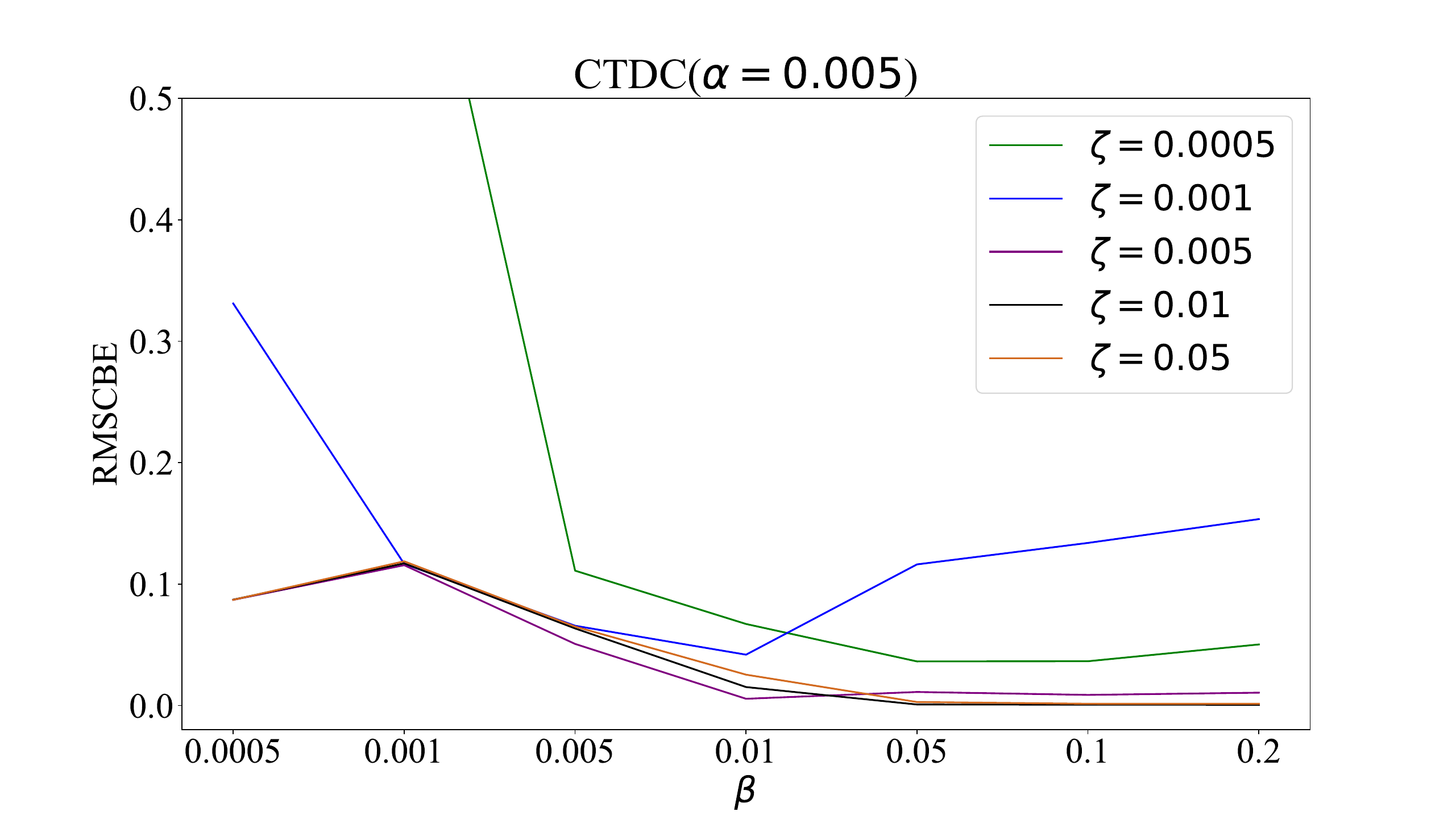}
        \label{ctdc_alpha_005_7state}
    }
    \subfigure[CTDC($\alpha=0.01$)]{
        \includegraphics[width=0.3\columnwidth, height=0.25\columnwidth]{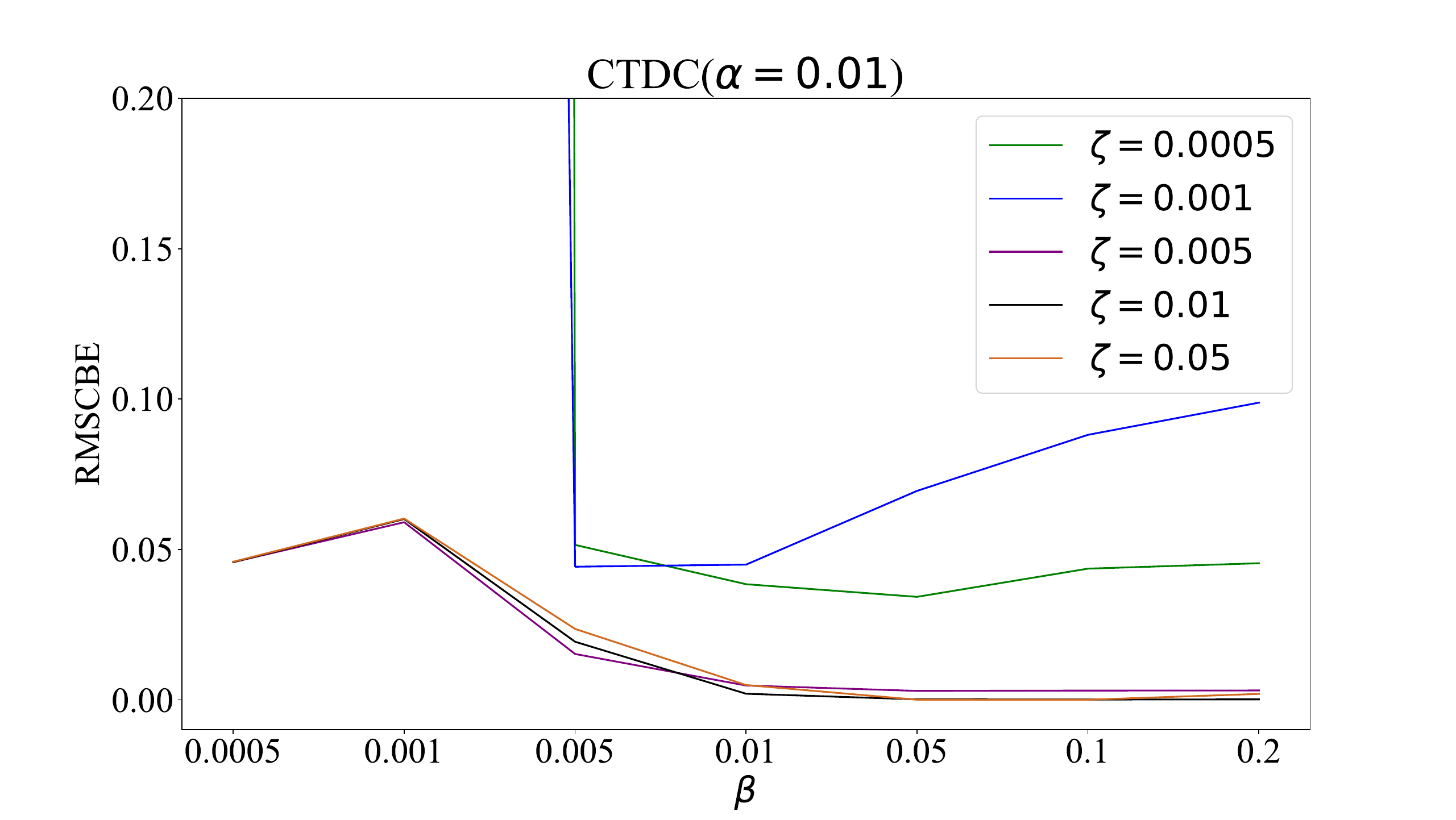}
        \label{ctdc_alpha_01_7state}
    }
        \caption{Sensitivity of various algorithms to learning rates for 7-state counterexample.}
        \label{Sensitivity7state}
    \end{center}
    \vskip -0.2in
\end{figure}

\end{document}